%% file: main.tex
\documentclass[10pt,journal,compsoc]{IEEEtran}
\usepackage{bm}
\usepackage{url}
\usepackage{bbding}
\usepackage{balance}
\usepackage{booktabs}
\usepackage{makecell}
\usepackage{multirow}
\usepackage{multicol}
\usepackage{hyperref}
\usepackage{proof}
\usepackage{amsmath,amsfonts}
\usepackage{amssymb}

\usepackage{setspace}
\usepackage{amsmath}
\usepackage{pifont}
\usepackage{adjustbox}
\usepackage{makecell}
\usepackage{mathtools}
\usepackage{enumitem}
\usepackage{colortbl}
\usepackage{amsfonts}       
\usepackage{amsmath}
\usepackage{bbm}
\definecolor{nlp-green}{rgb}{0.851,0.949,0.816}
\definecolor{vision-yellow}{rgb}{0.965,0.925,0.788}
\definecolor{multimodal-pink}{rgb}{0.965,0.867,0.929}
\definecolor{recsys-cyan}{rgb}{0.745,0.961,1.000}
\usepackage{tikz}
\usepackage[edges]{forest}
\definecolor{hiddendraw}{RGB}{205, 44, 36}
\definecolor{hidden-blue}{RGB}{194,232,247}
\definecolor{hidden-orange}{RGB}{243,202,120}
\definecolor{hidden-yellow}{RGB}{242,244,193}
\tikzstyle{mybox}=[
    rectangle,
    draw=hiddendraw,
    rounded corners,
    text opacity=1,
    minimum height=1.5em,
    minimum width=5em,
    inner sep=2pt,
    align=center,
    fill opacity=.5,
    ]
\usepackage{amssymb}
\usepackage{amsthm}

\usepackage[misc]{ifsym}

\ifCLASSOPTIONcompsoc
  \usepackage[nocompress]{cite}
\else
  \usepackage{cite}
\fi

\ifCLASSINFOpdf
\else
\fi

\begin{document}
%
\title{A Survey on Mixture of Experts in Large Language Models}
%
%
%
%

\author{Weilin Cai$^*$, Juyong Jiang$^*$, Fan Wang$^*$, Jing Tang$^\dag$, Sunghun Kim$^\dag$, Jiayi Huang$^\dag$
\IEEEcompsocitemizethanks{
\IEEEcompsocthanksitem The work of Weilin Cai and Jiayi Huang was supported in part by the National Natural Science Foundation of China (No. 62402411), the Guangdong Provincial Project (No. 2023QN10X252), the Guangdong Basic and Applied Basic Research Foundation (No. 2023A1515110353), the Guangzhou Municipal Science and Technology Project (No. 2024A04J4528), and the Guangzhou-HKUST(GZ) Joint Funding Program (No. 2024A03J0624).
Jing Tang's work is partially supported by National Key R\&D Program of China under Grant No. 2023YFF0725100 and No. 2024YFA1012701, by the National Natural Science Foundation of China (NSFC) under Grant No. 62402410 and No. U22B2060, by Guangdong Provincial Project (No. 2023QN10X025), by Guangdong Basic and Applied Basic Research Foundation under Grant No. 2023A1515110131, by Guangzhou Municipal Science and Technology Bureau under Grant No. 2023A03J0667 and No. 2024A04J4454, by Guangzhou Municipal Education Bureau (No. 2024312263), and by Guangzhou Municipality Big Data Intelligence Key Lab (No. 2023A03J0012), Guangzhou Industrial Information and Intelligent Key Laboratory Project (No. 2024A03J0628) and Guangzhou Municipal Key Laboratory of Financial Technology Cutting-Edge Research (No. 2024A03J0630).
\IEEEcompsocthanksitem The authors are with The Hong Kong University of Science and Technology (Guangzhou), Guangzhou, China. 
\IEEEcompsocthanksitem Weilin Cai, Juyong Jiang, and Fan Wang are the Equal Contributions (e-mail: \{wcai738, jjiang472, fwang380\}@connect.hkust-gz.edu.cn).
\IEEEcompsocthanksitem Jing Tang, Sunghun Kim, and Jiayi Huang are the Corresponding Authors (e-mail: \{jingtang, hunkim, hjy\}@hkust-gz.edu.cn).}
}

\markboth{Journal of \LaTeX\ Class Files,~Vol.~14, No.~8, August~2015}%
{J.Y. Jiang \MakeLowercase{\textit{et al.}}: Improving Sequential Recommendations via Bidirectional Temporal Data Augmentation with Pre-training}
%

\IEEEtitleabstractindextext{%
\begin{abstract}
Large language models (LLMs) have garnered unprecedented advancements across diverse fields, ranging from natural language processing to computer vision and beyond. The prowess of LLMs is underpinned by their substantial model size, extensive and diverse datasets, and the vast computational power harnessed during training, all of which contribute to the emergent abilities of LLMs (e.g., in-context learning) that are not present in small models. Within this context, the mixture of experts (MoE) has emerged as an effective method for substantially scaling up model capacity with minimal computation overhead, gaining significant attention from academia and industry. Despite its growing prevalence, there lacks a systematic and comprehensive review of the literature on MoE. This survey seeks to bridge that gap, serving as an essential resource for researchers delving into the intricacies of MoE. We first briefly introduce the structure of the MoE layer, followed by proposing a new taxonomy of MoE. Next, we overview the core designs for various MoE models including both algorithmic and systemic aspects, alongside collections of available open-source implementations, hyperparameter configurations and empirical evaluations. Furthermore, we delineate the multifaceted applications of MoE in practice, and outline some potential directions for future research. To facilitate ongoing updates and the sharing of cutting-edge advances in MoE research, we have established a resource repository at \href{https://github.com/withinmiaov/A-Survey-on-Mixture-of-Experts-in-LLMs}{https://github.com/withinmiaov/A-Survey-on-Mixture-of-Experts-in-LLMs}. 
\end{abstract}

}

\maketitle

\IEEEdisplaynontitleabstractindextext

%
\IEEEpeerreviewmaketitle

\input{section/1-introduction}

\input{section/2-background}
\input{section/3-taxonomy}
\input{section/4-overview}

\input{section/5-challenges_opportunities}
\input{section/6-conclusion}


%



\ifCLASSOPTIONcaptionsoff
  \newpage
\fi



%

\bibliographystyle{IEEEtran}
\bibliography{ref}

\end{document}

%% file: section/1-introduction.tex
\IEEEraisesectionheading{\section{Introduction}\label{sec:intro}}

\IEEEPARstart{I}{n} the current landscape of artificial general intelligence (AGI), the transformative impact of transformer-based large language models (LLMs) has permeated diverse fields such as natural language processing \cite{vaswani2017attention,brown2020language,chowdhery2023palm,achiam2023gpt,jiang2024survey}, computer vision \cite{riquelme2021scaling,liu2021swin}, and multimodality \cite{lu2019vilbert,zhou2022learning,zhu2023minigpt}. 
Building upon the foundational transformer architecture, LLMs demonstrate extraordinary capabilities, which are attributed to their sheer size, the breadth of data they are trained on, and the significant computational resources invested in their development \cite{kaplan2020scaling,wei2022emergent,yoo2024hyperclova}. 
Recognizing a scaling law \cite{kaplan2020scaling,hoffmann2022training} that underpins their evolution, it is imperative to identify and implement efficient methodologies for the sustainable scaling of LLMs.

\begin{figure*}[t]
\centering
\includegraphics[width=1\linewidth]{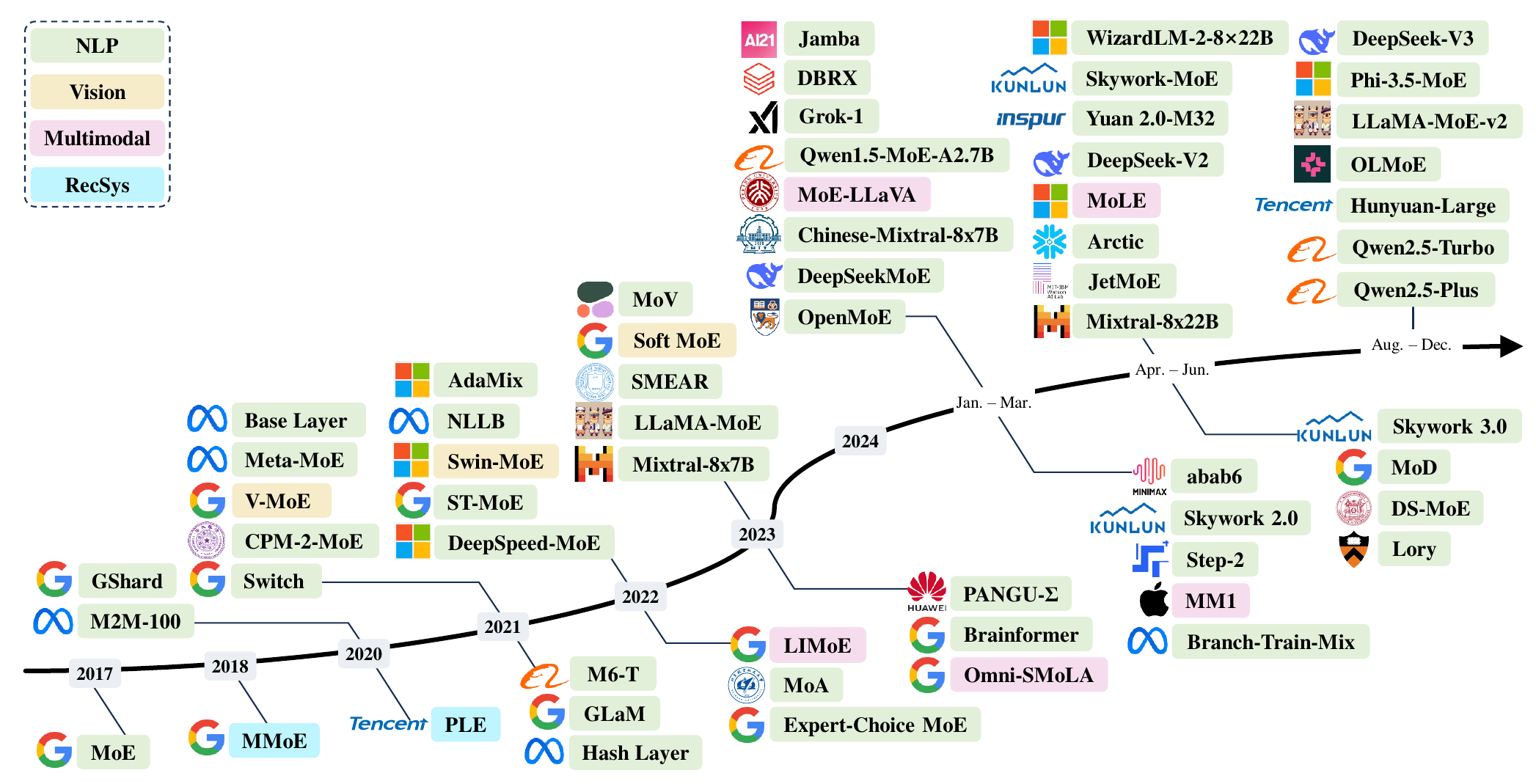}
\vspace{-0.25in}
\caption{A chronological overview of several representative mixture-of-experts (MoE) models in recent years. The timeline is primarily structured according to the release dates of the models. MoE models located above the arrow are open-source, while those below the arrow are proprietary and closed-source. MoE models from various domains are marked with distinct colors: Natural Language Processing (NLP) in \colorbox{nlp-green}{green}, Computer Vision in \colorbox{vision-yellow}{yellow}, Multimodal in \colorbox{multimodal-pink}{pink}, and Recommender Systems (RecSys) in \colorbox{recsys-cyan}{cyan}.}
\vspace{-0.05in}
\label{fig:moe_timeline}
\end{figure*}

The concept of mixture of experts (MoE), initially introduced in \cite{jacobs1991adaptive,jordan1994hierarchical}, has undergone extensive exploration and advancement as evidenced by subsequent studies \cite{collobert2001parallel, rasmussen2001infinite, shahbaba2009nonlinear, eigen2013learning, theis2015generative, deisenroth2015distributed, aljundi2017expert}. 
The emergence of sparsely-gated MoE \cite{shazeer2017outrageously}, particularly within the integration of transformer-based large language models \cite{lepikhin2020gshard}, has brought new vitality to this three-decade-old technology.
The MoE framework is based on a simple yet powerful idea: different parts of a model, known as experts, specialize in different tasks or aspects of the data. 
With this paradigm, only pertinent experts are engaged for a given input, keeping the computational cost in check while still benefiting from a large pool of specialized knowledge.
This scalable and flexible innovation has offered an effective approach for adhering to the scaling law, allowing for increased model capacity without a corresponding surge in computational demands.
As depicted in Figure~\ref{fig:moe_timeline}, MoE has maintained a robust trajectory of growth, particularly notable in 2024 with the advent of Mixtral-8x7B \cite{jiang2024mixtral} and a variety of subsequent industrial-scale LLMs such as Grok-1 \cite{Grok-1}, DBRX \cite{dbrx}, Arctic \cite{snowflake}, DeepSeek-V2 \cite{deepseekv2}, etc.

\textcolor{black}{
Despite the growing popularity and application of mixture of experts (MoE) models across various domains, comprehensive reviews that thoroughly examine and categorize advancements, particularly in the context of MoE in LLMs, remain scarce. 
Specifically, we identified two surveys preceding our work: the first, published in August 2012 \cite{yuksel2012twenty}, provides a comprehensive review of early studies on dense MoE, which significantly differ from the current mainstream focus on sparse MoE; the second, released in September 2022 \cite{fedus2022review}, predates the major developments following the ``ChatGPT moment'' and, as a result, does not cover the substantial advancements and increased interest from academia and industry that have since emerged. 
This gap in the literature not only hinders the progress of MoE research but also limits the broader dissemination of knowledge on this topic. 
Our survey aims to address this deficit by providing a clear and comprehensive overview of MoE in LLMs, introducing a novel taxonomy that organizes recent progress into three categories: algorithm, system, and application.
}

Under this taxonomy, we first delve into MoE algorithmic advancements, particularly the prevalent substitution of feed-forward network (FFN) layers with MoE layers in transformer-based LLMs \cite{lepikhin2020gshard,du2022glam,fedus2022switch,zoph2022st,jiang2024mixtral,xue2024openmoe,deepseekv2}. 
As each MoE layer integrates multiple FFNs---each designated as an expert---and employs a gating function to activate a selected subset of these experts, we explore the design choices of gating function and expert network, alongside collections of available open-source implementations, hyperparameter configurations, and empirical evaluations. 
Furthermore, to underscore the flexibility and versatility of MoE, we extend our analysis beyond the standard integration of MoE into model backbone, and discuss an array of novel MoE-related designs, such as soft MoE with token or expert merging \cite{puigcerver2023sparse,muqeeth2023soft,zhong2024lory,zadouri2023pushing,wu2023omni}, mixture of parameter-efficient experts (MoPEs) \cite{wang-etal-2022-adamix,zadouri2023pushing,dou2023loramoe,gou2023mixture,luo2024moelora,wu2024mixture}, training and inference schemes with model transition between dense and sparse \cite{komatsuzaki2022sparse,zhang2022moefication,llama-moe-2023,xue2022one,chen2022task,sukhbaatar2024branch}, and various derivatives \cite{chen2023lifelong,antoniak2023mixture,raposo2024mixture,xue2022go,tan2023sparse,choi2023smop}.

With the gradual convergence of model architecture design in industrial products, system design has emerged as a pivotal factor in enhancing the quality of LLM services. 
Given the close association of MoE models with machine learning system design, we provide a comprehensive overview of MoE system design, including computation, communication, and storage enhancements tailored to address the unique challenges posed by the sparse and dynamic nature of its computational workload. 
Additionally, we overview the applications of MoE across various domains, including natural language processing, computer vision, recommender system, and multimodal contexts.

The remainder of this survey is organized as follows.
Section~\ref{sec:background} provides a foundational understanding of MoE, contrasting sparse and dense activation of experts. 
Section~\ref{sec:taxonomy} introduces our proposed taxonomy for categorizing MoE advancements. 
Sections~\ref{sec:algorithm}, \ref{sec:system}, and \ref{sec:app} delve into the algorithmic designs, computing system support, and various applications of MoE models, respectively, following the structure outlined in our taxonomy in Figure~\ref{fig:taxonomy}. 
Finally, in Section~\ref{sec:challenges}, we highlight the critical challenges and opportunities for bridging the research-practicality gap, culminating in Section~\ref{sec:discussion} with our conclusions.


%% file: section/2-background.tex
\begin{figure*}[t]
\centering
\includegraphics[width=0.7\linewidth]{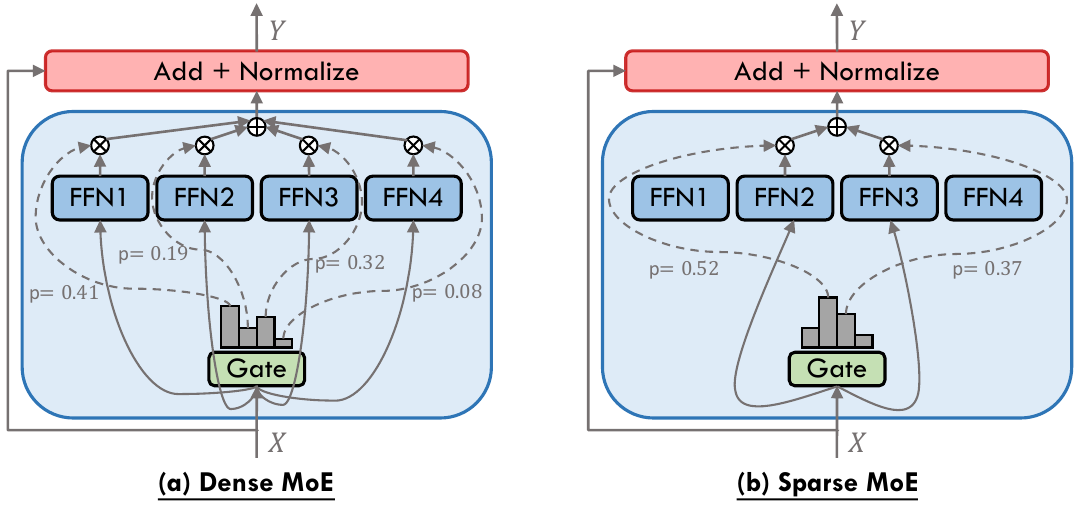}
\caption{An illustration of an MoE layer in Transformer-based models. For each input $X$, the linear-softmax gating will select all experts namely \textbf{(a) Dense MoE} or top-$k$ experts namely \textbf{(b) Sparse MoE} to perform conditional computation. 
The expert layer returns the output of the selected expert multiplied by the gate value (softmax of the gating function output).}
\label{fig:moe_layer}
\end{figure*}

\section{Background on Mixture of Experts}\label{sec:background}
In transformer-based large language models (LLMs), each mixture-of-experts (MoE) layer typically consists of a set of $N$ ``expert networks'' $\{f_{1}, \ldots, f_{N}\}$, alongside a ``gating network'' $\mathcal{G}$. 
The role of the gating network, which often takes the form of a linear network with a softmax activation function, is to direct the input to the appropriate expert networks \cite{shazeer2017outrageously,fedus2022switch}. 
The MoE layer is strategically placed to select the feed-forward network (FFN) within each Transformer block, typically following the self-attention (SA) sub-layer. 
This positioning is crucial because the FFN become increasingly computationally demanding as the model scales up. 
For instance, in the PaLM \cite{chowdhery2023palm} model with the parameter number of 540B, the 90\% of these parameters are within its FFN layers.

Formally, each expert network $f_i$, usually a linear-ReLU-linear network, is parameterized by $\mathbf{W}_i$, accepting the same input $\mathbf{x}$ and generating an output $f_{i}(\mathbf{x}; \mathbf{W}_i)$.
In parallel, the gating network $\mathcal{G}$, parameterized by $\mathbf{\Theta}$ and typically consisting of a linear-softmax network, yields the output $\mathcal{G}(\mathbf{x}; \mathbf{\Theta})$. 
Based on the design of gating function, MoE layers can be broadly classified into two categories: dense MoE and sparse MoE, which are described in detail in the following subsections.

\subsection{Dense MoE}
\label{sec:background_densemoe}
The dense mixture-of-experts layer activates all expert networks $\{f_{1}, \ldots, f_{N}\}$ during each iteration. 
This strategy has been extensively employed in a range of early proposals \cite{jacobs1991adaptive, jordan1994hierarchical, collobert2001parallel, rasmussen2001infinite, shahbaba2009nonlinear, eigen2013learning, theis2015generative, deisenroth2015distributed, aljundi2017expert,ma2018modeling}.
Most recently, the dense MoE concept has been revisited by studies such as EvoMoE\cite{nie2021evomoe}, MoLE \cite{wu2023mole}, LoRAMoE \cite{dou2023loramoe}, and DS-MoE\cite{pan2024dense}. 
The structure of the dense MoE layer is depicted in Figure \ref{fig:moe_layer}(a).
Consequently, the output of the dense MoE layer can be formulated as
\begin{align}
&\mathcal{F}_{\mathrm{dense}}^{\mathrm{MoE}}(\mathbf{x}; \mathbf{\Theta}, \{\mathbf{W}_i\}_{i=1}^{N}) =
\sum_{i=1}^{N}\mathcal{G}(\mathbf{x}; \mathbf{\Theta})_i f_{i}(\mathbf{x}; \mathbf{W}_i),\label{eq:dense_moe}\\
&\mathcal{G}(\mathbf{x}; \mathbf{\Theta})_i = \operatorname{softmax}(g(\mathbf{x}; \mathbf{\Theta}))_i = \frac{\exp(g(\mathbf{x}; \mathbf{\Theta})_i)}{\sum_{j=1}^{N} \exp(g(\mathbf{x}; \mathbf{\Theta})_j)}, \label{eq:dense_gating}
\end{align}
where $g(\mathbf{x}; \mathbf{\Theta})$ represents the gating value prior to the softmax operation.

\subsection{Sparse MoE}
\label{sec:background_sparsemoe}
While dense MoE typically yields higher prediction accuracy \cite{riquelme2021scaling}, it also incurs a significant increase in computational overhead. 
To address this, Shazeer \textit{et al.} \cite{shazeer2017outrageously} introduced the sparsely-gated MoE layer, which is designed to activate only a selected subset of experts during each forward pass. 
This strategy achieves sparsity by computing a weighted sum of the outputs from only the top-$k$ experts, rather than aggregating the outputs from all the experts. 
The structure of the sparse MoE layer is illustrated in Figure \ref{fig:moe_layer}(b).
Building on the framework established by \cite{shazeer2017outrageously}, Equation \eqref{eq:dense_gating} can be modified to reflect the sparsely-gated mechanism as follows:
\begin{align}
    &\mathcal{G}(\mathbf{x}; \mathbf{\Theta})_i = \operatorname{softmax}(\operatorname{TopK}(g(\mathbf{x}; \mathbf{\Theta}) + \mathcal{R}_{\mathrm{noise}}, k))_i,\label{eq:sparse_gating}\\
    &\operatorname{TopK}(g(\mathbf{x}; \mathbf{\Theta}), k)_i = 
    \begin{cases}
    g(\mathbf{x}; \mathbf{\Theta})_i, &\text{condition}, \\
    -\infty, &\text{otherwise}.
    \end{cases},\\
    &\mathrm{condition}:\text{if $g(\mathbf{x}; \mathbf{\Theta})_i$ is in the top-$k$ elements of $g(\mathbf{x}; \mathbf{\Theta})$}.
    \label{eq:topk}
\end{align}
To explain, $\operatorname{TopK}(\cdot, k)$ function retains only the top-$k$ entries of a vector at their original values, while setting all other entries to $-\infty$. 
Following the $\operatorname{softmax}$ operation, those entries assigned $-\infty$ become approximately zero. 
The hyper-parameter $k$ is selected based on the specific application, with common choices being $k=1$ \cite{fedus2022switch,clark2022unified} or $k=2$ \cite{lepikhin2020gshard,du2022glam,rajbhandari2022deepspeed,zoph2022st,jiang2024mixtral,wei2024skywork}.
The addition of a noise term $\mathcal{R}_{\operatorname{noise}}$ is a prevalent strategy for training a sparsely-gated MoE layer, fostering exploration among experts and enhancing the stability of MoE training \cite{shazeer2017outrageously,fedus2022switch}.

Although the sparse gate $\mathcal{G}(\mathbf{x}; \mathbf{\Theta})$ substantially expands the model's parameter space without a corresponding increase in computational cost, it can lead to a load balancing issue. Such an issue refers to the uneven distribution of workload across experts, with some being frequently utilized and others seldom engaged. 
To address this, each MoE layer incorporates an auxiliary loss function that promotes an even distribution of tokens across experts within each batch, as described in many studies \cite{lepikhin2020gshard,fedus2022switch,du2022glam,jiang2024mixtral,lieber2024jamba,dai2024deepseekmoe,wei2024skywork}.
To formulate this concept, consider a batch of queries $\mathcal{B} = \{\mathbf{x}_i, \mathbf{x}_2, \ldots, \mathbf{x}_T\}$, comprising $T$ tokens, and $N$ experts indexed from $i=1$ to $N$. 
Following \cite{lepikhin2020gshard,fedus2022switch}, the auxiliary load balancing loss for the batch is defined as
\begin{align}
&\mathcal{L}_{\operatorname{load-balancing}} = N\sum_{i=1}^{N}\mathcal{D}_i\mathcal{P}_i,\label{eq:load_balance}\\
&\mathcal{D}_i = \frac{1}{T}\sum_{x \in \mathcal{B}}\mathbbm{1}\{\operatorname{argmax} \mathcal{G}(\mathbf{x}; \mathbf{\Theta}) = i\},\label{eq:token_sum}\\
&\mathcal{P}_i = \frac{1}{T}\sum_{x \in \mathcal{B}} \mathcal{G}(\mathbf{x}; \mathbf{\Theta})_i,\label{eq:prob_sum}
\end{align}
where $\mathcal{D}_i$ represents the proportion of tokens distributed to expert $i$, while $\mathcal{P}_i$ denotes the proportion of the gating probability assigned to expert $i$. 
To ensure an even distribution of the batch of tokens across the $N$ experts, the load-balancing loss function $\mathcal{L}_{\operatorname{load-balancing}}$ should be minimized.
The optimal condition, i.e.,~$\min(\mathcal{L}_{\operatorname{load-balancing}}) = N\sum_{i=1}^{N}\mathcal{D}_i\mathcal{P}_i = N\sum_{i=1}^{N}\frac{1}{N}\frac{1}{N} = 1$, is achieved when each expert receives an equal number of dispatched tokens $\mathcal{D}_i = \frac{1}{N}$, and an equal proportion of the gating probability $\mathcal{P}_i = \frac{1}{N}$. 
The balance is thus maintained across all the experts, ensuring that the workload is uniformly distributed at all times.
Throughout the subsequent sections, unless explicitly stated otherwise, the term ``MoE'' refers to ``sparse MoE''.

%% file: section/3-taxonomy.tex
\section{Taxonomy of Mixture of Experts}\label{sec:taxonomy}
To effectively scale model parameters without a corresponding increase in computational demand, the mixture of experts (MoE) architecture has emerged as a viable solution. 
MoE leverages a collection of specialized models and a gating mechanism to dynamically select the appropriate ``expert networks'' for processing a given input. 
This enables the model to allocate computational resources on an as-needed basis, a concept known as conditional computation. 
The incorporation of MoE architectures into large language models (LLMs) is now a prevalent practice, allowing these models to achieve significant parameter scale-ups and consequent enhancements in capabilities \cite{fedus2022review,lepikhin2020gshard,fedus2022switch,jiang2024mixtral,wei2024skywork}.

\definecolor{line-color}{RGB}{0, 119, 182}
\definecolor{fill-color}{RGB}{114, 200, 222}

\tikzstyle{category}=[
    rectangle,
    draw=line-color,
    rounded corners,
    text opacity=1,
    minimum height=1.5em,
    minimum width=5em,
    inner sep=2pt,
    align=center,
    fill opacity=.5,
]
\tikzstyle{leaf}=[category,minimum height=1.5em,
fill=fill-color!40, text width=20em,  text=black,align=left,font=\scriptsize,
inner xsep=2pt,
inner ysep=1pt,
]

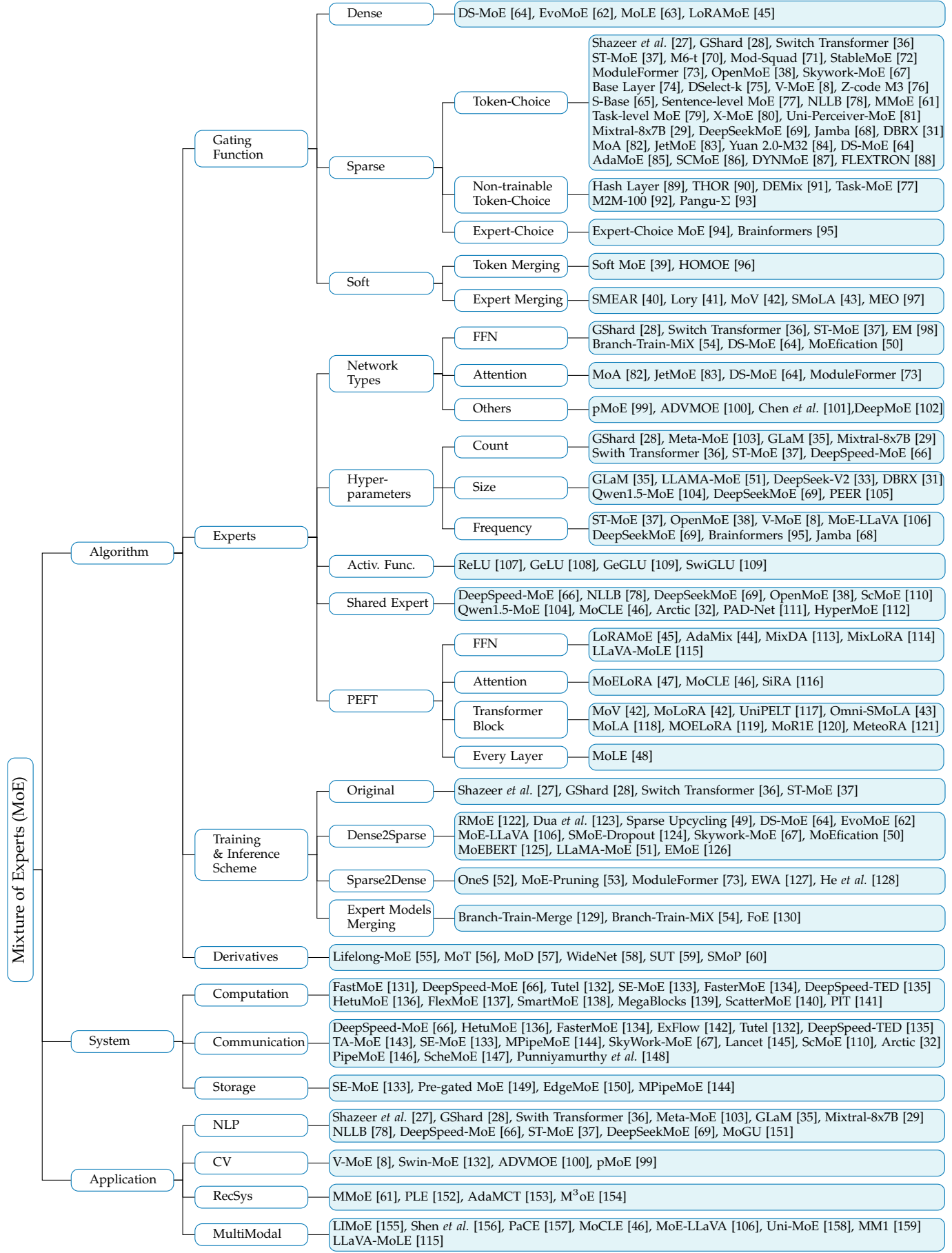
\begin{figure*}[tp]
  \centering
\begin{forest}
  forked edges,
  for tree={
  grow=east,
  reversed=true,
  anchor=base west,
  parent anchor=east,
  child anchor=west,
  base=left,
  font=\small,
  rectangle,
  draw=line-color,
  rounded corners,align=left,
  minimum width=2.5em,
s sep=4pt,
inner xsep=10pt,
inner ysep=2pt,
align=left,
ver/.style={rotate=90, child anchor=north, parent anchor=south, anchor=center},
  },
  where level=1{text width=3.8em,font=\scriptsize,}{},
  where level=2{text width=4.4em,font=\scriptsize}{},
  where level=3{text width=3.90em,font=\scriptsize}{},
  where level=4{text width=4.4em,font=\scriptsize}{},
  [Mixture of Experts (MoE), ver
    [Algorithm
        [Gating \\Function
            [Dense
                [
                DS-MoE\cite{pan2024dense}{,} EvoMoE\cite{nie2021evomoe}{,} MoLE\cite{wu2023mole}{,} LoRAMoE\cite{dou2023loramoe}
                ,leaf,text width=27.8em]
            ]
            [Sparse
                [Token-Choice
                    [Shazeer \textit{et al.}\cite{shazeer2017outrageously}{,} GShard\cite{lepikhin2020gshard}{,} Switch Transformer\cite{fedus2022switch}\\
                    ST-MoE\cite{zoph2022st}{,} M6-t\cite{yang2021m6}{,} Mod-Squad\cite{chen2023mod}{,} StableMoE\cite{dai2022stablemoe}\\
                    ModuleFormer\cite{shen2023moduleformer}{,} OpenMoE\cite{xue2024openmoe}{,} Skywork-MoE \cite{wei2024skywork}\\
                    Base Layer\cite{lewis2021base}{,} DSelect-k\cite{hazimeh2021dselect}{,} V-MoE\cite{riquelme2021scaling}{,} Z-code M3\cite{kim2021scalable}\\
                    S-Base\cite{clark2022unified}{,} Sentence-level MoE\cite{kudugunta2021beyond}{,} NLLB\cite{costa2022no}{,} MMoE\cite{ma2018modeling}\\
                    Task-level MoE\cite{ye2022eliciting}{,} X-MoE\cite{chi2022representation}{,} Uni-Perceiver-MoE\cite{zhu2022uni}\\
                    Mixtral-8x7B\cite{jiang2024mixtral}{,} DeepSeekMoE\cite{dai2024deepseekmoe}{,} Jamba\cite{lieber2024jamba}{,} DBRX\cite{dbrx}\\ MoA\cite{zhang2022mixture}{,} JetMoE \cite{shen2024jetmoe}{,} Yuan 2.0-M32 \cite{wu2024yuan}{,} DS-MoE\cite{pan2024dense} \\ AdaMoE\cite{zeng2024adamoe}{,} SCMoE\cite{shi2024unchosen}{,} DYNMoE\cite{guo2024dynamic}{,} FLEXTRON\cite{caiflextron}
                    ,leaf,text width=20em]
                ]
                [Non-trainable\\ Token-Choice
                    [Hash Layer\cite{roller2021hash}{,} THOR\cite{zuo2021taming}{,} DEMix\cite{gururangan2022demix}{,} Task-MoE\cite{kudugunta2021beyond}\\
                    M2M-100\cite{fan2021beyond}{,} Pangu-$\Sigma$\cite{ren2023pangu}
                    ,leaf,text width=20em]
                ]
                [Expert-Choice
                    [Expert-Choice MoE\cite{zhou2022mixture}{,} Brainformers\cite{zhou2023brainformers}
                    ,leaf,text width=20em]
                ]
            ]
            [Soft
                [Token Merging
                    [Soft MoE\cite{puigcerver2023sparse}{,} HOMOE\cite{dat2023homoe}
                    ,leaf,text width=20em]
                ]
                [Expert Merging
                    [SMEAR\cite{muqeeth2023soft}{,} Lory\cite{zhong2024lory}{,} MoV\cite{zadouri2023pushing}{,} SMoLA\cite{wu2023omni}{,} MEO\cite{he2023merging}
                    ,leaf,text width=20em]
                ]
            ]
        ]
        [Experts
            [Network \\Types
                [FFN
                    [GShard\cite{lepikhin2020gshard}{,} Switch Transformer\cite{fedus2022switch}{,} ST-MoE\cite{zoph2022st}{,} EM\cite{zhang2023emergent} \\Branch-Train-MiX\cite{sukhbaatar2024branch}{,} DS-MoE\cite{pan2024dense}{,} MoEfication\cite{zhang2022moefication}
                    ,leaf,text width=20em]
                ]
                [Attention
                    [MoA\cite{zhang2022mixture}{,} JetMoE\cite{shen2024jetmoe}{,} DS-MoE\cite{pan2024dense}{,} ModuleFormer\cite{shen2023moduleformer}
                    ,leaf,text width=20em]
                ]
                [Others
                    [pMoE\cite{chowdhury2023patch}{,} ADVMOE\cite{zhang2023robust}{,} Chen \textit{et al.}\cite{chen2022towards}{,}DeepMoE\cite{wang2020deep}
                    ,leaf,text width=20em]
                ]
            ]
            [Hyper-\\parameters
                [Count
                    [GShard\cite{lepikhin2020gshard}{,} Meta-MoE\cite{artetxe2021efficient}{,} GLaM\cite{du2022glam}{,} Mixtral-8x7B\cite{jiang2024mixtral} \\ Swith Transformer\cite{fedus2022switch}{,} ST-MoE\cite{zoph2022st}{,} DeepSpeed-MoE\cite{rajbhandari2022deepspeed}
                    ,leaf,text width=20em]
                ]
                [Size
                    [GLaM\cite{du2022glam}{,} LLAMA-MoE \cite{llama-moe-2023}{,} DeepSeek-V2\cite{deepseekv2}{,} DBRX\cite{dbrx} \\Qwen1.5-MoE\cite{qwen_moe}{,} DeepSeekMoE\cite{dai2024deepseekmoe}{,} PEER\cite{he2024mixture}
                    ,leaf,text width=20em]
                ]
                [Frequency
                    [ST-MoE\cite{zoph2022st}{,} OpenMoE\cite{xue2024openmoe}{,} V-MoE\cite{riquelme2021scaling}{,} MoE-LLaVA\cite{lin2024moe} \\DeepSeekMoE\cite{dai2024deepseekmoe}{,} Brainformers\cite{zhou2023brainformers}{,} Jamba\cite{lieber2024jamba}
                    ,leaf,text width=20em]
                ]
            ]
            [Activ. Func.
                [ReLU\cite{glorot2011deep}{,} GeLU\cite{hendrycks2016gaussian}{,} GeGLU\cite{shazeer2020glu}{,} SwiGLU\cite{shazeer2020glu}
                ,leaf,text width=27.8em]
            ]
            [Shared Expert
                [DeepSpeed-MoE\cite{rajbhandari2022deepspeed}{,} NLLB\cite{costa2022no}{,} DeepSeekMoE\cite{dai2024deepseekmoe}{,} OpenMoE\cite{xue2024openmoe}{,} ScMoE\cite{cai2024shortcut} \\ Qwen1.5-MoE\cite{qwen_moe}{,} MoCLE\cite{gou2023mixture}{,} Arctic\cite{snowflake}{,} PAD-Net\cite{he2023pad}{,} HyperMoE\cite{zhao2024hypermoe}
                ,leaf,text width=27.8em]
            ]
            [PEFT
                [FFN
                    [LoRAMoE\cite{dou2023loramoe}{,}
                    AdaMix\cite{wang-etal-2022-adamix}{,}
                    MixDA\cite{diao2023mixture}{,}
                    MixLoRA\cite{li2024mixlora}\\
                    LLaVA-MoLE\cite{chen2024llava}
                    ,leaf,text width=20em]
                ]
                [Attention
                    [MoELoRA\cite{luo2024moelora}{,}
                    MoCLE\cite{gou2023mixture}{,}
                    SiRA\cite{zhu2023sira}
                    ,leaf,text width=20em]
                ]
                [Transformer \\Block
                    [MoV\cite{zadouri2023pushing}{,}
                    MoLoRA\cite{zadouri2023pushing}{,}
                    UniPELT\cite{mao2022unipelt}{,}
                    Omni-SMoLA\cite{wu2023omni}\\
                    MoLA\cite{gao2024higher}{,}
                    MOELoRA\cite{liu2023moelora}{,}
                    MoR1E\cite{liu2024intuition}{,}
                    MeteoRA\cite{xu2024meteora}
                    ,leaf,text width=20em]
                ]
                [Every Layer
                    [MoLE\cite{wu2024mixture}
                    ,leaf,text width=20em]
                ]
            ]
        ]
        [Training \\ \& Inference \\Scheme
            [Original
                [Shazeer \textit{et al.}\cite{shazeer2017outrageously}{,} GShard\cite{lepikhin2020gshard}{,} Switch Transformer\cite{fedus2022switch}{,} ST-MoE\cite{zoph2022st}
                ,leaf,text width=27.8em]
            ]
            [Dense2Sparse
                [RMoE\cite{wu2022residual}{,} Dua \textit{et al.}\cite{dua2022tricks}{,} Sparse Upcycling\cite{komatsuzaki2022sparse}{,} DS-MoE\cite{pan2024dense}{,} EvoMoE\cite{nie2021evomoe} \\MoE-LLaVA\cite{lin2024moe}{,} SMoE-Dropout\cite{chen2022sparse}{,} Skywork-MoE\cite{wei2024skywork}{,} MoEfication \cite{zhang2022moefication} \\MoEBERT\cite{zuo2022moebert}{,} LLaMA-MoE\cite{llama-moe-2023}{,} EMoE\cite{qiu2024unlocking}
                ,leaf,text width=27.8em]
            ]
            [Sparse2Dense
                [OneS\cite{xue2022one}{,} MoE-Pruning\cite{chen2022task}{,} ModuleFormer \cite{shen2023moduleformer}{,} EWA\cite{huang2023experts}{,} He \textit{et al.}\cite{he2024demystifying}
                ,leaf,text width=27.8em]
            ]
            [Expert Models \\Merging
                [Branch-Train-Merge\cite{li2022branch}{,} Branch-Train-MiX\cite{sukhbaatar2024branch}{,} FoE\cite{wang2023fusing}
                ,leaf,text width=27.8em]
            ]
        ]
        [Derivatives
            [Lifelong-MoE\cite{chen2023lifelong}{,} MoT\cite{antoniak2023mixture}{,} MoD\cite{raposo2024mixture}{,} WideNet\cite{xue2022go}{,} SUT\cite{tan2023sparse}{,} SMoP\cite{choi2023smop}
            ,leaf,text width=35em]
        ]
    ]
    [System
        [Computation
            [FastMoE\cite{he2021fastmoe}{,} DeepSpeed-MoE\cite{rajbhandari2022deepspeed}{,} Tutel\cite{hwang2023tutel}{,} SE-MoE\cite{shen2022se}{,} FasterMoE\cite{he2022fastermoe}{,} DeepSpeed-TED\cite{singh2023hybrid} \\HetuMoE\cite{nie2022hetumoe}{,} FlexMoE\cite{nie2023flexmoe}{,} SmartMoE\cite{zhai2023smartmoe}{,} MegaBlocks\cite{gale2023megablocks}{,}
            ScatterMoE\cite{tan2024scattered}{,}
            PIT\cite{zheng2023pit}
            ,leaf,text width=35em]
        ]
        [Communication
            [DeepSpeed-MoE\cite{rajbhandari2022deepspeed}{,} HetuMoE\cite{nie2022hetumoe}{,} FasterMoE\cite{he2022fastermoe}{,} ExFlow\cite{yao2024exploiting}{,} Tutel\cite{hwang2023tutel}{,} DeepSpeed-TED\cite{singh2023hybrid} \\ TA-MoE\cite{chen2022ta}{,} SE-MoE\cite{shen2022se}{,} 
            MPipeMoE \cite{zhang2024mpmoe}{,} SkyWork-MoE\cite{wei2024skywork}{,} Lancet\cite{jiang2024lancet}{,} ScMoE\cite{cai2024shortcut}{,} Arctic\cite{snowflake} \\ PipeMoE\cite{shi2023pipemoe}{,} ScheMoE\cite{shi2024schemoe}{,} Punniyamurthy \textit{et al.}\cite{punniyamurthy2023optimizing}
            ,leaf,text width=35em]
        ]
        [Storage
            [SE-MoE\cite{shen2022se}{,} Pre-gated MoE\cite{hwang2023pre}{,} EdgeMoE\cite{yi2023edgemoe}{,} MPipeMoE\cite{zhang2024mpmoe}
            ,leaf,text width=35em]
        ]
    ]
    [Application
        [NLP
            [Shazeer \textit{et al.}\cite{shazeer2017outrageously}{,} GShard\cite{lepikhin2020gshard}{,} Swith Transformer\cite{fedus2022switch}{,} Meta-MoE\cite{artetxe2021efficient}{,} GLaM\cite{du2022glam}{,} Mixtral-8x7B\cite{jiang2024mixtral}\\ NLLB\cite{costa2022no}{,}
            DeepSpeed-MoE\cite{rajbhandari2022deepspeed}{,} ST-MoE\cite{zoph2022st}{,} DeepSeekMoE\cite{dai2024deepseekmoe}{,} MoGU\cite{du2024mogu}
            ,leaf,text width=35em]
        ]
        [CV
            [V-MoE\cite{riquelme2021scaling}{,} Swin-MoE\cite{hwang2023tutel}{,} ADVMOE\cite{zhang2023robust}{,} pMoE\cite{chowdhury2023patch}
            ,leaf,text width=35em]
        ]
        [RecSys
            [MMoE\cite{ma2018modeling}{,} PLE\cite{tang2020progressive}{,} AdaMCT\cite{jiang2023adamct}{,} M$^3$oE\cite{zhang2024m3oe}
            ,leaf,text width=35em]
        ]
        [MultiModal
            [LIMoE\cite{mustafa2022multimodal}{,} Shen \textit{et al.}\cite{shen2023scaling}{,} PaCE\cite{li2023pace}{,} MoCLE\cite{gou2023mixture}{,} MoE-LLaVA\cite{lin2024moe}{,} Uni-MoE\cite{li2024uni}{,} MM1\cite{mckinzie2024mm1}\\ LLaVA-MoLE\cite{chen2024llava}
            ,leaf,text width=35em]
        ]
    ]
  ]
\end{forest}
\caption{Taxonomy of Mixture of Experts (MoE).}
\label{fig:taxonomy}
\end{figure*}

For example, the Mixtral 8x7B \cite{jiang2024mixtral}, introduced by Mixtral AI, shares its foundational architecture with the earlier Mistral 7B \cite{jiang2023mistral}, but with a notable difference: each layer comprises eight feed-forward networks (FFN) (i.e., experts). Despite utilizing only 13 billion active parameters, the Mixtral-8x7B demonstrates superior or equivalent performance to the Llama-2-70B \cite{touvron2023llama} and GPT-3.5 \cite{gpt-3.5-turbo} across various benchmarks.
Similarly, the DeepSeek LLM \cite{bi2024deepseek}, developed by DeepSeek, has been extended with an MoE variant known as DeepSeekMoE \cite{dai2024deepseekmoe}. The DeepSeekMoE 16B, while requiring approximately 40\% less computation, attains performance on par with the Llama 2 7B \cite{touvron2023llama}.
The Qwen team has also contributed to this innovative field by developing the Qwen1.5-MoE \cite{qwen_moe}, a smaller MoE model with only 2.7B active parameters that rivals the performance of leading 7B parameter models such as the Mistral 7B \cite{jiang2023mistral} and the Qwen1.5-7B \cite{qwen1.5}.

To assist researchers in navigating the rapidly evolving landscape of LLMs equipped with MoE architectures, we have developed a taxonomy that categorizes these models from three perspectives: algorithm design, system design, and application. 
Figure \ref{fig:taxonomy} showcases our taxonomy alongside several representative studies. 
In the following sections, we provide a comprehensive and in-depth analysis of each category within our taxonomy.

%% file: section/4-overview.tex
\section{Algorithm Design of Mixture of Experts}\label{sec:algorithm}

\subsection{Gating Function}\label{sec:gating_function}
The gating function (also known as the routing function or router), which stands as a fundamental component of all the MoE architectures, orchestrates the engagement of expert computations and the combination of their respective outputs. 
We categorize this mechanism into three distinct types 
Based on the processing methodologies of each input, we categorize the gating mechanism into three distinct types: sparse, which activates a subset of experts; dense, which activates all experts; and soft, which encompasses fully-differentiable approaches including input token merging and expert merging. 


\subsubsection{Sparse}
The sparse gating functions activate a selected subset of experts for processing each individual input token, which can be considered as a form of conditional computation \cite{bengio2013estimating, davis2013low,almahairi2016dynamic}.
The gating functions have been studied extensively, which may be trained by various forms of reinforcement learning and back-propagation, making binary or sparse and continuous, stochastic or deterministic gating decisions \cite{eigen2013learning, bengio2015conditional, rosenbaum2017routing, rosenbaum2019routing, clark2022unified}.
Shazeer \textit{et al.} \cite{shazeer2017outrageously} pioneered a differentiable heuristic with auxiliary load balancing losses, in which the outputs from expert computations are weighted by their selection probabilities. 
This introduces a differentiable aspect to the gating process, thereby facilitating the derivation of gradients that can guide the gating function's optimization.
This paradigm has subsequently become predominant in the realm of MoE research.
Due to its selection of experts for each input token, this method can be recognized as a gating function with token choice.

\begin{figure*}
    \centering
    \includegraphics[width=1\linewidth]{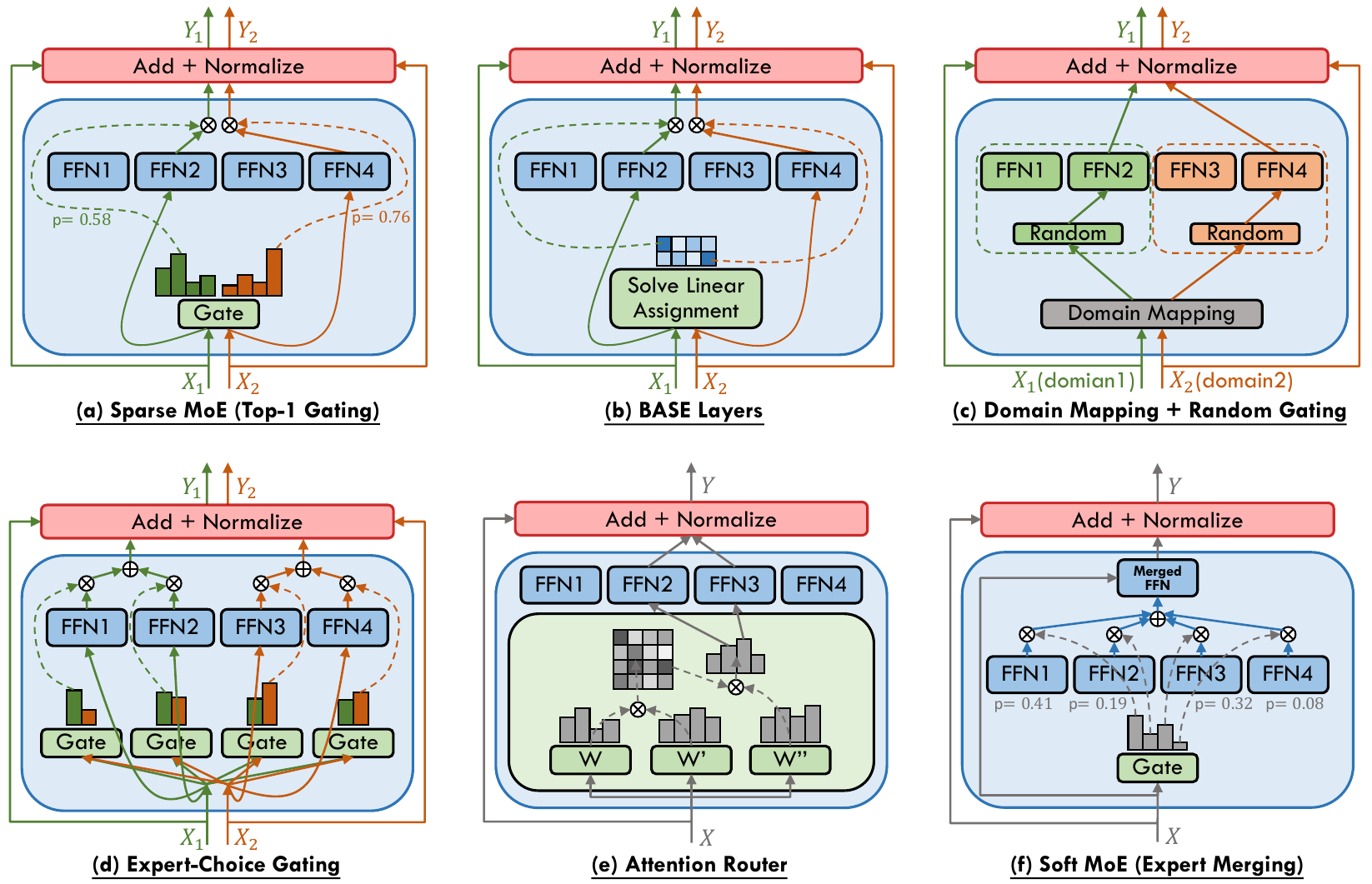}
    \caption{The illustration of various gating functions employed in MoE models, including (a) sparse MoE with top-1 gating \cite{fedus2022switch}, (b) BASE layers \cite{lewis2021base}, (c) the combination of grouped domain mapping and random gating \cite{ren2023pangu}, (d) expert-choice gating \cite{zhou2022mixture}, (e) attention router \cite{wu2024yuan}, and (f) soft MoE with expert merging \cite{muqeeth2023soft}.}
    \label{fig:gating_function}
\end{figure*}

\textbf{Token-Choice Gating.}
Shazeer \textit{et al.} \cite{shazeer2017outrageously} posited the necessity of gating inputs to the top-$k$ experts, with $k>1$, to enhance the efficacy of MoE. 
The rationale behind this approach is that by simultaneously consulting multiple experts for a given input, the network can effectively weigh and integrate their respective contributions, thereby improving performance. 
To accommodate the scalability to thousands of experts within a MoE layer, they employ a two-level hierarchical MoE to reduce the branching factor in the context of a large expert count.
Subsequent research has largely affirmed that increasing the value of k enhances performance, which has led to the widespread adoption of this top-$k$ strategy with $k>1$. Notwithstanding, the Switch Transformer model \cite{fedus2022switch} has shown that a top-1 gating strategy (as illustrated in Figure~\ref{fig:gating_function} (a)) can also yield competitive results, a finding that has been substantiated and adopted by later studies \cite{clark2022unified}. 
Furthermore, M6-t \cite{yang2021m6} proposed a novel variation of the top-1 gating called expert prototyping, which organizes experts into k groups and then applies top-1 gating in each group.
Their experimental results show the training and downstream perplexity of a 16-layer model in order of best to worst: expert prototyping with 4 top-1 gating, 1 top-4 gating, 1 top-16 gating, 1 top-1 gating.

\begin{table*}[t]
\caption{Overview of diverse auxiliary loss functions and their typical coefficient configurations. The originators introducing each auxiliary loss is highlighted as \textbf{bolded reference}, followed by references that adopts the same approach. Studies that have modified the original formulation are indicated with \underline{underlined reference}.}
\label{tab:auxiliary_loss}
\vskip -0.1in
\resizebox{1\textwidth}{!}{
\renewcommand\arraystretch{1.5}
\begin{tabular}{l|c|c} 
\toprule
\textbf{Reference} & \textbf{Auxiliary Loss} & \textbf{Coefficient} \\
\midrule
\textbf{Shazeer \textit{et al.}}\cite{shazeer2017outrageously}, \underline{V-MoE}\cite{riquelme2021scaling} & $L_{importance}+L_{load}$ & $w_{importance}=0.1$, $w_{load}=0.1$\\ 
\makecell[l]{\textbf{GShard}\cite{lepikhin2020gshard}, \textbf{Switch-T}\cite{fedus2022switch}, GLaM\cite{du2022glam}, Mixtral-8x7B\cite{jiang2024mixtral}, DBRX\cite{dbrx}, \\Jamba\cite{lieber2024jamba}, \underline{DeepSeekMoE}\cite{dai2024deepseekmoe}, \underline{DeepSeek-V2}\cite{deepseekv2}, \underline{Skywork-MoE}\cite{wei2024skywork}}
 & $L_{aux}$ & $w_{aux}=0.01$\\ 
\textbf{ST-MoE}\cite{zoph2022st}, OpenMoE\cite{xue2024openmoe}, MoA\cite{zhang2022mixture}, JetMoE \cite{shen2024jetmoe} & $L_{aux}+L_{z}$ & $w_{aux}=0.01$, $w_{z}=0.001$\\
\textbf{Mod-Squad}\cite{chen2023mod}, \underline{Moduleformer}\cite{shen2023moduleformer}, \underline{DS-MoE}\cite{pan2024dense}& $L_{MI}$ & $w_{MI}=0.001$\\ 
\bottomrule
\end{tabular}
}
\end{table*}

\textbf{Auxiliary Loss for Token-Choice Gating.} 
Token-choice gating algorithms frequently incorporate an auxiliary loss during training to promote equitable token distribution across experts. Table~\ref{tab:auxiliary_loss} shows prevalent auxiliary loss functions leveraged in the field.
\textcolor{black}{
Shazeer \textit{et al.} \cite{shazeer2017outrageously} quantify the importance of an expert in relation to a training batch via the batchwise sum of the gate values for that expert, defined as 
\begin{align}
&\operatorname{Importance}(\mathcal{B}) = \sum_{x \in \mathcal{B}} \mathcal{G}(\mathbf{x}; \mathbf{\Theta}).\label{eq:importance}
\end{align}
Furthermore, they introduce an additional loss $\mathcal{L}_{\mathrm{importance}}$, which is defined as the square of the coefficient of variation of the set of importance values, and can be formulated as
\begin{align}
&\mathcal{L}_{\mathrm{importance}} = \operatorname{CV}(\operatorname{Importance}(\mathcal{B}))^2.\label{eq:loss_importance}
\end{align}
This loss is multiplied by a manually adjusted scaling factor $w_{\mathrm{importance}}$, and then integrated into the overall loss function for the model, encouraging all experts to have equal importance.
Although $\mathcal{L}_{\mathrm{importance}}$ promotes balance in importance, it does not guarantee an even distribution of training examples among experts, which can lead to execution inefficiencies in distributed computing environments. To address this, they introduce a second loss $\mathcal{L}_{\mathrm{load}}$, which employs a smooth estimator of the number of examples assigned to each expert for a batch of inputs, thereby facilitating gradient backpropagation.
Simplifying the above design, GShard \cite{lepikhin2020gshard} defines a new differentiable auxiliary loss $\mathcal{L}_{\mathrm{aux}}$ using a differentiable approximation (the dot-product of mean gates and mean gating decisions per expert). This is equivalent to $\mathcal{L}_{\mathrm{load-balancing}}$ in Equations \eqref{eq:load_balance}--\eqref{eq:prob_sum}, as detailed in Section~\ref{sec:background_sparsemoe}. 
Switch Transformers \cite{fedus2022switch} and many other subsequent studies \cite{du2022glam,jiang2024mixtral,dbrx,lieber2024jamba} have embraced this $\mathcal{L}_{\mathrm{aux}}$ design, and enhancements \cite{dai2024deepseekmoe, deepseekv2, wei2024skywork} have been made to cater to diverse requirements.
Nevertheless, ST-MoE \cite{zoph2022st} identified limitations with $\mathcal{L}_{\mathrm{aux}}$, particularly at larger scales, leading to unreliable training outcomes. 
To enhance training stability without compromising quality, the $z$-loss $\mathcal{L}_{z}$ is integrated, defined as
\begin{align}
&\mathcal{L}_{z} = \frac{1}{T}\sum_{x \in \mathcal{B}} (\log \sum_{i=1}^{N} e^{x_i} )^2,\label{eq:loss_z}
\end{align}
which functions by penalizing large logits entering the gating network. 
Since this loss encourages absolute magnitude of values to be smaller, roundoff errors are reduced, which can be quite impactful for exponential functions such as gating.
Additionally, Mod-Squad \cite{chen2023mod} posits the difficulty of training multi-task models under such an expert-balancing loss, which may inadvertently force experts to set parameters on conflicting tasks or hinder the potential synergies from parameter sharing across complementary tasks.
Instead, it proposes to maximize the mutual information (MI) between experts and tasks to build task-expert alignment, which is accomplished through $\mathcal{L}_{\mathrm{MI}}$. 
Differently, ModuleFormer \cite{shen2023moduleformer} proposes to maximize the Mutual Information between experts and tokens.
Furthermore, DS-MoE \cite{pan2024dense} extends the application of $\mathcal{L}_{\mathrm{MI}}$, calibrating different weightings $w_{\mathrm{MI}}$, in Mixture-of-Attention (MoA, as illustrated in Figure~\ref{fig:moa_share} (a)) and FFN MoE modules of different size models.
Although existing research has introduced many different gating functions and auxiliary losses, the methods proposed by GShard \cite{lepikhin2020gshard} remain the predominant choice in industry \cite{du2022glam,jiang2024mixtral,dbrx,dai2024deepseekmoe,lieber2024jamba}.}

\textbf{Expert Capacity for Token-Choice Gating.} 
In conjunction with load balancing via auxiliary loss, GShard \cite{lepikhin2020gshard} incorporates an expert capacity limit, defining a threshold for the number of tokens an expert can process.
This can lead to token overflow, where excess tokens are not processed by the designated expert.
GShard also proposes a random routing mechanism that selects a secondary expert with a probability proportional to its weight, under the intuition that the contribution of a secondary expert can be negligible, given that the output is a weighted average and the secondary weight is typically small.
For the task of image classification with Vision Transformer (ViT) models, Riquelme \textit{et al.} \cite{riquelme2021scaling} enhance the top-$k$ gating strategy with Batch Prioritized Routing (BPR), which assigns priority based on higher gating scores rather than the sequence order of tokens. 
Zoph \textit{et al.} \cite{zoph2022st} have demonstrated the efficacy of BPR in the context of MoE language models. 
Kim \textit{et al.} \cite{kim2021scalable} suggest randomizing token prioritization within sequences to mitigate routing bias towards early-positioned tokens.
OpenMoE \cite{xue2024openmoe} provides a comprehensive analysis of gating mechanisms, highlighting the ``Drop-towards-the-End'' phenomenon whereby tokens later in a sequence are at greater risk of being dropped due to experts reaching their maximum capacity limits, an issue that is exacerbated in instruction-tuning datasets. 
Moreover, OpenMoE identifies a tendency within MoE systems to route tokens based on token-level semantic similarities, leading to ``Context-independent Specialization''. 
Additionally, this token ID routing specialization is established early in pretraining and remains largely fixed, resulting in a consistent pattern of token processing by the same experts throughout training, a phenomenon referred to as ``Early Routing Learning''.

\textbf{Other Advancements on Token-Choice Gating.} 
Despite the implementation of gating heuristics and auxiliary expert-balancing loss functions aimed at achieving a balanced workload distribution among experts, the issue of load imbalance persists as a prevalent challenge within MoE architectures.
To solve it, the Balanced Assignment of Sparse Experts (BASE) layer, as conceptualized by Lewis \textit{et al.} \cite{lewis2021base} and illustrated in Figure~\ref{fig:gating_function} (b), re-envisions the token-to-expert allocation process by casting it as a linear assignment problem, aiming to maximize the token-expert affinities under the constraints that each expert is assigned an equal quantity of tokens.
Subsequently, Clark \textit{et al.} \cite{clark2022unified} introduce a variant of the BASE layer, termed S-BASE, using an optimal transport formulation. 
Additionally, they devise a reinforcement learning based gating algorithm employing top-1 routing, with the reward function defined as the negative cross-entropy of the predicted token.
The discrete optimization of gating function can lead to convergence and statistical performance issues when training with gradient-based methods. To address these issues, Hazimeh \textit{et al.} \cite{hazimeh2021dselect} introduce DSelect-k, which is a smooth version of the top-$k$ gating algorithm that improves over standard top-$k$ gating.
This method constitutes a refined version of the top-$k$ gating algorithm, featuring enhanced smoothness properties that yield improvements over the conventional top-$k$ gating approach.
Kudugunta \textit{et al.} \cite{kudugunta2021beyond} diverge from the prevalent token-level gating strategies by introducing a sentence-level gating mechanism. 
This approach involves generating a sentence representation by averaging the tokens within a sequence and subsequently routing it to an expert.
Chi \textit{et al.} \cite{chi2022representation} observe that prevailing gating mechanisms tend to push hidden representations clustering around expert centroids, implying a trend toward representation collapse, which in turn harms model performance.
To counteract this issue, they project hidden vectors into a lower-dimensional space before gating and implement L2 normalization for both token representations and expert embeddings, thus calculating gating scores within a low-dimensional hypersphere. 
Cai \textit{et al.} \cite{caiflextron} identify the difficulty in training the adaptive routers due to gradient vanishing, and introduce a novel strategy involving the training of a Surrogate Model (SM) that predicts LLM’s language loss given only router choices.
Once trained, the SM is frozen, and the router is subsequently tuned to minimize language loss, relying exclusively on feedback from the SM.
Skywork-MoE \cite{wei2024skywork} proposes two innovative techniques: gating logit normalization, which improves expert diversification, and adaptive auxiliary loss coefficients, which provides layer-specific adjustment of auxiliary loss coefficients.
Yuan 2.0-M32 \cite{wu2024yuan} proposes a new router network, Attention Router (as illustrated in Figure~\ref{fig:gating_function} (e)), which implements a more efficient selection of experts and yields an enhancement in model accuracy over classical linear router network.
Zeng \textit{et al.} \cite{zeng2024adamoe} posit that the complexity of token feature abstraction may necessitate a variable number of experts to process. In response, they propose AdaMoE, a novel approach that enables token-adaptive gating for MoE, allowing for a dynamic number of selected experts per token. 
AdaMoE subtly modifies the standard top-$k$ MoE by incorporating a predetermined set of null experts and increasing the value of $k$.
Importantly, AdaMoE does not mandate a uniform allocation of null experts across tokens but ensures the average engagement of null experts with a load-balancing loss, resulting in an adaptive number of null/true experts used by each token.
Dynamic Mixture of Experts (DYNMoE) \cite{guo2024dynamic} also introduces an innovative gating mechanism that enables individual tokens to automatically determine the number of activated experts via the trainable per-expert thresholds, incorporating an adaptive process that automatically add or remove experts during training.
Shi \textit{et al.} \cite{shi2024unchosen} introduce Self-Contrast Mixtureof-Experts (SCMoE), a training-free strategy that utilizes the contrastive information among different gating strategies to engage unchosen experts during inference.

\textbf{Non-trainable Token-Choice Gating.}
The dynamic training of gating functions within MoE models is standard practice; however, some research has ventured into the realm of non-trainable token-choice gating mechanisms.
The most significant benefit of non-trainable token-choice gating is that no additional gating network parameters are required and the full load balancing can be achieved through specific gating mechanisms.
The Hash Layer \cite{roller2021hash} utilizes a random fixed gating approach by hashing the input token, achieving competitive results without the necessity of training the gating network. 
The load balancing is facilitated by the selection of hash functions prior to training, which can equitably distribute token batches.
Zuo \textit{et al.} \cite{zuo2021taming} introduces THOR, an algorithm that randomly allocates two experts to each input during training and inference with a consistency regularized loss promoting consistent predictions.
Gururangan \textit{et al.} \cite{gururangan2022demix} propose the DEMix model, which explicitly assigns distinct experts to discrete pretraining domains, with domain matching being employed to select experts corresponding to the training inputs. 
Given the potential suboptimality of domain categorization and its limited scope in encompassing test-time domains, a single domain expert selection could undermine the model's generalizability. 
To address this, DEMix adopts a parameter-free probabilistic method that dynamically estimates the domain-weighted mixture at inference.
Kudugunta \textit{et al.} \cite{kudugunta2021beyond} explore task-level gating incorporating prior knowledge tags, and similarly, M2M-100 model \cite{fan2021beyond} utilizes explicit language-specific sublayers with deterministically routing input tokens based on their language.
Building upon the aforementioned non-trainable gating strategies—random gating and domain mapping—PanGu-$\sum$ \cite{ren2023pangu} presents the Random Routed Experts (RRE) mechanism. As illustrated in Figure~\ref{fig:gating_function} (c), this approach initially routes tokens to a domain-specific expert group, followed by a random selection within that group.

In contrast to explicit language-specific expert selection, NLLB \cite{costa2022no} leverages trainable gating to manage multilingual machine translation tasks, outperforming the M2M-100 approach \cite{fan2021beyond}. 
Addressing task interference in generalist models, Zhu \textit{et al.} \cite{zhu2022uni} introduce the Conditional MoE, which augments MoE with trainable gating by integrating conditional information at various levels, such as token-level, context-level, modality-level, task-level, and predefined token attributes.
Ye \textit{et al.} \cite{ye2022eliciting} further investigate the incorporation of trainable gating at task-level MoE.
Additionally, STABLEMOE \cite{dai2022stablemoe} identifies a challenge with existing learning-to-route MoE methods: the phenomenon of gating fluctuation. 
To counter this, STABLEMOE employs a two-stage training process. The first stage focuses on acquiring a balanced and cohesive gating strategy, which is then distilled into a lightweight gate function, decoupled from the backbone model. 
Subsequently, the second stage leverages the distilled gate for token-to-expert assignments and freezes it to ensure a stable gating strategy throughout further training.

\textbf{Expert-Choice Gating.}
Zhou \textit{et al.} \cite{zhou2022mixture} propose an inversion of the conventional token-choice gating paradigm, wherein each expert selects the top-$k$ tokens they will process, as illustrated in Figure~\ref{fig:gating_function} (d). 
This approach circumvents the necessity for auxiliary load balancing losses during training, ensuring a uniform distribution of tokens across experts. 
However, this method may result in uneven token coverage, with some tokens potentially being processed by multiple experts or not at all. 
Despite this, the technique demonstrates strong empirical performance and offers an adaptive computational interpretation where the model can implicitly apply more computation to certain tokens. 
The effectiveness of expert-choice gating is further validated by Zhou \textit{et al.} in their subsequent Brainformers study \cite{zhou2023brainformers}. 
Additionally, Komatsuzaki \textit{et al.} \cite{komatsuzaki2022sparse} integrate the expert-choice gating strategy within the Vision Transformer and adapt it for the encoder in T5 models, while maintaining token-choice gating for the T5 decoder.

\subsubsection{Dense}
In Section~\ref{sec:background_densemoe}, we discuss the enduring relevance of dense MoE, which activates all the experts for each input process. 
This dense paradigm continues to inform current innovations in MoE training and inference methodologies, as elaborated in Section~\ref{sec:dense_to_sparse}. 
While sparse activation of experts, as a trade-off, may yield computational efficiency gains at the expense of some performance loss when compared to a densely activated MoE with an equivalent number of total parameters \cite{dai2024deepseekmoe, pan2024dense, shen2023moduleformer}, it represents a strategic adjustment to balance computational demands with model capability.
Notably, dense activation performs well in the context of LoRA-MoE fine-tuning, where the computational overhead of LoRA experts is comparatively low. This approach enables the effective and flexible integration of multiple LoRAs across a variety of downstream tasks. It preserves the generative capabilities of the original pretrained model and maintains the unique characteristics of individual LoRAs for each task \cite{wu2023mole, dou2023loramoe}.

\subsubsection{Soft}
Deciding the allocation of appropriate experts to each input token pose the fundamental discrete optimization challenge for sparse MoE.
This often necessitates heuristic auxiliary losses to ensure balanced expert engagement and to minimize unassigned tokens. 
These issues become more pronounced in scenarios involving out-of-distribution data, such as small inference batches, novel inputs, or during transfer learning.
Similar to dense MoE, the soft MoE approach maintains full differentiability by leveraging all the experts for processing each input, thus avoiding issues inherent to discrete expert selection. 
We distinguish soft MoE from dense MoE to highlight the characteristic that mitigates computational demands through the gating-weighted merging of input tokens or experts.

\textbf{Token Merging.} 
Puigcerver \textit{et al.} \cite{puigcerver2023sparse} proposed the Soft MoE, which eschews the conventional sparse and discrete gating mechanism in favor of a soft assignment strategy that merges tokens. 
This method computes several weighted averages of all tokens, with weights depending on both tokens and experts, and processes each aggregate with its respective expert. 
Their experimental results in image classification demonstrate that soft MoE enhances the stability of gating function training and inherently maintains balance. 
HOMOE \cite{dat2023homoe} follows the design of Soft MoE and combines it with Hopfield network to address the the challenges of Compositional Zero-Shot Learning tasks.
Yet, merging input tokens complicates its application in auto-regressive decoders, as future tokens required for averaging are inaccessible during inference.

\textbf{Expert Merging.} In contrast to the merging of input tokens, Muqeeth \textit{et al.} \cite{muqeeth2023soft} introduced the Soft Merging of Experts with Adaptive Routing (SMEAR) framework, which circumvents discrete gating by merging all the experts' parameters through a weighted average, as illustrated in Figure~\ref{fig:gating_function} (f). 
They argue that conventional sparse MoE models often fail to match the performance of their parameter-matched dense counterparts or those utilizing non-learned heuristic gating functions, potentially due to flawed gradient estimation methods for training modules with non-differentiable, discrete gating decisions. 
By processing the input tokens through a single merged expert, SMEAR does not incur a significant increase in computational costs and enables standard gradient-based training. 
Empirical evaluations on T5-GLUE and ResNet-DomainNet benchmarks reveal that SMEAR-equipped models surpass those with metadata-based \cite{gururangan2022demix, kudugunta2021beyond} or gradient-estimated learning gating strategies. On ResNet-DomainNet, SMEAR achieved a 1.5\% higher average accuracy than Soft MoE \cite{puigcerver2023sparse} with single ``slot'' per expert, at the expense of a near 10\% reduction in throughput.
Subsequent contributions by Zhong \textit{et al.} \cite{zhong2024lory} argue that SMEAR's demonstrated advantages are confined to downstream fine-tuning on classification tasks. They present Lory, an innovative approach for scaling such expert merging architectures to auto-regressive language model pretraining. 
Lory \cite{zhong2024lory} introduces a causal segment routing strategy, conducting expert merging at the segment level while maintaining the auto-regressive nature of language models. Furthermore, it employs similarity-based data batching to direct expert specialization in particular domains or topics. 
Lory's empirical validation on LLaMA models showcases significant improvements over parameter-matched dense models in terms of perplexity (by 13.9\%) and on diverse downstream tasks (by 1.5\%-11.1\%), highlighting the potential of fully-differentiable MoE architectures for language model pretraining and encouraging further investigation in this area.
In addition, expert merging methods have demonstrated efficacy in parameter-efficient fine-tuning (PEFT) MoE contexts. 
Zadouri \textit{et al.} \cite{zadouri2023pushing} substantiate that soft merging of experts significantly outperforms sparse gating mechanisms (top-1, top-2) in the T5 models \cite{raffel2020exploring} fine-tuning with the MoV-10 setting of 10 (IA)$^3$ vector expert. 
Wu \textit{et al.} \cite{wu2023omni} propose Omni-SMoLA, an architecture leveraging the soft method to mix multimodal low-rank experts, improving the generalist performance across a broad range of generative vision-language tasks.
He \textit{et al.} \cite{he2023merging} introduce Merging Experts into One (MEO), merging multiple selected experts into one to reduce the expert computation cost. 
Moreover, they perform the expert selection at the sequence level and employ a token attention mechanism for capturing the identification of each token, thus preserving context information without the necessity of merging distinct weights and biases for individual tokens.

\subsection{Experts}
In this section, we delineate the architecture of expert networks within MoE framework, following our discussion on the gating function that orchestrates the activation of these experts.

\begin{figure*}
    \centering
    \includegraphics[width=0.9\linewidth]{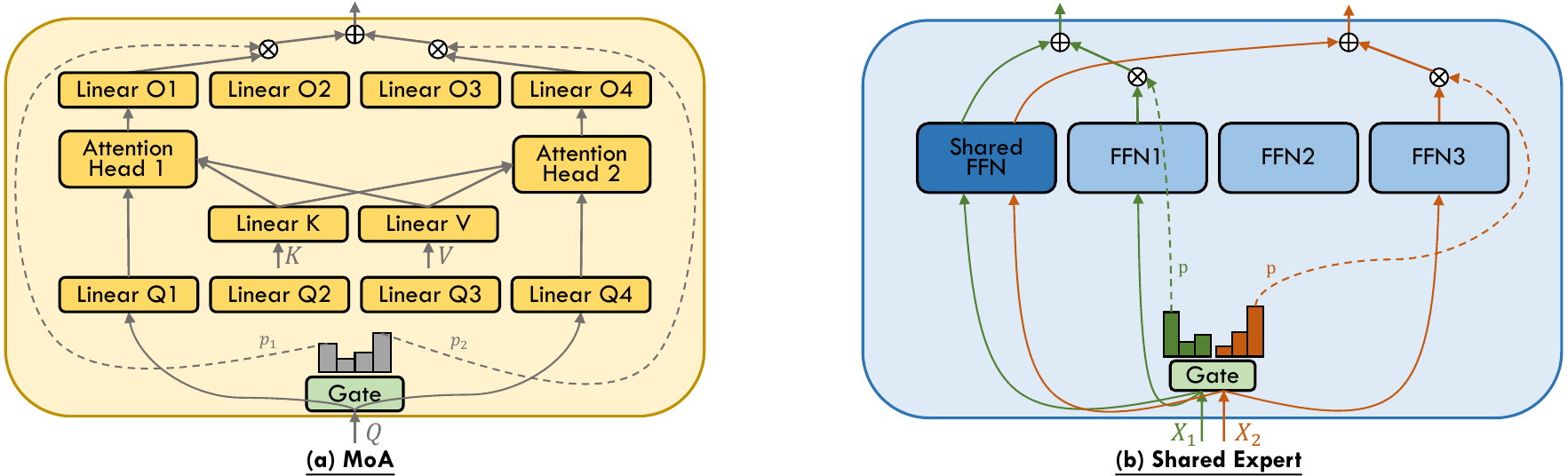}
    \caption{The illustration of Mixture of Attention Heads \cite{zhang2022mixture} (a) and Shared Expert \cite{rajbhandari2022deepspeed} (b) architectures.}
    \label{fig:moa_share}
\end{figure*}

\subsubsection{Network Types}
Since the initial integration of MoE into transformer architectures \cite{lepikhin2020gshard,fedus2022switch,zoph2022st}, MoE has served as a substitute for Feed-Forward Network (FFN) modules within these models. 
Typically, each expert within a MoE layer replicates the architecture of the FFN it replaces.
This paradigm, wherein FFNs are utilized as experts, remains predominant, and subsequent refinements will be expounded upon in Sections~\ref{sec:hyperparameters} to \ref{sec:share_expert}.

\textbf{Feed-Forward Network.} 
As discussed in existing work \cite{sukhbaatar2024branch}, the predilection for leveraging MoE in the context of FFNs is rooted in the hypothesis that self-attention layers exhibit lower sparsity and less domain specificity than FFN layers. 
Pan \textit{et al.} \cite{pan2024dense} provide empirical support for this, revealing marked sparsity in FFN layers compared to self-attention layers, through their analysis of downstream Wikitext tasks using their pretrained DS-MoE models. 
Their results indicate a mere 20\% active expert engagement in FFN layers, in contrast to the 80\% observed within self-attention layers.
In earlier investigation of FFN computational patterns, Zhang \textit{et al.} \cite{zhang2022moefication} and Li \textit{et al.} \cite{li2022lazy} observe that most inputs only activate a small proportion of neurons of FFNs, highlighting the inherent sparsity of FFNs.
Subsequently, Zhang \textit{et al.} \cite{zhang2023emergent} observe the Emergent Modularity (EM) phenomenon within pretrained Transformers, revealing a significant correlation between neuron activation and specific tasks (evidenced by the functional specialization of neurons and function-based neuron grouping). This discovery supports the proposition that the MoE structure reflects the modularity of pre-trained Transformers.

\textbf{Attention.} While the focus of MoE research has predominantly been on FFN layers within the Transformer architecture, Zhang \textit{et al.} \cite{zhang2022mixture} introduce the Mixture of Attention Heads (MoA), an innovative architecture that combines multi-head attention layers with MoE to further enhance performance and restrain computational cost. 
As delineated in Figure~\ref{fig:moa_share} (a), MoA employs two sets of experts, one for query projection and one for output projection. Both sets select the experts with the same indices through a common gating network.
To reduce computational complexity, MoA shares the key ($W_k$) and value ($W_v$) projection weights across attention experts, with experts differentiated only by their respective query ($q_tW_i^q$) and output ($o_{i,t}W_i^o$) projection weights, allowing for shared pre-computation of key ($KW_k$) and value ($VW_v$) sequences. 
Subsequent work such as DS-MoE \cite{pan2024dense}, JetMoE \cite{shen2024jetmoe}, and ModuleFormer \cite{shen2023moduleformer} follows the design of MoA and further refines the combination of MoE and attention layer.

\textbf{Others.} 
In addition to the aforementioned expert network types, researchers have explored the use of Convolutional Neural Network (CNN) as expert \cite{chowdhury2023patch, zhang2023robust,chen2022towards,wang2020deep,gross2017hard}. 
Moreover, recent endeavors that integrate Parameter-Efficient Fine-Tuning (PEFT) techniques with MoE, such as employing Low-Rank Adaptation (LoRA) \cite{hu2021lora} as expert, have shown promising results, which are discussed in Section~\ref{sec:mope}.

\begin{table*}[t]
\caption{Comparative configurations of MoE with FFN experts in selected models. Model differentiation in each reference is achieved by using the model size, indicated either by total or activated/total parameter count. Both activated and total expert counts encompass the count of shared experts when utilized. $d_{model}$ is the hidden size, $d_{ffn}$ is the intermediate size of FFNs, $d_{expert}$ is the intermediate size of FFN experts, \#L is the number of layers, \#H and $d_{head}$ are the number of attention heads and attention head dimensions.}
\label{tab:expert}
\vskip -0.1in
\resizebox{1\textwidth}{!}{ 
\renewcommand\arraystretch{1.2}
\begin{tabular}{cccccccccccc} 
\toprule
\textbf{Reference} & \textbf{Models} & \makecell[c]{\textbf{Expert Count} \\ (Activ./Total)} & \textbf{$d_{model}$} & \textbf{$d_{ffn}$} & \textbf{$d_{expert}$ }& \textbf{\#L} & \textbf{\#H} & \textbf{$d_{head}$} & \makecell[c]{\textbf{Placement} \\\textbf{Frequency}} & \makecell[c]{\textbf{Activation} \\\textbf{Function}} & \makecell[c]{\textbf{Share Expert} \\\textbf{Count}}\\
\midrule
\multirow{4}{*}{\shortstack{GShard \cite{lepikhin2020gshard}\\(2020)}} & 600B & 2/2048 & 1024& 8192& $d_{ffn}$ & 36 & 16& 128 & 1/2 & ReLU & 0\\ 
& 200B & 2/2048 & 1024& 8192& $d_{ffn}$ & 12 & 16& 128 & 1/2 & ReLU & 0\\ 
& 150B & 2/512 & 1024& 8192& $d_{ffn}$ & 36 & 16& 128 & 1/2 & ReLU & 0\\ 
& 37B & 2/128 & 1024& 8192& $d_{ffn}$ & 36 & 16& 128 & 1/2 & ReLU & 0\\ 
\hline
\multirow{4}{*}{\shortstack{Switch \cite{fedus2022switch}\\(2021)}} & 7B & 1/128 & 768& 2048& $d_{ffn}$& 12 & 12& 64 & 1/2 & GEGLU & 0\\
& 26B & 1/128 & 1024& 2816& $d_{ffn}$ & 24 & 16& 64 & 1/2 & GEGLU & 0\\
& 395B & 1/64 & 4096& 10240& $d_{ffn}$ & 24 & 64& 64 & 1/2 & GEGLU & 0\\
& 1571B & 1/2048 & 2080& 6144& $d_{ffn}$ & 15 & 32& 64 & 1 & ReLU & 0\\
\hline
\multirow{4}{*}{\shortstack{GLaM \cite{du2022glam}\\(2021)}} & 0.1B/1.9B & 2/64 & 768& 3072& $d_{ffn}$ & 12 & 12& 64 & 1/2 & GEGLU & 0\\ 
& 1.7B/27B & 2/64 & 2048& 8192& $d_{ffn}$ & 24 & 16& 128 & 1/2 & GEGLU & 0\\ 
& 8B/143B & 2/64 & 4096& 16384& $d_{ffn}$ & 32 & 32& 128 & 1/2 & GEGLU & 0\\ 
& 64B/1.2T & 2/64 & 8192& 32768& $d_{ffn}$ & 64 & 128& 128 & 1/2 & GEGLU & 0\\ 
\hline
\multirow{4}{*}{\shortstack{DeepSpeed-MoE \cite{rajbhandari2022deepspeed}\\(2022)}} & 350M/13B & 2/128 & 1024& $4d_{model}$& $d_{ffn}$ & 24 & 16& 64 & 1/2 & GeLU & 0 \\ 
& 1.3B/52B & 2/128 & 2048& $4d_{model}$& $d_{ffn}$ & 24 & 16& 128 & 1/2 & GeLU & 0 \\ 
& \makecell[c]{PR-350M/4B} & 2/32-2/64 & 1024& $4d_{model}$& $d_{ffn}$ & 24 & 16& 64 & \makecell[c]{1/2, 10L-32E, 2L-64E} & GeLU & 1 \\ 
& \makecell[c]{PR-1.3B/31B} & 2/64-2/128 & 2048& $4d_{model}$& $d_{ffn}$ & 24 & 16& 128 & \makecell[c]{1/2, 10L-64E, 2L-128E} & GeLU & 1 \\ 
\hline
\multirow{2}{*}{\shortstack{ST-MoE \cite{zoph2022st}\\(2022)}} & 0.8B/4.1B & 2/32 & 1024& 2816& $d_{ffn}$ & 27 & 16& 64 & \makecell[c]{1/4, add extra FFN} & GEGLU & 0\\
& 32B/269B & 2/64 & 5120& 20480& $d_{ffn}$ & 27 & 64& 128 & \makecell[c]{1/4, add extra FFN} & GEGLU & 0\\
\hline
\multirow{2}{*}{\shortstack{Mixtral \cite{jiang2024mixtral} \\(2023)}} & 13B/47B & 2/8 & 4096& 14336 & $d_{ffn}$ & 32 & 32& 128 & 1 & SwiGLU & 0 \\
& 39B/141B & 2/8 & 6144 & 16384 & $d_{ffn}$ & 56 & 48 & 128 & 1 & SwiGLU & 0 \\
\hline
\multirow{3}{*}{\shortstack{LLAMA-MoE \cite{llama-moe-2023} \\(2023)}} & 3.0B/6.7B & 2/16 & 4096 & 11008 & 688 & 32 & 32& 128 & 1 & SwiGLU & 0 \\
& 3.5B/6.7B & 4/16 & 4096 & 11008 & 688 & 32 & 32& 128 & 1 & SwiGLU & 0 \\
& 3.5B/6.7B & 2/8 & 4096 & 11008 & 1376 & 32 & 32 & 128 & 1 & SwiGLU & 0 \\
\hline
\multirow{3}{*}{\shortstack{DeepSeekMoE \cite{dai2024deepseekmoe}\\(2024)}} & 0.24B/1.89B & 8/64 & 1280& -& $\frac{1}{4}d_{ffn}$ & 9 & 10& 128 & 1 & SwiGLU & 1 \\
& 2.8B/16.4B & 8/66 & 2048& 10944& 1408 & 28 & 16& 128 & \makecell[c]{1, except 1st layer} & SwiGLU & 2 \\
& 22B/145B & 16/132 & 4096& -& $\frac{1}{8}d_{ffn}$ & 62 & 32& 128 & \makecell[c]{1, except 1st layer}  & SwiGLU & 4 \\
\hline
\multirow{3}{*}{\shortstack{OpenMoE \cite{xue2024openmoe}\\(2024)}} & 339M/650M & 2/16 & 768& 3072& $d_{ffn}$ & 12 & 12& 64 & 1/4 & SwiGLU & 1\\
& 2.6B/8.7B & 2/32 & 2048& 8192& $d_{ffn}$ & 24 & 24& 128 & 1/6 & SwiGLU & 1\\
& 6.8B/34B & 2/32 & 3072& 12288& $d_{ffn}$ & 32 & 24& 128 & 1/4 & SwiGLU & 1\\
\hline
\makecell[c]{Qwen1.5-MoE \cite{qwen_moe} \\(2024)} & 2.7B/14.3B & 8/64 & 2048& 5632& 1408 & 24 & 16& 128 & 1 & SwiGLU & 4 \\
\hline
\makecell[c]{DBRX \cite{dbrx} \\(2024)} & 36B/132B & 4/16 & 6144& 10752 & $d_{ffn}$ & 40 & 48& 128 & 1 & SwiGLU & 0 \\
\hline
\makecell[c]{Jamba \cite{lieber2024jamba} \\(2024)} & 12B/52B & 2/16 & 4096 & 14336 & $d_{ffn}$ & 32 & 32 & 128 & \makecell[c]{1/2, \\ 1:7 Attention:Mamba} & SwiGLU & 0 \\
\hline
\makecell[c]{Skywork-MoE \cite{wei2024skywork} \\(2024)} & 22B/146B & 2/16 & 4608& 12288& $d_{ffn}$ & 52 & 36& 128 & 1 & SwiGLU & 0 \\
\hline
\makecell[c]{Yuan 2.0-M32 \cite{wu2024yuan} \\(2024)} & 3.7B/40B & 2/32 & 2048 & 8192 & $d_{ffn}$ & 24 & 16 & 256 & 1 & SwiGLU & 0 \\
\hline
\makecell[c]{OLMoE \cite{muennighoff2024olmoe} \\(2024)} & 1.3B/6.9B & 8/64 & 2048 & 1024 & $d_{ffn}$ & 16 & 16 & 128 & 1 & SwiGLU & 0 \\
\hline
\makecell[c]{DeepSeek-V3 \cite{liu2024deepseek} \\(2024)} & 37B/671B & 9/257 & 7168 & 18432 & 2048 & 61 & 128 & 128 & 1, except first 3 layers & SwiGLU & 1 \\
\bottomrule
\end{tabular}
}
\end{table*}

\subsubsection{Hyperparameters} 
\label{sec:hyperparameters} 
The scale of sparse MoE models is governed by several critical hyperparameters that extend beyond those of dense transformer models. 
These include (1) the count of experts per MoE layer, (2) the size of each expert, and (3) the placement frequency of MoE layers throughout the model. 
The selection of these hyperparameters is crucial, as it profoundly influences model performance and computational efficiency across various tasks. 
Optimal hyperparameter choices are thus contingent upon the specific application requirements and the constraints of the computational infrastructure. 
Our subsequent analysis, informed by the exemplified models listed in Table~\ref{tab:expert}, explores these hyperparameter decisions in depth. 
Meanwhile, we enumerate some recent open-source models, summarizing their number of parameters and benchmark performance in Table~\ref{tab:open_models}.

\textbf{Expert Count.} 
Initial investigations employing thousands of experts per layer yielded impressive gains in pretraining and translation quality \cite{shazeer2017outrageously,lepikhin2020gshard,fedus2022switch}. 
Nonetheless, the quality of sparse MoE models is disproportionately reduced under domain shift \cite{artetxe2021efficient} or when fine-tuning on diverse task distributions \cite{fedus2022switch}. 
GLaM \cite{du2022glam} adopts a configuration of 64 experts, guided by their findings that a 64-expert setup with top-2 gating strikes an optimal balance between execution efficiency and performance across zero-shot, one-shot, and few-shot scenarios. 
Reflecting this trend, more recent sparse MoE models \cite{zoph2022st,jiang2024mixtral,xue2024openmoe,qwen_moe,dbrx,wei2024skywork,lieber2024jamba,wu2024yuan} commonly utilize no more than 64 experts.
Additionally, DeepSpeed-MoE \cite{rajbhandari2022deepspeed} adopts a Pyramid-MoE approach, positioning MoE layers with a larger expert count towards the network's end.

\textbf{Expert Size.} 
To scale the model effectively, GLaM \cite{du2022glam} prioritizes the expansion of the 
intermediate hidden dimension per expert while standardizing the expert count at 64, a strategy that often requires the implementation of tensor parallelism across multiple accelerators to maintain computational efficiency \cite{du2022glam, rajbhandari2022deepspeed, fedus2022switch}.
From this period forward, MoE models \cite{zoph2022st,jiang2024mixtral,dbrx,wei2024skywork} typically featured larger expert dimensions.
Differently, DeepSeekMoE \cite{dai2024deepseekmoe, deepseekv2} introduces the concept of fine-grained expert segmentation by subdividing the intermediate hidden dimension of FFN expert, while preserving the overall parameter count. 
Specifically, DeepSeekMoE-145B employs a reduced intermediate hidden dimension at one-eighth that of its dense FFN counterpart, increasing both the number of experts (from 16 to 128) and the number of active experts (from top-2 to top-16) by a factor of eight. 
They believe that this strategy not only refines the decomposition of knowledge across experts, facilitating more precise learning, but also enhances the flexibility of expert activation combinations, allowing for more specialized and targeted knowledge capture.
Qwen1.5-MoE \cite{qwen_moe} and DBRX \cite{dbrx} adopt a similar fine-grained expert segmentation strategy. 
Results from LLAMA-MoE \cite{llama-moe-2023}, which allocates dense FFN parameters across non-overlapping experts to maintain a consistent parameter count, indicate that activating 4 out of 16 experts with a dimensionality of $d_{expert}=688$ marginally outperforms the activation of 2 out of 8 experts with $d_{expert}=1376$.
Furthermore, Parameter Efficient Expert Retrieval (PEER) \cite{he2024mixture}, an innovative layer design employing the product key technique \cite{lample2019large} for sparse retrieval from a vast pool of tiny experts (over a million single-neuron experts), surpasses dense FFNs and coarse-grained MoEs in terms of performance-compute trade-off on language modeling tasks.

\textbf{Frequency of MoE Layers.} 
Sparse MoE models typically evolve from dense architectures by interspersing MoE layers in place of the dense FFN layers at regular intervals. 
Although a higher frequency of MoE layers can enlarge the model size, it also introduces greater system overhead. 
In practice, most MoE models features alternate FFN replacement (1/2) with MoE layers \cite{lepikhin2020gshard,du2022glam,artetxe2021efficient,rajbhandari2022deepspeed}. Nevertheless, variations exist, with some models incorporating MoE layers every fourth layer (1/4) \cite{zoph2022st,xue2024openmoe} or in every layer (1/1) \cite{fedus2022switch,dai2024deepseekmoe}.
Following the introduction of Mixtral 8x7B \cite{jiang2024mixtral}, the trend seems to shift towards placing MoE in every layer of the model, with a common choice of only 8 or 16 experts mirroring the dimensionality of a dense FFN \cite{dai2024deepseekmoe,qwen_moe,dbrx,wei2024skywork}.

Research into the optimal configuration of MoE layers has been extensive. 
V-MoE \cite{riquelme2021scaling} employs MoE in the last few even-numbered Transformer layers, noting that, despite using fewer MoE layers, the impact on performance is minimal while computational speed is significantly enhanced. 
DeepSeekMoE-16B/-145B \cite{dai2024deepseekmoe} replaces all FFNs with MoE layers, excluding the first, due to the observed slower convergence of load balance status in the first layer.
MoE-LLaVA \cite{lin2024moe}, a recently popular open Large Vision-Language Model (LVLM), demonstrates that alternating MoE placement yields superior model quality and execution efficiency on multimodal tasks, compared to every-layer MoE placement or "First-Half" and "Second-Half" configurations.
ST-MoE \cite{zoph2022st} found that adding a dense FFN adjacent to each MoE layer can improve model quality. 
Brainformers \cite{zhou2023brainformers} introduce a nonuniform architecture that integrates MoE layers, dense FFNs, attention mechanisms, and a variety of layer normalizations and activation functions without strict sequential layering, trading architectural regularity for the flexibility of sub-layer composition.
Jamba \cite{lieber2024jamba} integrates the architecture of Mamba \cite{gu2023mamba} by adopting a 1:7 ratio of attention-to-Mamba layers.

\begin{table*}[t]
\caption{A collection of recent open-source models detailing activated and total parameter counts, alongside performance benchmarks such as MMLU \cite{hendrycks2020measuring} (5-shot), GSM8K \cite{cobbe2021training} (5-shot), MATH \cite{hendrycks2021measuring} (4-shot), and HumanEval \cite{chen2021evaluating} (0-shot), unless specified otherwise.}
\vskip -0.1in
\label{tab:open_models}
\resizebox{1\textwidth}{!}{
\renewcommand\arraystretch{1.3}
\begin{tabular}{l|c|c|cc|cccc|c} 
\toprule
\multirow{2}{*}{\textbf{Name}} & \multirow{2}{*}{\textbf{Time}} & \multirow{2}{*}{\textbf{Affiliation}} & \multicolumn{2}{c|}{\textbf{Params.}} & \multicolumn{4}{c|}{\textbf{Benchmarks}} & \multirow{2}{*}{\textbf{Link}}\\ 
 \cline{4-9} & & & Activ. & Total & MMLU & GSM8K & MATH & HumanEval &\\ 
\midrule
Mixtral-8x7B-v0.1 & 2023.12 & Mistral & 13B & 47B & 70.6 & 58.4, 74.4 (8-shot) & 28.4 & 40.2 & \href{https://huggingface.co/mistralai/Mixtral-8x7B-v0.1}{https://huggingface.co/mistralai/Mixtral-8x7B-v0.1} \\
DeepSeekMoE-16B-Base & 2024.1 & DeepSeek & 3B & 16B & 45.0 & 18.8 (8-shot) & 4.3 & 26.8 & \href{https://huggingface.co/deepseek-ai/deepseek-moe-16b-base}{https://huggingface.co/deepseek-ai/deepseek-moe-16b-base} \\
Grok-1 & 2024.3 & xAI & 86B & 314B & 73.0 & 62.9 & 23.9 & 63.2 & \href{https://github.com/xai-org/grok-1}{https://github.com/xai-org/grok-1} \\ 
Qwen1.5-MoE-A2.7B & 2024.3 & Alibaba & 3B & 14B & 62.5 & 61.5 (8-shot) & - & 34.2 & \href{https://huggingface.co/Qwen/Qwen1.5-MoE-A2.7B}{https://huggingface.co/Qwen/Qwen1.5-MoE-A2.7B}  \\ 
DBRX Instruct & 2024.3 & Databricks & 36B & 132B & 73.7 & 72.8 & - & 70.1 & \href{https://huggingface.co/databricks/dbrx-instruct}{https://huggingface.co/databricks/dbrx-instruct} \\ 
Jamba-v0.1 & 2024.3 & AI21 Labs & 12B & 52B & 67.4 & 59.9 (3-shot) & - & 29.3 & \href{https://huggingface.co/ai21labs/Jamba-v0.1}{https://huggingface.co/ai21labs/Jamba-v0.1} \\ 
Mistral-8x22B-v0.1 & 2024.4 & Mistral & 39B & 141B & 77.8 & 78.6, 88.4 (8-shot) & 41.8 & 45.1 & \href{https://huggingface.co/mistralai/Mixtral-8x22B-v0.1}{https://huggingface.co/mistralai/Mixtral-8x22B-v0.1} \\ 
Arctic Instruct & 2024.4 & Snowflake & 17B & 480B & 67.3 & 74.2 & - & - & \href{https://huggingface.co/Snowflake/snowflake-arctic-instruct}{https://huggingface.co/Snowflake/snowflake-arctic-instruct} \\ 
DeepSeek-V2 & 2024.5 & DeepSeek & 21B & 236B & 78.5 & 79.2 (8-shot) & 43.6 & 48.8 &  \href{https://huggingface.co/deepseek-ai/DeepSeek-V2}{https://huggingface.co/deepseek-ai/DeepSeek-V2} \\ 
DeepSeek-V2-Chat (RL) & 2024.5 & DeepSeek & 21B & 236B & 77.8 & 92.2 (8-shot) & 53.9 & 81.1 & \href{https://huggingface.co/deepseek-ai/DeepSeek-V2-Chat}{https://huggingface.co/deepseek-ai/DeepSeek-V2-Chat} \\ 
Yuan 2.0-M32 & 2024.5 & IEIT & 4B & 40B & 72.2 & 92.7 (8-shot) & 55.9 (8-shot) & 74.4 & \href{https://huggingface.co/IEITYuan/Yuan2-M32}{https://huggingface.co/IEITYuan/Yuan2-M32} \\
Skywork-MoE-Base & 2024.6 & Kunlun & 22B & 146B & 77.4 & 76.1 & 31.9 & 43.9 & \href{https://huggingface.co/Skywork/Skywork-MoE-Base}{https://huggingface.co/Skywork/Skywork-MoE-Base}\\ 
OLMoE-0924-Base & 2024.9 & Ai2 & 1.3B & 6.9B & 54.1 & 3.0 (8-shot) & - & 22.4 & \href{https://huggingface.co/allenai/OLMoE-1B-7B-0924}{https://huggingface.co/allenai/OLMoE-1B-7B-0924}\\ 
DeepSeek-V3-Base & 2024.12 & DeepSeek & 37B & 671B & 87.1 & 89.3 (8-shot) & 61.6 & 65.2 & \href{https://huggingface.co/deepseek-ai/DeepSeek-V3-Base}{https://huggingface.co/deepseek-ai/DeepSeek-V3-Base}\\ 
\bottomrule
\end{tabular}
}
\end{table*}

\subsubsection{Activation Function}
\label{sec:activ_func}
Building upon dense Transformer architectures, sparse MoE models have adopted a progression of activation functions paralleling those in leading dense large language models, including BERT \cite{devlin2018bert}, T5 \cite{raffel2020exploring}, GPT \cite{brown2020language}, LLAMA \cite{touvron2023llama} and so on. 
The evolution of activation functions has seen a shift from ReLU \cite{glorot2011deep} to more advanced options such as GeLU \cite{hendrycks2016gaussian}, GeGLU \cite{shazeer2020glu}, and SwiGLU \cite{shazeer2020glu}. 
This trend extends to other components of MoE models, which now frequently incorporate Root Mean Square Layer Normalization (RMSNorm) \cite{zhang2019root}, Grouped Query Attention (GQA) \cite{ainslie2023gqa}, and Rotary Position Embeddings (RoPE) \cite{su2024roformer}.

\subsubsection{Shared Expert}
\label{sec:share_expert}
DeepSpeed-MoE \cite{rajbhandari2022deepspeed} innovatively introduces the Residual-MoE architecture, wherein each token is processed by a fixed expert and another selected through gating, achieving two experts engagement per layer without increasing the communication cost beyond that of top-1 gating.
This approach considers the gating-selected MoE expert as an error-correcting adjunct to the fixed dense FFN. 
A conceptually similar approach, Conditional MoE Routing (CMR), is employed in NLLB \cite{costa2022no}, which also combines the outputs of dense FFN and MoE layers. 
This paradigm of integrating fixed FFN with sparse MoE, often referred to as shared expert and illustrated in Figure~\ref{fig:moa_share} (b), has gained traction in recent language models such as DeepSeekMoE \cite{dai2024deepseekmoe}, OpenMoE \cite{xue2024openmoe}, Qwen1.5-MoE \cite{qwen_moe}, and MoCLE \cite{gou2023mixture}, indicating its ascension to a mainstream configuration.
Instead of using a single shared expert, DeepSeekMoE \cite{dai2024deepseekmoe} and Qwen1.5-MoE \cite{qwen_moe} employ multiple shared experts, due to their fine-grained expert segmentation design. 
He \textit{et al.} \cite{he2023pad} introduce Partially Dynamic Networks (PAD-Net), iteratively transforming partial parameters of gating-selected experts into static parameters (akin to shared experts) based on their impact on loss values. 
Zhao et al. \cite{zhao2024hypermoe} introduce HyperMoE, an innovative MoE framework that integrates expert-shared and layer-shared hypernetwork to effectively capture cross-expert and cross-layer information.
Additionally, based on the design of shared expert, ScMoE \cite{cai2024shortcut} decouples the MoE process to separately handle the representations from preceding layers and integrate them with the outputs processed by the shared expert of the current layer, thus improving efficiency by facilitating overlap in communication and computation. 
A comparable method to enhance overlapping is employed in the Dense-MoE hybrid transformer architecture, as delineated in Snowflake Arctic \cite{snowflake}, which bears resemblance to the LoRA MoE framework discussed in Section~\ref{sec:transformer_block} and illustrated in Figure~\ref{fig:mope} (d).

\subsection{Mixture of Parameter-Efficient Experts}
\label{sec:mope}
LLMs pretrained on generic massive datasets have demonstrated impressive abilities, enabling their deployment across diverse tasks \cite{gao2024higher}. However, to tailor a pretrained LLM for a specific downstream task, fine-tuning is essential. Traditional full fine-tuning, which updates all the parameters of the base model, is computationally intensive, especially as model sizes continue to grow \cite{ding2022delta}. To address this issue, research into parameter-efficient fine-tuning (PEFT) has emerged, intending to reduce computational demands during the adaptation of a generic pretrained model to particular tasks \cite{han2024parameter}. PEFT methods only update a small set of parameters while maintaining the rest of the base model untouched \cite{lialin2023scaling}. 
As an example of PEFT, LoRA \cite{hu2021lora} introduces two low-rank matrices to receive incremental updates associated with the task-specific fine-tuning. Only the LoRA matrices are updated while the base model is kept untouched during fine-tuning. 
These techniques have achieved state-of-the-art performance across numerous NLP tasks \cite{liu2022few, hu2021lora}.

Despite these successes, PEFT approaches often struggle with generalizing across multiple tasks due to their limited scope of trainable parameters and the potential for catastrophic forgetting \cite{li2024mixlora}. 
To mitigate these limitations, a line of mixture of parameter-efficient experts (MoPE) research has emerged, focusing on integrating the MoE framework with PEFT \cite{li2024mixlora, liu2023moelora}. MoPE incorporates the MoE's gating mechanism and multi-expert architecture, but with each expert constructed using PEFT techniques \cite{ostapenko2023case}. The subtle combination boosts PEFT's performance under the multi-task scenario \cite{zhu2023sira}. Additionally, by leveraging PEFT for constructing experts, MoPE operates with fewer parameters, achieving greater resource efficiency compared to traditional MoE models \cite{zadouri2023pushing}.

\begin{figure*}[t]
\centering
\includegraphics[width=0.95\linewidth]{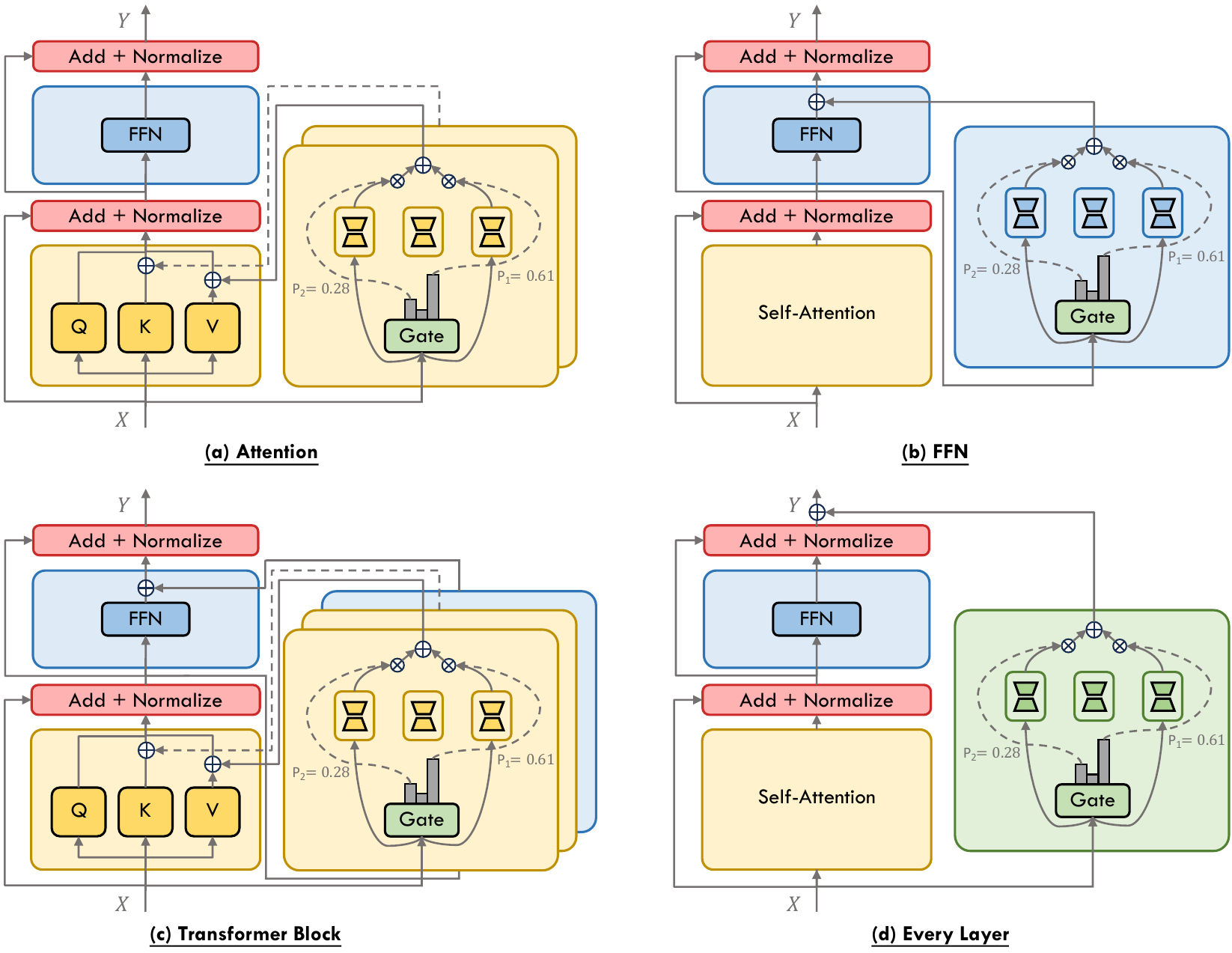}
\caption{The illustration of the taxonomy of MoPEs based placement within the Transformer model architecture. 
(a) exemplifies the integration of MoPE with the Key and Value modules of the attention mechanism, with applicability extending to Query and Output projection modules.
(b) represents the application of MoPE to the FFN.
(c) refers to the MoPE integration at the level of the Transformer block, wherein two distinct groups of experts are applied to attention and FFN, where separate sets of experts are allocated to both attention and FFN, each regulated by its own gating mechanism.
(d) illustrates a layer-wise integration of MoPE, in which each Transformer layer is regarded as a unified entity with a gating orchestrating the interplay among experts.}
\label{fig:mope}
\end{figure*}

MoPE harnesses the best of both fields: the task versatility of MoE and the resource efficiency of PEFT \cite{li2024mixlora}, positioning it as a promising area of study that pushes the boundaries of both fields. In the following subsection, we will give a taxonomy of MoPE, as depicted in Figure ~\ref{fig:mope}, based on their placement within the Transformer model architecture. We will then review recent MoPE research, summarizing the methodologies and contributions.

\subsubsection{Feed-Forward Network}
Following the conventional MoE structure, a series of investigations introduce the MoPE framework to the FFN layer of every Transformer block. During the training process, the focus is on optimizing the parameter-efficient experts and the gating mechanism, leaving the rest of the pretrained model intact.
As illustrated in Figure \ref{fig:mope}(b), the forward process under the MoPE framework integrated with FFN can be expressed as:
\begin{align}
    &\operatorname{FFN}^{MoE}(\mathbf{x}^\prime) = \operatorname{FFN}(\mathbf{x}^\prime) + \sum_{i=1}^n \mathbf{x}^\prime \Delta \mathbf{W}_i^{ffn} \cdot G^{ffn}(\mathbf{x}^\prime)_i,\label{eq:FFN}\\
    &\mathbf{x}^\prime = \operatorname{LayerNorm}(\operatorname{SA}(\mathbf{x}) + \mathbf{x}),\label{eq:FFN_x} 
\end{align}
where $\Delta \mathbf{W}^{ffn}$ and $G^{ffn}(\mathbf{x})$ is the parameter-efficient expert and gating function applied to the FFN layer, respectively.

One of the pioneering studies in this domain, LoRAMoE \cite{dou2023loramoe}, efficiently applies the MoPE structure to FFN.
LoRAMoE integrates a few plug-in LoRA experts into the FFN layer, employing a gating mechanism to orchestrate the experts' contributions.
Realizing the diversity in data distributions, LoRAMoE separates the experts into two distinct groups: one focuses on learning various downstream tasks, and the other is dedicated to aligning pretrained world knowledge with human instructions. To ensure that each group of experts maintains its focus, LoRAMoE defines a localized balancing constraint loss, which preserves the importance of each expert within its group while allowing different groups to concentrate on their respective tasks. This design enables LoRAMoE to effectively resolve the knowledge forgetting issue and enhance model performance on downstream tasks.
In a similar vein, AdaMix \cite{wang-etal-2022-adamix} injects a set of Adapter \cite{houlsby2019parameter} experts after the FFN layer in each Transformer block. Adapter tuning is a PEFT method that integrates a pair of feed-forward up and down projection matrices into the Transformer block. During fine-tuning, only the incremental Adapter blocks are updated, with the rest of the model unchanged.
AdaMix utilizes a stochastic routing policy that randomly selects the projection matrices during training, maintaining computational costs equivalent to a single adapter. To minimize service costs during inference, AdaMix averages the outputs of all the experts.

Taking a different approach, MixDA\cite{diao2023mixture} includes two training stages to leverage domain-specific knowledge while preserving learned information. 
During the first stage, MixDA only fine-tunes the domain-adapters that work in parallel with the FFN to acquire domain-specific knowledge while simultaneously retaining the world knowledge.
In the second stage, MixDA introduces a gating network and task-adapters on top of the FFN layer for tailoring the model to specific downstream tasks. This strategy allows for a more nuanced adaptation to the task at hand.
LLaVA-MoLE\cite{chen2024llava} extends the application of MoPE to multimodal tasks. It creates a set of LoRA experts for the FFN layer to handle inputs from different domains, enhancing the model's versatility. LLaVA-MoLE adopts a top-1 routing strategy, activating the most relevant expert based on the router's output distribution, thus maintaining computational costs close to a standard FFN with LoRA. This framework is effective in addressing data conflicts and consistently surpasses plain-LoRA baselines across diverse data configurations.

Contrasting with the MoPE implementations we have discussed, MixLoRA\cite{li2024mixlora} creates a LoRA-MoE framework that closely aligns with the conventional MoE models. Rather than just plugging in multiple lightweight experts, MixLoRA fuses LoRA experts with the shared FFN layer. By leveraging the base weights from a single FFN of the base model, MixLoRA streamlines the creation of the MoPE architecture. Furthermore, MixLoRA implements a high-throughput framework that significantly reduces token computation latency and memory usage during both training and inference, optimizing performance and efficiency.

\subsubsection{Attention}
A branch of research has been exploring the application of the MoPE framework with the attention mechanism.
These studies typically involve augmenting the attention mechanism by incorporating a gating network and a set of parallel experts. The MoPE framework can be applied to the Query, Key, Value, and Output projection modules, individually or in various combinations, within the attention mechanism. During the fine-tuning process, only the parameters of the activated experts and the gating network are updated, while the remaining parameters of the model are kept frozen.
For example, as shown in Figure \ref{fig:mope}(a), the integration of MoPE with the Key and Value module of the attention mechanism can be formalized as follows:
\begin{align}
    \operatorname{SA}^{MoE}(\mathbf{x}) = Softmax(\frac{\mathbf{Q}(\mathbf{K^T} + \sum_{i=1}^n \mathbf{x} \Delta \mathbf{W}_i^k \cdot G^k(\mathbf{x})_i)}{\sqrt{d_{head}}})\nonumber \\(\mathbf{V} + \sum_{i=1}^n \mathbf{x} \Delta \mathbf{W}_i^v \cdot G^v(\mathbf{x})_i),\label{eq:SA}    
\end{align}
where $\mathbf{Q}, \mathbf{K}, \mathbf{V}$ represents the Query, Key and Value modules, respectively. $\Delta \mathbf{W}^k$ and $G^k(\mathbf{x})$ denote the parameter-efficient expert and its corresponding gating function for the Key module. Similarly, $\Delta \mathbf{W}^v$ and $G^v(\mathbf{x})$ indicate the expert and the gating function for the Value module. Here, $n$ is the number of experts, and $d_{head}$ is the dimensions in the Multi-head Attention mechanism.

Recent studies have demonstrated the effectiveness of extending MoE to the attention layer \cite{zhang2022mixture, shen2024jetmoe, shen2023moduleformer}. Additionally, there is a new line of research has focused on the fusion of MoPE with the attention mechanism to enhance the model's efficiency and adaptability. For instance, MoELoRA \cite{luo2024moelora} applies MoE to the attention mechanism in a resource-efficient manner by leveraging LoRA \cite{hu2021lora} to construct the experts. 
Specifically, MoELoRA sets multiple LoRA experts to the Query and Value matrices of the attention mechanism, and utilizes a gating network to activate the top-$k$ experts related to the specific tasks during both training and inference phases.
To alleviate routing randomness, MoELoRA employs a contrastive learning loss to control the training of experts. The contrastive learning loss is designed to accentuate the differences in output distributions between experts, thereby encouraging them to capture diverse features relevant to the downstream tasks.
MoELoRA offers a solution for flexibly combining various computational modules tailored to downstream tasks.

Another framework, MoCLE\cite{gou2023mixture}, aims to resolve task conflicts that arise from the diversity of training tasks of different sources and formats.
MoCLE utilizes a clustering model to categorize different tasks and then leverages a router to direct the clustered input to LoRA experts inserted into the Query and Value modules of the attention mechanism. These LoRA experts contain a group of multiple task experts and a universal expert. Each task expert is dedicated to a particular task to reduce task conflicts, while the universal expert, trained on all tasks, helps to maintain model generalization.
SiRA\cite{zhu2023sira} introduces several lightweight LoRA adapters as experts, along with a top-$k$ gating mechanism. To mitigate load imbalance and over-fitting issues, SiRA incorporates a capacity constraint that limits the number of tokens each expert can process. Additionally, it employs an auxiliary loss to promote load balancing and an expert dropout mechanism to equalize the gating distribution.
SiRA provides an efficient and fine-grained approach to improving the quality of LoRA.

\subsubsection{Transformer Block}
\label{sec:transformer_block}
The integration of MoPE with the Transformer architecture has received substantial attention in recent research. This approach involves creating two groups of experts: one for the attention mechanism, and another for the FFN within the Transformer block. Each group is regulated by its gating mechanism to control the activation of the experts.
As exhibited in Figure \ref{fig:mope}(c), the forward process under the MoPE framework integrated with the Transformer block can be denoted as:
\begin{align}
    &y = \operatorname{LayerNorm}(\mathbf{x}^\prime + \operatorname{FFN}^{MoE}(\mathbf{x}^\prime)),\label{eq:Transformer}\\
    &\mathbf{x}^\prime = \operatorname{LayerNorm}(\operatorname{SA}^{MoE}(\mathbf{x}) + \mathbf{x}).\label{eq:Transformer_sa} 
\end{align}
MoV \cite{zadouri2023pushing} is one of the notable attempts that combine MoPE with the Transformer block to pursue parameter efficiency. 
Utilizing the PEFT method, $\text{(IA)}^3$ \cite{liu2022few}, MoV introduces tunable vectors that re-scale the Key and Value modules in the attention mechanism, as well as the activation within the FFN. By substituting conventional experts with $\text{(IA)}^3$ vectors and updating only these lightweight experts and their corresponding gating during fine-tuning, MoV significantly reduces the computational burden associated with gradient calculations and lessens the memory footprint required for model storage. Similarly, MoLORA \cite{zadouri2023pushing} employs multiple LoRA experts to the attention and FFN blocks, outperforming the standard LoRA approach.
UniPELT \cite{mao2022unipelt} proposed a hybrid framework that integrates three representative PEFT methods as experts, namely Adapter \cite{houlsby2019parameter}, Prefix-tuning \cite{li2021prefix}, and LoRA \cite{hu2021lora}. Prefix-tuning is a method that freezes the base model and optimizes the continuous task-specific vectors prepended to the input of the attention. Within the UniPELT framework, LoRA matrices are applied to the weight matrices of Query and Key in the attention mechanism, Prefix vectors are added to the Key and Value modules, and the Adapter block is inserted after the FFN layer. UniPELT leverages different gating mechanisms to dynamically activate the experts, efficiently finding the approaches that best suit the given task.

Further broadening the scope of the LoRA-MoE framework, Omni-SMoLA\cite{wu2023omni} extends the MoPE with three sets of LoRA experts, each tailored to handle text tokens, visual tokens, and multimodal tokens, respectively. The specialization enables the architecture to enhance performance across various vision-and-language tasks.
In the context of MoPE research, the number of experts emerges as a critical hyperparameter influencing downstream task performance \cite{zadouri2023pushing, wu2024mixture}. Additionally, the use of many experts may lead to redundancy \cite{chen2023sparse}. 
MoLA \cite{gao2024higher} is one of the pioneering work that explores the expert allocation issue. It proposes a LoRA-MoE framework with a Layer-wise Expert Allocation, which enables the flexible employment of varying numbers of experts across different layers. The expert allocation strategy proposed by MoLA further improves the effectiveness of the LoRA-MoE framework.
In the specialized field of medical applications, MOELoRA\cite{liu2023moelora} tackles the challenges of task variety and high adaptation cost. It integrates LoRA experts and task-motivated gate functions into the attention and FFN of each layer.
The gating utilizes task identity to modulate expert contributions, creating unique parameter sets tailored to individual tasks.
The design of MOELoRA combines the strengths of both MoE and LoRA, strengthening LLM's capability in medical domains.
Liu \textit{et al.} \cite{liu2024intuition} design a novel framework, named Intuition-MoR1E, which leverages the inherent semantic clustering of instances to emulate cognitive processes in the human brain for multitasking purposes. 
This framework provides implicit guidance to the gating mechanism, thereby enhancing feature allocation.
Furthermore, they introduce a cutting-edge rank-1 expert architecture. This architecture advances beyond the conventional 1-1 mapping of two weight matrices $W_{up}^i$ and $W_{down}^i$ in LoRA expert composition, facilitating a flexible combination of any $W_{up}^i$ with any $W_{down}^j$ to form an expert. 
They implement MoE in the transformer blocks, specifically targeting the Query, Key, Value, and FFN modules.
Xu \textit{et al.} \cite{xu2024meteora} present Multiple-Tasks Embedded LoRA (MeteoRA), a scalable and efficient framework that embeds multiple task-specific LoRA adapters into the base LLM using a full-mode MoE architecture. This framework incorporates custom GPU kernel operators to accelerate LoRA computation while maintaining memory overhead.

\subsubsection{Every Layer}
There has been considerable interest in incorporating MoPE into fundamental components such as the attention, FFN, and Transformer block. Existing work often approaches the attention mechanism and FFN independently, employing distinct gating mechanisms to modulate them separately. 
Rather than treating these elements isolated, there is a new direction that considers the Transformer layer as an integrated whole. This shift in perspective allows for the application of the MoPE framework to the entire Transformer layer, capturing the combined dynamics of the attention and FFN within a unified approach.
As illustrated in Figure \ref{fig:mope}(d), the forward process under the MoPE framework integrated with every layer is as follows:
\begin{align}
    &y = \operatorname{LayerNorm}(\mathbf{x}^\prime + \operatorname{FFN}(\mathbf{x}^\prime)) + \sum_{i=1}^n \mathbf{x} \Delta \mathbf{W}_i^{layer} \cdot G^{layer}(\mathbf{x})_i,\label{eq:EveryLayer}\\ 
    &\mathbf{x}^\prime = \operatorname{LayerNorm}(\operatorname{SA}(\mathbf{x}) + \mathbf{x}),\label{eq:Transformer_sa_sub} 
\end{align}
where $\Delta \mathbf{W}^{layer}$ and $G^{layer}(\mathbf{x})$ is the parameter-efficient expert and gating function applied to the entire layer, respectively.

In this context, the approach presented by MoLE\cite{wu2024mixture} provides innovative insights. MoLE identifies that various layers within LoRA exhibit unique features. In response to this finding, MoLE pursues to enhance the composition effect of trained LoRAs by dynamically adjusting the layer-specific weights according to the desired objective.
This is achieved by integrating a set of trained LoRAs and a gating function into each layer. MoLE treats each layer of trained LoRAs as an individual expert and only trains the gating to learn the optimal composition weights for a specified domain. This dynamic linear composition strategy significantly extends the versatility of LoRA, enabling its application across a broader spectrum of practical scenarios.

\subsection{Training \& Inference Scheme}
\label{sec:train_inference}
The architectural advancements of Mixture-of-Experts (MoE) models have been complemented by extensive research into training and inference schemes, with the objective of optimizing both computational efficiency and model quality.

\textbf{Original Training \& Inference Scheme.}
Initial training methodologies, as established in seminal works \cite{shazeer2017outrageously,lepikhin2020gshard,fedus2022switch,zoph2022st,riquelme2021scaling}, involve constructing an MoE model and training it from scratch, with inference directly following the model configurations of training. 

\begin{figure*}
    \centering
    \includegraphics[width=1\linewidth]{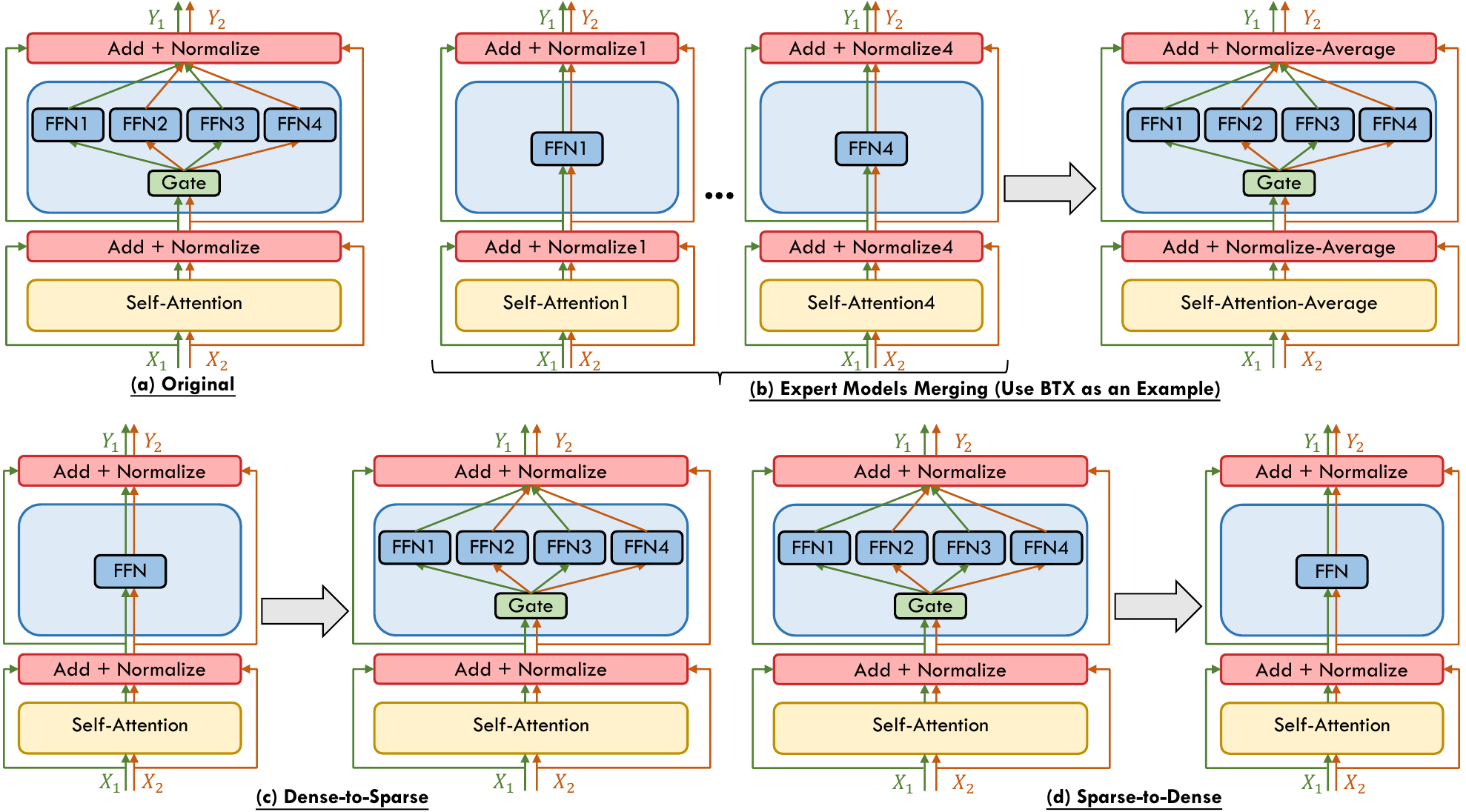}
    \caption{Schematic representation of training and inference schemes related to MoE. It provides an abstracted view of model transition, without focusing specific model states during training or inference. Subfigure (a) depicts the original scheme without architectural transformation. Subfigure (b) depicts the merging of distinct expert models, exemplified by BTX \cite{sukhbaatar2024branch}. Subfigure (c) depicts the transition from a dense model to a sparse model. Subfigure (d) depicts the inverse process, where a sparse model is converted to a dense model.}
    \label{fig:training_inference_scheme}
\end{figure*}

The advent of MoE models has introduced novel paradigms for training and inference, enabling a flexible approach that synergizes the strengths of dense and sparse models while mitigating their respective weaknesses.
As depicted in Figure~\ref{fig:training_inference_scheme}, we divide the emerging schemes into three distinct categories: Dense-to-Sparse, which entails initiating with dense model training and progressively transitioning to a sparse MoE configuration \cite{wu2022residual,dua2022tricks,komatsuzaki2022sparse,wei2024skywork,lin2024moe,pan2024dense,nie2021evomoe,chen2022sparse}; Sparse-to-Dense, which involves degrading a sparse MoE model to a dense form that is more conducive to hardware implementation for inference \cite{xue2022one,chen2022task,huang2023experts}; and Expert Models Merging, a process of integrating multiple pretrained dense expert models into a unified MoE model \cite{li2022branch,sukhbaatar2024branch,wang2023fusing}.

\subsubsection{Dense-to-Sparse}
\label{sec:dense_to_sparse}
To mitigate the training overhead associated with vision MoE transformer models, the Residual Mixture-of-Experts (RMoE) approach \cite{wu2022residual} commences with training a dense, non-MoE model on the upstream task, followed by an efficient fine-tuning stage to transition into a MoE model. 
This process reveals that directly inheriting expert weights from a pretrained non-MoE model's FFN can lead to suboptimal performance, necessitating an alignment between the MoE and FFN outputs during the fine-tuning phase. 
Similarly, Dua \textit{et al.} \cite{dua2022tricks} advocate for initially training a dense model, subsequently introducing sparsity by incorporating a randomly initialized gating module, and further training the model's feed-forward layers under sparsity conditions—specifically, by updating the weights locally within each compute node rather than averaging gradients across nodes.

Nie \textit{et al.} \cite{nie2021evomoe} present EvoMoE, an efficient end-to-end MoE training framework. 
EvoMoE decouples the joint learning of experts and the sparse gate, emphasizing the acquisition of foundational knowledge through a single expert at the inception of training. Subsequently, it spawns multiple diverse experts and advances the diversification of experts by training with the novel Dense-to-Sparse gate (DTS-Gate). The DTS-Gate initially operates as a dense activation of all experts, then progressively and adaptively constricting to route tokens to a reduced number of experts.
A similar strategy is employed in the development of the MoE-LLaVA \cite{lin2024moe} large vision-language model, which commences with a dense model, subsequently multiplies the feed-forward network (FFN) to create expert initializations, and proceeds to train exclusively the MoE layers, while keeping the remaining model components static.

Komatsuzaki \textit{et al.} \cite{komatsuzaki2022sparse} highlight the efficiency of sparse models in terms of quality and computational cost, yet acknowledge their significant data requirements and the expense of training from scratch at scale. 
To address this, they introduce a scheme termed "sparse upcycling," which leverages pre-existing training investments by initializing a sparsely activated MoE model from a pretrained dense checkpoint. 
This involves transferring all parameters—and optionally their associated optimizer states—from the original checkpoint, with the exception of the MoE gating network parameters, which are not present in the dense model.
Notably, the new MoE layers are populated with identical copies of the original dense model's FFN layers, and the gating mechanism weights are initialized randomly.
A critical obstacle in model upcycling is the initial performance decrease resulting from structural modifications to a trained network. 
To mitigate this performance regression during upcycling, the researchers propose normalizing the gating scores for each input token to 1, which are used to combine the outputs of multiple experts. 
This approach is grounded in the notion that, in the dense model, each token was processed by a singular ``expert" FFN. 
While this normalization proved beneficial for upcycled vision models, it was found to be detrimental to the performance of upcycled language models.
In the first introduction of sparse MoE by \cite{shazeer2017outrageously}, the softmax function was applied to the selected top-$k$ gating values, which normalizes the combination gating scores to 1, formulated as $\operatorname{softmax}(\operatorname{TopK}(g(\mathbf{x}; \mathbf{\Theta}), k))$. 
However, subsequent LLMs \cite{lepikhin2020gshard,fedus2022switch,xue2024openmoe} with MoE have evolved to apply the softmax function across all potential gating values before isolating the top-$k$ subset, formulated as $\operatorname{TopK}(\operatorname{softmax}(g(\mathbf{x}; \mathbf{\Theta})), k)$.

Building upon the sparse upcycling technique \cite{komatsuzaki2022sparse}, the Skywork-MoE model \cite{wei2024skywork} leverages the foundational architecture of its pre-developed Skywork-13B model \cite{wei2023skywork}, adopting its dense checkpoints as a foundation for initial states. 
Their empirical evidence indicates that the decision between sparse upcycling and training from scratch should be informed by both the performance of available dense checkpoints and the MoE-specific training resources, as models trained from scratch consistently surpass their upcycled counterparts in performance.
The study observes a decline in average expert similarity throughout the training of upcycled MoEs, suggesting a diversification of experts emerges during the process. Importantly, the Skywork-MoE analysis reveals that models with greater expert similarity tend to underperform, establishing expert similarity as a potential diagnostic tool during MoE training for upcycled models. 
Conversely, the expert similarity in models trained from scratch remains minimal, implying that non-uniform expert initialization promotes diversification.

Pan \textit{et al.} \cite{pan2024dense} posit that the parameter inefficiency observed in MoE models stems from conventional sparse training methodologies, where only a selected group of experts is engaged and refined for each input token. 
To counteract this, they introduce a hybrid framework for MoE models, denoted as DS-MoE, which integrates dense training (activating all the experts) with sparse inference (sparse expert activation) to achieve higher computation and parameter efficiency. 
Notably, DS-MoE maintains activation for all self-attention experts (MoA \cite{zhang2022mixture}) during inference but selectively activates FFN experts, reflecting the observation that self-attention layers manifest considerably less sparsity compared to FFN layers.

Chen \textit{et al.} \cite{chen2022sparse} introduce SMoE-Dropout, an innovative plug-and-play training framework, which initially modularizes the FFN into a sequence of smaller FFNs then employs a random policy parameterized by fixed weights to route token to k experts with the largest response.
Progressively, the framework activates an increasing number of experts, preventing overfitting to the amounts of used network capacity during training.
MoEfication \cite{zhang2022moefication} investigates various strategies for expert construction in T5 models, including random splitting, parameter clustering, and building co-activation graphs. 
MoEBERT \cite{zuo2022moebert} implements an importance-based method for adapting FFNs into experts within BERT models.
LLaMA-MoE \cite{llama-moe-2023} conducts an extensive examination of different expert construction methods, ultimately proposing a straightforward random division approach that partitions the parameters of FFNs into non-overlapping experts.
Emergent MoEs (EMoE) \cite{qiu2024unlocking} splits certain FFN layers of the original model into MoE layers with clustering-based experts construction and avg-k gating, which ameliorates the parameter updating during fine-tuning and can even be abandoned afterward to preserve the original model architecture.

\subsubsection{Sparse-to-Dense}
Switch Transformer \cite{fedus2022switch} studies the distillation of large sparse models into smaller dense counterparts to achieve parameter efficiency for deployment. 
The study reports that initializing the corresponding weights of dense model from non-expert layers of MoE model modestly enhances performance, facilitated by the consistent dimension of non-expert layers.
Furthermore, an improvement in distillation is observed when employing a mixture of 0.25 for the teacher probabilities and 0.75 for the ground truth label.
Leveraging both methods, the distillation preserves approximately 30\% of the sparse model's quality gains using only about 5\% of the parameters.
Similarly, Xue \textit{et al.} \cite{xue2022one} address the challenges of overfitting, deployment difficulty, and hardware constraints associated with sparse MoE models. 
Drawing inspiration from human learning paradigms, they propose a new concept referred to as `knowledge integration' aimed at creating a dense student model (OneS) that encapsulates the expertise of a sparse MoE model. 
Their framework first implements knowledge gathering, explored through a variety of methods such as summation, averaging, top-$k$ Knowledge Gathering, and their Singular Value Decomposition Knowledge Gathering. 
Then, they refine the dense student model by knowledge distillation to mitigate noise introduced by the knowledge gathering. 
The OneS model retains 61.7\% of the MoE's benefits on ImageNet and 88.2\% on NLP datasets.
Further investigations into MoE model distillation are also conducted by other researchers \cite{rajbhandari2022deepspeed,costa2022no}.

Chen \textit{et al.} \cite{chen2022task} highlight the challenges associated with deploying MoE models on resource-constrained platforms, such as cloud or mobile environments. 
Observing that only a fraction of experts contribute significantly to MoE fine-tuning and inference, they propose a method for the progressive elimination of non-essential experts. This approach retains the advantages of MoE while simplifying the model into a single-expert dense structure for the target downstream task.
Similarly, ModuleFormer \cite{shen2023moduleformer} applies a comparable pruning technique, removing task-unrelated experts for a lightweight deployment.
Huang \textit{et al.} \cite{huang2023experts} separate the training and inference stages for Vision Transformers (ViTs). They substitute certain FFNs in the ViT with custom-designed, efficient MoEs during training. 
These MoEs assign tokens to experts using a random uniform partition and incorporate Experts Weights Averaging (EWA) on these MoEs at the end of each iteration. 
After training, the MoEs are converted back to FFNs through averaging of expert weights, thus reverting the model to its original dense ViT for inference.
He \textit{et al.} \cite{he2024demystifying} propose a unified framework for MoE model compression, encompassing two strategies: Expert Slimming, which compresses individual experts via pruning and quantization, and Expert Trimming, which structurally removes unimportant experts.

\subsubsection{Expert Models Merging}
Li \textit{et al.} \cite{li2022branch} introduce the Branch-Train-Merge (BTM) algorithm, a method for the communication-efficient training of language models (LMs). 
BTM independently trains a set of expert LMs (ELMs), each tailored to a specific domain within the training corpus, such as scientific or legal text. 
These ELMs, which operate without shared parameters, can be ensembled or parameter-averaged at inference to coalesce into a singular LM.
Expanding on this concept, Sukhbaatar \textit{et al.} \cite{sukhbaatar2024branch} present Branch-Train-MiX (BTX), designed to combine the strengths of BTM and Mixture-of-Experts while addressing their respective limitations. 
BTX maintains separate training for multiple expert LLMs, akin to BTM, but subsequently integrates these experts within a unified MoE model. 
Specifically, it consolidates the FFNs from all ELMs into a singular MoE module at each layer, with a gating network determining the appropriate FFN expert for each token. 
Other components, such as the self-attention layers from ELMs, are merged by averaging their weights. 
The resulting model then undergoes MoE fine-tuning on all the combined data to enable the gate to effectively mix the FFN experts.

Wang \textit{et al.} \cite{wang2023fusing} point out that while the emergence of Foundation Models made it easier to obtain expert models tailored to specific tasks, the heterogeneity of data at test time necessitates more than a single expert. 
Accordingly, they explore the Fusion of Experts (FoE) challenge, which aims to integrate outputs from expert models that provide diverse but complementary insights into the data distribution, formulating it as an instance of supervised learning.

\begin{figure*}
\centering
\includegraphics[width=\linewidth]{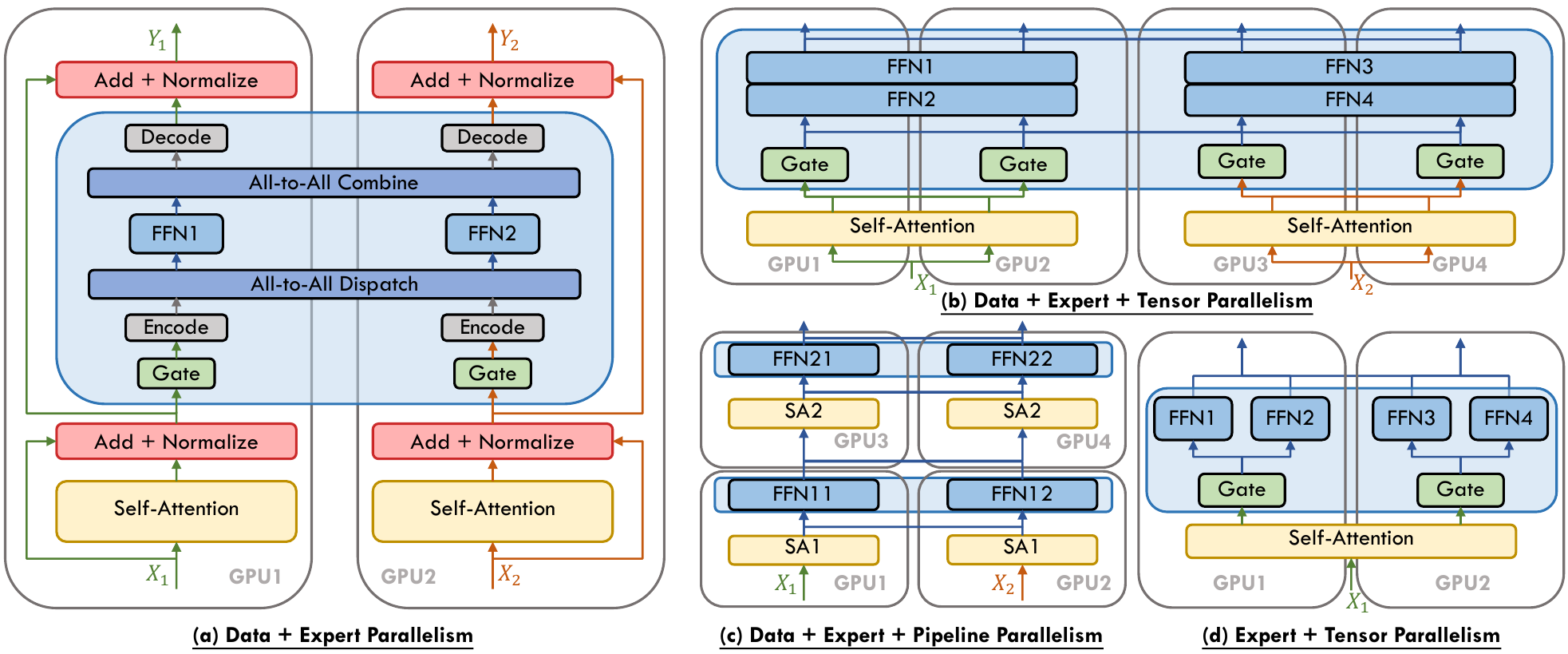}
\caption{Schematic depiction of diverse parallel strategies for MoE. For clarity and conciseness, this illustration omits some All-to-All, All-Reduce, Point-to-Point communication within parallelism, and Normalization, Encode, Decode, Gate in subfigures (b), (c), and (d).}
\label{fig:expertparallelism}
\end{figure*}

\subsection{Derivatives}
Building upon the principles of algorithm design highlighted earlier, numerous studies have drawn inspiration from the Mixture of Experts (MoE) framework, proposing a range of MoE variants. 
We categorize these innovative models as derivatives of the MoE.
For instance, Xue \textit{et al.} \cite{xue2022go} introduced WideNet, an approach that increases model width by substituting the feed-forward network (FFN) with an MoE layer while maintaining shared trainable parameters across Transformer layers, except for the normalization layers. 
Subsequently, Tan \textit{et al.} \cite{tan2023sparse} presented the Sparse Universal Transformer (SUT), an efficient enhancement of the Universal Transformer, which is characterized by parameter-sharing across its layers. 
SUT incorporates a Sparse Mixture of Experts and a novel stick-breaking-based dynamic halting mechanism, thus reducing computational complexity without compromising parameter efficiency or generalization capabilities.
Moreover, the traditional MoE models often employ discrete matching between experts and tokens \cite{shazeer2017outrageously,lepikhin2020gshard,du2022glam,artetxe2021efficient,zhou2022mixture,fedus2022switch,zoph2022st}, a practice associated with training instability and uneven expert utilization. 
To address these challenges, Antoniak \textit{et al.} \cite{antoniak2023mixture} innovatively proposes the Mixture of Tokens (MoT), which blends tokens from different examples before presenting them to the experts. 
Thus, MoT enables the model to benefit from a wider array of token-expert combinations.


Recently, the MoE's principle of assigning specialized knowledge to individual experts has been adapted to parameter-efficient fine-tuning (PEFT). 
Choi \textit{et al.} \cite{choi2023smop} propose the sparse mixture-of-prompts (SMoP), a method that utilizes a gating mechanism to train multiple short soft prompts, each adept at processing distinct subsets of data. 
This addresses the inefficiencies encountered with long soft prompts during prompt tuning.
The MoE framework has also been integrated into lifelong learning (LLL), which seeks to facilitate continuous learning from an ongoing stream of data. 
The Lifelong-MoE model \cite{chen2023lifelong} dynamically expands model capacity by adding experts with regularized pretraining, effectively mitigating the issue of catastrophic forgetting \cite{kirkpatrick2017overcoming} typically associated with straightforward fine-tuning.
In a recent development, the MoE concept of conditional computation has been further refined to optimize resource allocation in transformer-based language models (LMs). 
The Mixture-of-Depths (MoD) \cite{raposo2024mixture} employs a binary gating network to decide whether a token should be processed by a given Transformer layer. 
As a result, MoD transformers can dynamically allocate computational resources (FLOPs) to specific sequence positions, achieving a lower overall FLOP footprint compared to vanilla or MoE-based transformers.

In summary, the evolution of MoE derivatives reveals a trend where models either integrate the conditional computation aspect of the gating mechanism or merge the MoE structure with various tasks achieved by assigning specialized knowledge to individual experts, such as aforementioned prompt tuning \cite{choi2023smop} and lifelong learning \cite{chen2023lifelong} with MoE, demonstrating the versatility and adaptability of the MoE architecture across different domains.

\begin{table*}[t]
\caption{Comparative overview of the open-source MoE system frameworks, arranged chronologically by reference publication date from newest to oldest. We give the count of GitHub stars as of June 2024.}
\vskip -0.1in
\label{tab:moe_system}
\resizebox{1\textwidth}{!}{
\renewcommand\arraystretch{1.2}
\begin{tabular}{l|l|ccc|cc} 
\toprule
\multirow{2}{*}{\textbf{Reference}} & \multirow{2}{*}{\textbf{Affiliation}} & \multicolumn{3}{c|}{\textbf{Optimizations}} & \multirow{2}{*}{\textbf{Link}} & \multirow{2}{*}{\textbf{Star}}\\ 
 \cline{3-5} & & Computation & Communication & Storage & \\ 
\midrule
OpenMoE \cite{xue2024openmoe}  & Colossal-AI & \checkmark & \checkmark & & \href{https://github.com/hpcaitech/ColossalAI}{https://github.com/hpcaitech/ColossalAI} & 38K\\ 
ScatterMoE \cite{tan2024scattered}  & Mila Quebec & \checkmark & & & \href{https://github.com/shawntan/scattermoe}{https://github.com/shawntan/scattermoe} & 140\\ 
Megablocks \cite{gale2023megablocks}  & Stanford University & \checkmark & & & \href{https://github.com/stanford-futuredata/megablocks}{https://github.com/stanford-futuredata/megablocks} & 1.1K\\ 
Tutel \cite{hwang2023tutel} & Microsoft & \checkmark & \checkmark & & \href{https://github.com/microsoft/tutel}{https://github.com/microsoft/tutel} & 672\\
SE-MoE \cite{shen2022se} & Baidu &	\checkmark & \checkmark & \checkmark & \href{https://github.com/PaddlePaddle/Paddle}{https://github.com/PaddlePaddle/Paddle}  & 21K\\ 
HetuMoE \cite{nie2022hetumoe}  & Peking University & \checkmark & \checkmark & & \href{https://github.com/PKU-DAIR/Hetu}{https://github.com/PKU-DAIR/Hetu} & 236\\ 
Deepspeed-MoE \cite{rajbhandari2022deepspeed} & Microsoft & \checkmark & \checkmark &  & \href{https://github.com/microsoft/DeepSpeed}{https://github.com/microsoft/DeepSpeed} & 33K\\ 
FastMoE \cite{he2021fastmoe} & Tsinghua University & \checkmark & \checkmark & & \href{https://github.com/laekov/fastmoe}{https://github.com/laekov/fastmoe} & 1.4K\\ 
Fairseq \cite{ott2019fairseq,artetxe2021efficient}  &	Meta & & & & \href{https://github.com/facebookresearch/fairseq/tree/moe}{https://github.com/facebookresearch/fairseq/tree/moe}  & 29K\\ 
Mesh-TensorFlow \cite{shazeer2018mesh} & Google & & & & \href{https://github.com/tensorflow/mesh}{https://github.com/tensorflow/mesh}  & 1.6K\\ 
\bottomrule
\end{tabular}
}
\end{table*}

\section{System Design of Mixture of Experts}\label{sec:system}

While Mixture of Experts (MoE) has been increasingly leveraged to enhance the capabilities of large language models, its adoption introduces new challenges to existing training and inference systems, due to the inherently sparse and dynamic nature of its computational workload.
GShard \cite{lepikhin2020gshard} introduces expert parallelism that implements parallel gating and expert computation by dispatching partitioned local tokens with load balancing limit of expert capacity. Since then, expert parallelism has emerged as a fundamental strategy to facilitate efficient scaling of MoE models.
This approach can be viewed as an augmentation of data parallelism \cite{rajbhandari2020zero, ren2021zero, rajbhandari2021zero}, where each expert in an MoE layer is assigned to a distinct device, while all non-expert layers are duplicated across devices.
As depicted in Figure~\ref{fig:expertparallelism}(a), the process flow of expert parallelism consists of the following sequential operations: gate routing, input encode, All-to-All dispatch, expert computation, All-to-All combine, and output decode.
In general, the input size for general matrix multiply (GEMM) needs to be large enough to achieve optimal utilization and throughput that computing device necessitates.
Therefore, input encode is employed to aggregate the input tokens of a same expert into a contiguous memory space, as determined by the token-expert mapping from gate routing. 
Subsequently, the All-to-All dispatch is employed to send the input tokens to their corresponding experts across the distributed devices.
Following the localized computation by the experts, the inverse process---All-to-All combine and output decode---reinstates the original data layout according to the gating indices.

Furthermore, the synergy of expert parallelism \cite{fedus2022switch, hwang2023tutel, singh2023hybrid, ma2022bagualu, zheng2022alpa} with other existing parallel strategies (tensor \cite{shoeybi2019megatron, smith2022using, narayanan2021efficient}, pipeline \cite{huang2019gpipe, narayanan2019pipedream, qi2023zero}, sequence parallelism \cite{li2021sequence, korthikanti2023reducing, jacobs2023deepspeed}) has been investigated to enhance the scalability and efficiency of MoE models in large-scale distributed environments. 
As shown in Figure~\ref{fig:expertparallelism}, we illustrate several examples of hybrid parallelism, encompassing (b) data + expert + tensor parallelism \cite{fedus2022switch,rajbhandari2022deepspeed,singh2023hybrid,hwang2023tutel,zhai2023smartmoe}, (c) data + expert + pipeline parallelism \cite{he2022fastermoe,zhai2023smartmoe,hwang2023tutel}, and (d) expert + tensor parallelism \cite{wei2024skywork}.
It is imperative to recognize that the choice of distributed parallelism strategies influences a complex interplay between computation efficiency, communication overhead, memory occupation, potentially affected by various hardware configurations. 
Consequently, the deployment strategies for practical applications necessitate nuanced trade-offs and bespoke designs tailored to specific use-case scenarios.

In the subsequent discussion, we delineate the challenges introduced by MoE models from computation, communication, and storage aspects, concurrently reviewing existing research addressing these issues. Table~\ref{tab:moe_system} shows an overview of the open-source MoE frameworks.

\subsection{Computation}
Despite MoE is designed to scale model parameters efficiently without increasing computational demand, it encounters challenges pertaining to computational efficiency. 
One concern is the imbalance of computational load across distributed devices employing expert parallelism, which incurs significant synchronization overhead as the system awaits the processing completion of the most heavily loaded expert. 
Such issues are typically addressed through algorithmic strategies, such as optimized gating mechanisms and expert capacity adjustments, as discussed in Section~\ref{sec:gating_function}.
Besides, solutions like SE-MoE \cite{shen2022se}, Tutel \cite{hwang2023tutel}, FlexMoE \cite{nie2023flexmoe} and SmartMoE \cite{zhai2023smartmoe} have introduced dynamic expert placement strategies to distribute the workload as equally as possible among devices. 
Additionally, FasterMoE \cite{he2022fastermoe} has implemented a novel dynamic shadowed expert strategy, replicating experts on multiple devices to mitigate severe load imbalance.
These model placement related strategies impact both computation and communication efficiency.

Another concern is that MoE introduces additional computational overhead through operations including gate routing, input encode, and output decode. 
Unlike expert computations, which mirror operations in dense models and benefit from extensive optimization on prevalent hardware such as GPUs, these MoE operations are characterized by redundant computation and memory movement, resulting in low efficiency on computing devices.
Therefore, recent studies like DeepSpeed-MoE\cite{rajbhandari2022deepspeed}, FastMoE \cite{he2021fastmoe}, HetuMoE \cite{nie2022hetumoe} and Tutel \cite{hwang2023tutel} have focused on the development of tailored GPU kernels to enhance the efficiency of MoE operations.

In contexts where multiple experts are deployed on a single GPU device, MegaBlocks \cite{gale2023megablocks} reformulates MoE computation in terms of block-sparse operations, developing specialized block-sparse GPU kernels that efficiently handle the dynamic workloads without dropping tokens.
Zheng \textit{et al.} \cite{zheng2023pit} propose PIT, a deep-learning compiler tailored for dynamic sparsity of MoE, which can find feasible PIT rules for all the operators within a model and generate optimized GPU kernels for them. PIT employs a novel tiling mechanism, utilizing the Permutation Invariant Transformation (PIT)—--a mathematically proven property---to transform multiple sparsely located micro-tiles into a GPU-efficient dense tile without changing the computation results, thereby achieving both high GPU utilization and low coverage waste. 
Despite these advancements, Tan \textit{et al.} \cite{tan2024scattered} highlight remaining optimization potential within current MoE frameworks such as MegaBlocks and PIT, which commence with an initial scatter-to-group data copy that increases memory footprint and requires a translation of the MoE problem into the sparse matrix format. 
Although this translation contributes minimally to computation overhead, it imposes limitations on the transparency and adaptability of extending MegaBlocks to modules beyond the FFN.
To address these issues, Tan \textit{et al.} \cite{tan2024scattered} propose ScatterMoE, a MoE implementation designed to effectively minimize the memory footprint. 
ScatterMoE leverages ParallelLinear, a linear module capable of executing grouped matrix operations on scattered groups. 
This approach yields intermediate representations (e.g., the hidden states of an SMoE MLP) that are directly accessible as standard PyTorch tensors, allowing for easy extensions of MoE methods to other types of expert modules.

\subsection{Communication}\label{sec:communication}
In expert parallelism, the quadruple invocation of All-to-All communication during both the forward and backward propagation phases within each MoE layer causes a significant overhead, even emerging as the primary constraint on efficiency. 
The All-to-All communication paradigm encompasses both intra-node (via PCIe, pre-4th-generation NVLink) and inter-node (Ethernet, Infiniband, 4th-generation NVLink) communication channels. The efficiency of such communication is contingent upon a multitude of factors, including the heterogeneity of channel bandwidths, network topology, and the collective communication algorithms. 
Moreover, load imbalances intrinsic to MoE may exacerbate these inefficiencies by inducing synchronization delays.

To optimize the use of high intra-node bandwidth and low inter-node bandwidth, DeepSpeed-MoE \cite{rajbhandari2022deepspeed}, HetuMoE \cite{nie2022hetumoe} and ScheMoE \cite{shi2024schemoe} have introduced hierarchical All-to-All communication strategies that enhance intra-node process and reduce inter-node data exchanges.
Besides, FasterMoE \cite{he2022fastermoe}, TA-MoE \cite{chen2022ta} and SE-MoE \cite{shen2022se} have introduced topology-aware routing strategies aimed at mitigating cross-node expert selection, thereby reducing inter-node communication burdens. 
Additionally, ExFlow \cite{yao2024exploiting} exploits expert affinity, anticipating expert allocation across layers to maximize the retention of token processing within local GPU confines.
The strategic allocation of experts to minimize network traffic and leverage high-bandwidth connections is a prevalent approach in distributed MoE system \cite{rajbhandari2022deepspeed,singh2023hybrid,wei2024skywork}. Moreover, this is often integrated with the placement design of non-expert modules to optimize overall system performance.

Given the concurrent feature of communication and computation, pipelining \cite{huang2019gpipe, narayanan2019pipedream, qi2023zero} is commonly employed to overlap their execution, thereby reducing the total time cost.
This technique, which is integrated in systems such as Tutel \cite{hwang2023tutel}, FasterMoE \cite{he2022fastermoe}, PipeMoE \cite{shi2023pipemoe} and MPipeMoE \cite{zhang2024mpmoe}, orchestrates overlapping between All-to-All communication and expert computation. 
Notably, Lancet \cite{jiang2024lancet} underscores the inherent constraints of these pipelining methods, particularly the bounded duration for which expert computation and communication can overlap. 
To address this limitation, Lancet partitions non-MoE computations and integrates them into the pipeline during forward pass, and strategically schedules gradient weight computations to augment overlap in the backward pass.
Punniyamurthy \textit{et al.} \cite{punniyamurthy2023optimizing} also emphasize the challenge posed by collective communications, which are often on the critical path, noting the difficulty of hiding their latency by overlapping kernel-granular communication and computation due to the absence of independent computation. 
Their solution involves fusing computation with dependent collective communication by leveraging GPU's massive parallelism and GPU-initiated communication.

Aiming to break the inherent dependencies and thereby extend the overlap duration, ScMoE \cite{cai2024shortcut} restructures the MoE architecture to simultaneously process representations from preceding layers while engaging with current-layer representations. This decoupling of communication dependencies facilitates substantial, and in certain cases, complete overlapping between communication and computation.
Snowflake Arctic \cite{snowflake} employs a similar design, utilizing a Dense-MoE hybrid transformer architecture to overlap communication with computation.

\subsection{Storage}
The ever-increasing parameters in MoE models exacerbate the constraints posed by memory capacity in compute devices, a challenge already pronounced in dense models. 
While expert parallelism offers a mitigation strategy through the distribution of experts across multiple devices, individual devices may still struggle to accommodate numerous experts, particularly in inference contexts where device capacity—--such as that of edge devices (PCs, smartphones, IoTs)--—is inherently more restricted.

Considering the hierarchical storage pyramid, solutions like SE-MoE \cite{shen2022se}, Pre-gated MoE \cite{hwang2023pre}, and EdgeMoE \cite{yi2023edgemoe} selectively retain only essential non-expert parameters and the active expert parameters within the GPU's High-Bandwidth Memory (HBM), offloading inactive expert parameters to CPU memory or SSDs. 
These patterns incur additional overhead from data transfer across the storage hierarchy, thus they integrate expert selection forecasting and expert parameter prefetching techniques to overlap parameter access with computation.

In addition, MPipeMoE \cite{zhang2024mpmoe} introduces a strategy to reduce the memory overhead associated with activations and temporary buffers. This is achieved by sharing buffer for various partitions of tensors, while leveraging recomputation/communication and CPU offloading to recover the requisite activations in the backward pass.

\section{Applications of Mixture of Experts}\label{sec:app}
In the current landscape dominated by Transformer-based large language models (LLMs), the mixture of experts (MoE) paradigm offers a compelling method to significantly expand model capacity while avoiding a corresponding surge in computational demands during training and inference phases.
These models have been instrumental in enhancing the performance of LLMs across a spectrum of downstream tasks, with some applications achieving results that eclipse human performance \cite{fedus2022review,jiang2024mixtral,dai2024deepseekmoe}. 
Rumors suggest that the formidable GPT-4 may employ an MoE architecture with an array of $8\times220$B experts, trained on diverse datasets and tasks, and utilizing a 16-iteration inference process \footnote{\href{https://x.com/soumithchintala/status/1671267150101721090}{https://x.com/soumithchintala/status/1671267150101721090}}.
Given these, MoE has garnered widespread adoption across fields such as natural language processing, computer vision, recommender systems, and multimodal applications. 
The essence of these applications lies in leveraging conditional computation to significantly boost the number of model parameters, thereby augmenting model capacities with a fixed computational cost, or implementing dynamic expert selection through gating mechanisms for efficient multi-task learning. 
In the following, we explore several representative applications of MoE in various domains to provide an overall understanding of how MoE can be utilized to specific tasks.

\textbf{Natural Language Processing}. 
The integration of MoE architectures with LLMs has unlocked extraordinary capabilities in a range of natural language understanding (NLU) and generation (NLG) tasks, including machine translation \cite{shazeer2017outrageously,costa2022no}, open-domain question answering \cite{du2022glam,artetxe2021efficient}, code generation \cite{jiang2024mixtral,dai2024deepseekmoe,qwen1.5,wei2024skywork}, and mathematical problem-solving \cite{dai2024deepseekmoe,jiang2024mixtral,deepseekv2,qwen1.5}. 
The methods of integrating MoE into LLMs have been thoroughly discussed and analyzed in Section \ref{sec:algorithm} (algorithm design) and Section \ref{sec:system} (system design), and will not be reiterated in depth here.
Beyond augmenting LLM capabilities, MoE has been instrumental in enhancing LLM's safety while preserving its usability. A notable implementation is MoGU \cite{du2024mogu}, which leverages dynamic routing to balance the contribution between usable LLM and safe LLM.

\textbf{Computer Vision}. 
The great success of sparsely-gated Mixture of Experts networks (MoE) in NLP has inspired their application in computer vision. 
For example, Riquelme \textit{et al.} \cite{riquelme2021scaling} introduced Vision MoE (V-MoE), which incorporates a sparsely activated mixture of MLPs into selected ViT \cite{dosovitskiy2020image} blocks. 
In image recognition tasks, V-MoE rivals the performance of state-of-the-art networks while requiring substantially less computational power during inference. 
This demonstrates the potential of MoE to discern distinct image semantics through specialized experts. 
Hwang \textit{et al.} \cite{hwang2023tutel} develop Tutel, a scalable system design and implementation for MoE with dynamic parallelism and pipelining, which they demonstrate with SwinV2-MoE, built upon Swin Transformer V2 \cite{liu2021swin}. 
Moreover, Zhang \textit{et al.} \cite{zhang2023robust} explore adversarial robustness in CNN-based MoE models, proposing a novel router-expert alternating adversarial training framework called ADVMOE. 
In most recent work, Chowdhury \textit{et al.} \cite{chowdhury2023patch} introduce the concept of patch-level routing in MoE (pMoE) that segments each input image into $n$ patches (or tokens) and allocates $l$ patches ($l\ll n$) to each expert for processing through prioritized routing to enhance efficiency.

\textbf{Recommender System}. 
Recommender systems are quintessential in various large-scale applications where they are required to balance and optimize multiple objectives simultaneously \cite{zheng2022survey}. 
A prime example is in the domain of movie recommendations, where the aim is not only to suggest movie that align with users' immediate preferences but also to ensure subsequent user satisfaction for the selected movies \cite{ma2018modeling}.
The effectiveness of multi-task models hinges on the intricate interplay between task-specific goals and the relationships between tasks. Consequently, understanding the trade-offs inherent in these relationships is crucial.
Mixture-of-experts (MoE) models with gating mechanisms have emerged as a popular paradigm for tackling the complexities of multi-task learning in recommender systems. 
Ma \textit{et al.} \cite{ma2018modeling} introduce the multi-gate mixture-of-experts (MMOE) approach, which capitalizes on the concept of shared expert submodels across all tasks, guided by a gating network tailored to each individual task.
Addressing the ``seesaw phenomenon" where the improvement of one task's performance can detrimentally affect another is another challenge in multi-task learning. 
To counteract this, Tang \textit{et al.} \cite{tang2020progressive} propose the Progressive Layered Extraction (PLE) model for personalized recommendations. PLE distinctly segregates shared and task-specific components and employs a progressive routing mechanism to incrementally extract and refine the semantic knowledge, thereby enhancing the efficacy of joint representation learning and the routing of information across tasks.
Recently, in the pursuit of capturing both the long-term and short-term user preferences that are particularly salient in sequential recommendation scenarios, a novel method named AdaMCT \cite{jiang2023adamct} has been proposed. AdaMCT utilizes layer-aware adaptive mixture units to dynamically blend CNN and Transformer experts, thereby tailoring the recommendations to individual user patterns.
Zhang et al. \cite{zhang2024m3oe} present M$3$oE, an adaptive Multi-domain Multi-task Mixture-of-Experts framework designed for improving multi-domain multi-task recommendation. 
This framework incorporates a shared expert module, a domain expert module, and a task expert module to address the common information learning, domain-aspect user preferences, and task-aspect user preferences, respectively. 
Moreover, it employs a two-level fusion mechanism, powered by AutoML, ensuring precise control over feature extraction and fusion across diverse domains and tasks.


\textbf{Multimodal Applications}. 
Multimodal models are designed to process and integrate various data types within a single neural network framework \cite{ngiam2011multimodal}. These models often simultaneously encompass two primary data modalities: images and text \cite{baltruvsaitis2018multimodal,uppal2022multimodal,zhou2020unified}. 
The Mixture of Experts (MoE) architecture has gained considerable traction as the foundation of multimodal models due to its capacity for expert layers to learn distinct modality partitioning \cite{mustafa2022multimodal}.
One notable implementation of this approach is the LIMoE model \cite{mustafa2022multimodal}, a sparse mixture of expert models tailored for multimodal learning. LIMoE is trained on both images and text data, employing contrastive loss and an entropy-based regularization technique to address load balancing challenges inherent in MoE systems.
Subsequently, Shen \textit{et al.} \cite{shen2023scaling}, Li \textit{et al.} \cite{li2023pace} and Lin \textit{et al.} \cite{lin2024moe} have further investigated the potential of MoE for scaling vision-language models, offering valuable insights that contribute to the development of more efficient and effective multimodal learning systems.
Furthermore, to address the specific issue of task conflicts in instruction tuning of Large Vision-Language Models (LVLMs), MoCLE \cite{gou2023mixture} integrates MoE with LoRA \cite{hu2021lora} experts and a distinct universal expert to activate task-specific model parameters based on clusters of instructions.
In parallel, to mitigate data conflicts, LLaVA-MoLE \cite{chen2024llava} deploys a set of LoRA experts, specifically for the MLP layer, combined with a top-1 gating mechanism to refine instruction tuning in Multimodal Large Language Models (MLLMs).
While the MLLMs employing MoE architectures have demonstrated impressive performances, they generally involve a limited number of experts and modalities \cite{li2024uni}. 
To address this limitation, Li \textit{et al.} \cite{li2024uni} introduce the pioneering Uni-MoE, a unified MLLM with MoE architecture capable of managing an extensive range of modalities. They introduce a progressive training strategy to bolster expert collaboration and generalization across modalities, and they utilize LoRA \cite{hu2021lora}, a lightweight fine-tuning methodology, to minimize computational demands.


%% file: section/5-challenges_opportunities.tex
\section{Challenges \& Opportunities}\label{sec:challenges}
Mixture of Experts (MoE) models present a compelling approach for significantly increasing model capacity at a constant computational cost. 
Despite their promise, several intrinsic challenges remain, necessitating further collaborative design and engineering across algorithm, system, and application aspects.
In this section, we identify critical challenges and promising directions for future investigation as follows:

\textbf{Training Stability and Load Balancing.} 
MoE models that utilize sparse gating have become a popular means to expand model capacity without proportionally increasing computational demands. 
However, the discrete nature of assigning a fixed number of experts to tokens leads to significant challenges in maintaining balanced workloads of experts and training stability across varying inputs \cite{shazeer2017outrageously,lepikhin2020gshard,fedus2022switch,zhou2022mixture,antoniak2023mixture}. 
Load imbalances, where certain experts become over-utilized while others are underutilized can hinder expert specialization and further degrade model performance. 
Although current efforts \cite{lepikhin2020gshard,fedus2022switch,du2022glam,jiang2024mixtral,lieber2024jamba,dai2024deepseekmoe,wei2024skywork} have attempted to address this challenge by incorporating auxiliary loss functions to encourage even token distribution across experts, these solutions can still lead to training instability \cite{zoph2022st} and often neglect the relative importance of different tokens \cite{zhou2022mixture}. 
Therefore, future studies should focus on more effective regularization techniques \cite{zoph2022st} or innovative gating algorithms \cite{zhou2022mixture,antoniak2023mixture,muqeeth2023soft,zhong2024lory} that encourage equitable load distribution among experts and enhance model training stability.

\textbf{Scalability and Communication Overhead.} 
As the escalating sizes of LLMs with MoE necessitate more expansive distributed systems, the imperative for efficient communication during model training becomes increasingly critical, as elaborated in Section \ref{sec:communication}.
The trade-off between model complexity, indicated by the number of parameters, and the communication overhead represents a significant bottleneck in distributed training processes \cite{lepikhin2020gshard}.
To address these challenges, it is essential to develop and implement effective strategies that enhance the efficiency of information transfer from system aspect or streamline information exchange without compromising model performance from algorithm aspect. 
Innovations such as DeepSpeed \cite{rajbhandari2022deepspeed}, FasterMoE \cite{he2022fastermoe}, and ScMoE \cite{cai2024shortcut} are at the forefront of minimizing communication overhead.
For example, the shared expert approach \cite{rajbhandari2022deepspeed,dai2024deepseekmoe,qwen_moe,snowflake}, advancing MoE with parameter-sharing frameworks, holds promise for reducing the volume of data transmitted between distributed systems while concurrently enhancing model performance in natural language processing tasks.
Such innovations are pivotal in facilitating more scalable and efficient distributed training architectures for MoE models.


\textbf{Expert Specialization and Collaboration.} 
Expert specialization refers to the concept where each expert develops non-overlapping and focused knowledge. 
Encouraging experts to concentrate their skills on distinct sub-tasks or domains has been shown to enhance the performance and generalization of the MoE model. 
The prevailing strategy involves designating a select number of experts as shared ones, with the goal of capturing commonalities in knowledge and reducing redundancy among those experts that are routed dynamically \cite{rajbhandari2022deepspeed,xue2024openmoe,qwen_moe,dai2024deepseekmoe}.
However, fostering effective collaboration among these specialized experts is an ongoing challenge. 
Relying solely on a sparsely computed weighted sum of outputs from the top-$k$ experts can overlook the intricate internal relationships that exist across the entire experts. 
Consequently, exploring new mechanisms for enhancing both the specialization and collaboration among experts is crucial for the development of more integrated and powerful MoE models.

\textbf{Sparse Activation and Computational Efficiency.} 
One of the primary benefits of MoE models lies in their capacity for sparse activations, which theoretically enhances computational efficiency. 
Nevertheless, implementing this efficiency in practice poses substantial challenges. This is attributed to the non-uniformity of sparse operations within hardware accelerators \cite{dao2022flashattention,kwon2023efficient}. 
Furthermore, optimizing the balance between activating a select top-$k$ subset of experts from an entire pool of experts entails intricate coordination. 
This optimization is crucial for ensuring that each expert develops a specialized niche \cite{dai2024deepseekmoe}.
Thus, there is a pressing need for further research into hardware optimization techniques that more adeptly accommodate sparse computations. Such advancements would not only preserve the model's capacity but could also significantly enhance the performance and efficiency of MoE models.

\textbf{Generalization and Robustness.} 
MoE models have demonstrated increased computational efficiency during pretraining phases. However, there is a notable propensity for sparse MoE architectures to overfit to specific tasks or datasets, which undermines their ability to generalize effectively \cite{fedus2022switch,zoph2022st,shen2023mixture,dou2023art}. 
To enhance the generalization and robustness of MoE models when encountering unseen data and diverse input variations, various strategies have been explored. 
These include regularization techniques such as dropout \cite{fedus2022switch} and token dropping \cite{zoph2022st}, as well as multi-task instruction tuning \cite{shen2023mixture,dou2023art}.
Looking ahead, there is potential for further advancements in this challenge. Future endeavors could explore innovative regularization methods, refined multi-task learning frameworks, or the incorporation of meta-learning concepts that bolster the MoE models' robustness and extend their generalization capabilities across an even broader spectrum of downstream tasks.

\textbf{Interpretability and Transparency.} 
The inherent complexity of MoE models, coupled with their dynamic gating of inputs to specialized experts, poses significant challenges to interpretability. 
This becomes particularly problematic in contexts where comprehending the rationale behind the model's decisions is essential. 
Enhancing the interpretability of MoE models is therefore critical, not only to facilitate a clearer understanding of their decision-making processes but also to address underlying challenges such as load balancing \cite{lepikhin2020gshard,fedus2022switch,du2022glam} and the mitigation of knowledge redundancy \cite{qwen_moe,dai2024deepseekmoe}.
In light of these considerations, there is a pressing need for future studies focused on the development of methods and tools that can effectively visualize and explain the behavior of individual experts within MoE models, as well as the nature of their interactions. 
Such advancements would significantly improve our grasp of MoE models and bolster their ongoing development, ensuring their gating decisions are transparent and trustworthy.

\textbf{Optimal Expert Architecture.}
The design of MoE architectures, encompassing the selection of network types and the quantity of experts, significantly influences the efficacy of multi-task learning across various domains. 
A plethora of network architectures has been adopted as experts, including LSTM \cite{shazeer2017outrageously}, CNN \cite{chowdhury2023patch,zhang2023robust}, FFNs (MLPs) \cite{lepikhin2020gshard,fedus2022switch,zoph2022st,pan2024dense}, Attention \cite{zhang2022mixture,shen2023moduleformer}, and LoRA \cite{dou2023loramoe,li2024mixlora,luo2024moelora}. Among these, FFNs as experts remain the most prevalent.
Despite their considerable achievements, the exploration of various hybrids of network types within experts (as the distinct features processing capabilities of different network architectures), as well as the development of innovative expert architectures, remains nascent areas of research.
Furthermore, the strategic allocation of a varying number of experts across different layers of the model presents an area ripe for investigation. This is due to two primary considerations: 1) different layers of the model capture semantic information at varying levels of granularity; 2) an excessive number of experts can complicate the training process and augment computational costs, while an insufficient number of experts might lead to knowledge redundancy and diminish the specialization capabilities of the experts.
To navigate these challenges, the development of automated architecture search methods specifically designed for MoE models is imperative \cite{zhou2023brainformers}. 
Such approaches could systematically identify optimal configurations, balancing the trade-offs between computational efficiency and the specialization of experts.

\textbf{Integration with Existing Frameworks.} 
Ensuring seamless integration of MoE models into existing large language models (LLMs) is crucial for their broad adoption. 
It is particularly vital to enable adaptation of LLMs to MoE architecture without necessitating training from scratch, as it can significantly reduce resource consumption. 
Recent studies \cite{dou2023loramoe,chen2024llava,li2024mixlora,luo2024moelora,zadouri2023pushing,gao2024higher,wu2024mixture} have demonstrated the efficacy of combining Parameter-efficient Fine-tuning (PEFT) techniques with MoE frameworks, offering a promising method for incorporating MoE into established LLMs. 
However, these methods may compromise model performance or complicate the existing parallel strategies of pretraining and inference efforts \cite{han2024parameter}.
Advancing the development of modular and plug-and-play MoE components is essential. 
Additionally, optimizing these components for training and deployment across diverse computing environments and hardware platforms will expand their applicability. 
Such advancements are expected to enhance the versatility and efficiency of MoE models, making them more accessible for a wide range of applications and platforms.

By addressing these challenges, we can unlock the full potential of MoE models, paving the way for more efficient and powerful machine learning systems, particular for large language models (LLMs), that are capable of handling the ever-growing complexity and diversity of real-world tasks.

%% file: section/6-conclusion.tex
\section{Conclusion}\label{sec:discussion}
In this survey, we present a systematic and comprehensive review of the literature on MoE models, serving as a valuable compendium for researchers exploring the landscape of MoE technologies.
We introduce a new taxonomy for MoE models and provide an in-depth analysis that encompasses three distinct vantage points: algorithm design, system design, and practical applications, complemented by a curated collection of open-source implementations, detailed hyperparameter configurations, and thorough empirical assessments.
Moreover, we highlight the critical challenges faced in the field and outline the most promising avenues for future investigation. 
To support the continuous dissemination of knowledge and advancements, we have established a dedicated resource repository to facilitate ongoing updates and the sharing of cutting-edge developments in MoE research.
We hope this survey can contribute to an essential reference for researchers seeking to rapidly acquaint themselves with MoE models, and that it will actively contribute to the vibrant progression.

%% file: main.bbl
\begin{thebibliography}{100}
\providecommand{\url}[1]{#1}
\csname url@samestyle\endcsname
\providecommand{\newblock}{\relax}
\providecommand{\bibinfo}[2]{#2}
\providecommand{\BIBentrySTDinterwordspacing}{\spaceskip=0pt\relax}
\providecommand{\BIBentryALTinterwordstretchfactor}{4}
\providecommand{\BIBentryALTinterwordspacing}{\spaceskip=\fontdimen2\font plus
\BIBentryALTinterwordstretchfactor\fontdimen3\font minus \fontdimen4\font\relax}
\providecommand{\BIBforeignlanguage}[2]{{%
\expandafter\ifx\csname l@#1\endcsname\relax
\typeout{** WARNING: IEEEtran.bst: No hyphenation pattern has been}%
\typeout{** loaded for the language `#1'. Using the pattern for}%
\typeout{** the default language instead.}%
\else
\language=\csname l@#1\endcsname
\fi
#2}}
\providecommand{\BIBdecl}{\relax}
\BIBdecl

\bibitem{vaswani2017attention}
A.~Vaswani, N.~Shazeer, N.~Parmar, J.~Uszkoreit, L.~Jones, A.~N. Gomez, {\L}.~Kaiser, and I.~Polosukhin, ``{Attention is all you need},'' \emph{Advances in neural information processing systems}, vol.~30, 2017.

\bibitem{brown2020language}
T.~Brown, B.~Mann, N.~Ryder, M.~Subbiah, J.~D. Kaplan, P.~Dhariwal, A.~Neelakantan, P.~Shyam, G.~Sastry, A.~Askell \emph{et~al.}, ``{Language models are few-shot learners},'' \emph{Advances in neural information processing systems}, vol.~33, pp. 1877--1901, 2020.

\bibitem{chowdhery2023palm}
A.~Chowdhery, S.~Narang, J.~Devlin, M.~Bosma, G.~Mishra, A.~Roberts, P.~Barham, H.~W. Chung, C.~Sutton, S.~Gehrmann \emph{et~al.}, ``{Palm: Scaling language modeling with pathways},'' \emph{Journal of Machine Learning Research}, vol.~24, no. 240, pp. 1--113, 2023.

\bibitem{achiam2023gpt}
J.~Achiam, S.~Adler, S.~Agarwal, L.~Ahmad, I.~Akkaya, F.~L. Aleman, D.~Almeida, J.~Altenschmidt, S.~Altman, S.~Anadkat \emph{et~al.}, ``{Gpt-4 technical report},'' \emph{arXiv preprint arXiv:2303.08774}, 2023.

\bibitem{jiang2024survey}
J.~Jiang, F.~Wang, J.~Shen, S.~Kim, and S.~Kim, ``{A Survey on Large Language Models for Code Generation},'' \emph{arXiv preprint arXiv:2406.00515}, 2024.

\bibitem{riquelme2021scaling}
C.~Riquelme, J.~Puigcerver, B.~Mustafa, M.~Neumann, R.~Jenatton, A.~Susano~Pinto, D.~Keysers, and N.~Houlsby, ``{Scaling vision with sparse mixture of experts},'' \emph{Advances in Neural Information Processing Systems}, vol.~34, pp. 8583--8595, 2021.

\bibitem{liu2021swin}
Z.~Liu, Y.~Lin, Y.~Cao, H.~Hu, Y.~Wei, Z.~Zhang, S.~Lin, and B.~Guo, ``{Swin transformer: Hierarchical vision transformer using shifted windows},'' in \emph{Proceedings of the IEEE/CVF international conference on computer vision}, 2021, pp. 10\,012--10\,022.

\bibitem{lu2019vilbert}
J.~Lu, D.~Batra, D.~Parikh, and S.~Lee, ``{Vilbert: Pretraining task-agnostic visiolinguistic representations for vision-and-language tasks},'' \emph{Advances in neural information processing systems}, vol.~32, 2019.

\bibitem{zhou2022learning}
K.~Zhou, J.~Yang, C.~C. Loy, and Z.~Liu, ``{Learning to prompt for vision-language models},'' \emph{International Journal of Computer Vision}, vol. 130, no.~9, pp. 2337--2348, 2022.

\bibitem{zhu2023minigpt}
D.~Zhu, J.~Chen, X.~Shen, X.~Li, and M.~Elhoseiny, ``{Minigpt-4: Enhancing vision-language understanding with advanced large language models},'' \emph{arXiv preprint arXiv:2304.10592}, 2023.

\bibitem{kaplan2020scaling}
J.~Kaplan, S.~McCandlish, T.~Henighan, T.~B. Brown, B.~Chess, R.~Child, S.~Gray, A.~Radford, J.~Wu, and D.~Amodei, ``{Scaling laws for neural language models},'' \emph{arXiv preprint arXiv:2001.08361}, 2020.

\bibitem{wei2022emergent}
J.~Wei, Y.~Tay, R.~Bommasani, C.~Raffel, B.~Zoph, S.~Borgeaud, D.~Yogatama, M.~Bosma, D.~Zhou, D.~Metzler \emph{et~al.}, ``{Emergent abilities of large language models},'' \emph{arXiv preprint arXiv:2206.07682}, 2022.

\bibitem{yoo2024hyperclova}
K.~M. Yoo, J.~Han, S.~In, H.~Jeon, J.~Jeong, J.~Kang, H.~Kim, K.-M. Kim, M.~Kim, S.~Kim \emph{et~al.}, ``{HyperCLOVA X Technical Report},'' \emph{arXiv preprint arXiv:2404.01954}, 2024.

\bibitem{hoffmann2022training}
J.~Hoffmann, S.~Borgeaud, A.~Mensch, E.~Buchatskaya, T.~Cai, E.~Rutherford, D.~d.~L. Casas, L.~A. Hendricks, J.~Welbl, A.~Clark \emph{et~al.}, ``{Training compute-optimal large language models},'' \emph{arXiv preprint arXiv:2203.15556}, 2022.

\bibitem{jacobs1991adaptive}
R.~A. Jacobs, M.~I. Jordan, S.~J. Nowlan, and G.~E. Hinton, ``{Adaptive mixtures of local experts},'' \emph{Neural computation}, vol.~3, no.~1, pp. 79--87, 1991.

\bibitem{jordan1994hierarchical}
M.~I. Jordan and R.~A. Jacobs, ``{Hierarchical mixtures of experts and the EM algorithm},'' \emph{Neural computation}, vol.~6, no.~2, pp. 181--214, 1994.

\bibitem{collobert2001parallel}
R.~Collobert, S.~Bengio, and Y.~Bengio, ``{A parallel mixture of SVMs for very large scale problems},'' \emph{Advances in Neural Information Processing Systems}, vol.~14, 2001.

\bibitem{rasmussen2001infinite}
C.~Rasmussen and Z.~Ghahramani, ``{Infinite mixtures of Gaussian process experts},'' \emph{Advances in neural information processing systems}, vol.~14, 2001.

\bibitem{shahbaba2009nonlinear}
B.~Shahbaba and R.~Neal, ``{Nonlinear models using Dirichlet process mixtures.}'' \emph{Journal of Machine Learning Research}, vol.~10, no.~8, 2009.

\bibitem{eigen2013learning}
D.~Eigen, M.~Ranzato, and I.~Sutskever, ``{Learning factored representations in a deep mixture of experts},'' \emph{arXiv preprint arXiv:1312.4314}, 2013.

\bibitem{theis2015generative}
L.~Theis and M.~Bethge, ``{Generative image modeling using spatial lstms},'' \emph{Advances in neural information processing systems}, vol.~28, 2015.

\bibitem{deisenroth2015distributed}
M.~Deisenroth and J.~W. Ng, ``{Distributed gaussian processes},'' in \emph{International conference on machine learning}.\hskip 1em plus 0.5em minus 0.4em\relax PMLR, 2015, pp. 1481--1490.

\bibitem{aljundi2017expert}
R.~Aljundi, P.~Chakravarty, and T.~Tuytelaars, ``{Expert gate: Lifelong learning with a network of experts},'' in \emph{Proceedings of the IEEE conference on computer vision and pattern recognition}, 2017, pp. 3366--3375.

\bibitem{shazeer2017outrageously}
N.~Shazeer, A.~Mirhoseini, K.~Maziarz, A.~Davis, Q.~Le, G.~Hinton, and J.~Dean, ``{Outrageously large neural networks: The sparsely-gated mixture-of-experts layer},'' \emph{arXiv preprint arXiv:1701.06538}, 2017.

\bibitem{lepikhin2020gshard}
D.~Lepikhin, H.~Lee, Y.~Xu, D.~Chen, O.~Firat, Y.~Huang, M.~Krikun, N.~Shazeer, and Z.~Chen, ``{Gshard: Scaling giant models with conditional computation and automatic sharding},'' \emph{arXiv preprint arXiv:2006.16668}, 2020.

\bibitem{jiang2024mixtral}
A.~Q. Jiang, A.~Sablayrolles, A.~Roux, A.~Mensch, B.~Savary, C.~Bamford, D.~S. Chaplot, D.~d.~l. Casas, E.~B. Hanna, F.~Bressand \emph{et~al.}, ``{Mixtral of experts},'' \emph{arXiv preprint arXiv:2401.04088}, 2024.

\bibitem{Grok-1}
\BIBentryALTinterwordspacing
xAI, ``{Grok-1},'' March 2024. [Online]. Available: \url{https://github.com/xai-org/grok-1}
\BIBentrySTDinterwordspacing

\bibitem{dbrx}
\BIBentryALTinterwordspacing
Databricks, ``{Introducing DBRX: A New State-of-the-Art Open LLM},'' March 2024. [Online]. Available: \url{https://www.databricks.com/blog/introducing-dbrx-new-state-art-open-llm}
\BIBentrySTDinterwordspacing

\bibitem{snowflake}
\BIBentryALTinterwordspacing
S.~A.~R. Team, ``{Snowflake Arctic: The Best LLM for Enterprise AI — Efficiently Intelligent, Truly Open},'' April 2024. [Online]. Available: \url{https://www.snowflake.com/blog/arctic-open-efficient-foundation-language-models-snowflake/}
\BIBentrySTDinterwordspacing

\bibitem{deepseekv2}
DeepSeek-AI, ``{DeepSeek-V2: A Strong, Economical, and Efficient Mixture-of-Experts Language Model},'' 2024.

\bibitem{yuksel2012twenty}
S.~E. Yuksel, J.~N. Wilson, and P.~D. Gader, ``Twenty years of mixture of experts,'' \emph{IEEE transactions on neural networks and learning systems}, vol.~23, no.~8, pp. 1177--1193, 2012.

\bibitem{fedus2022review}
W.~Fedus, J.~Dean, and B.~Zoph, ``{A review of sparse expert models in deep learning},'' \emph{arXiv preprint arXiv:2209.01667}, 2022.

\bibitem{du2022glam}
N.~Du, Y.~Huang, A.~M. Dai, S.~Tong, D.~Lepikhin, Y.~Xu, M.~Krikun, Y.~Zhou, A.~W. Yu, O.~Firat \emph{et~al.}, ``{Glam: Efficient scaling of language models with mixture-of-experts},'' in \emph{International Conference on Machine Learning}.\hskip 1em plus 0.5em minus 0.4em\relax PMLR, 2022, pp. 5547--5569.

\bibitem{fedus2022switch}
W.~Fedus, B.~Zoph, and N.~Shazeer, ``{Switch transformers: Scaling to trillion parameter models with simple and efficient sparsity},'' \emph{Journal of Machine Learning Research}, vol.~23, no. 120, pp. 1--39, 2022.

\bibitem{zoph2022st}
B.~Zoph, I.~Bello, S.~Kumar, N.~Du, Y.~Huang, J.~Dean, N.~Shazeer, and W.~Fedus, ``{St-moe: Designing stable and transferable sparse expert models},'' \emph{arXiv preprint arXiv:2202.08906}, 2022.

\bibitem{xue2024openmoe}
F.~Xue, Z.~Zheng, Y.~Fu, J.~Ni, Z.~Zheng, W.~Zhou, and Y.~You, ``{Openmoe: An early effort on open mixture-of-experts language models},'' \emph{arXiv preprint arXiv:2402.01739}, 2024.

\bibitem{puigcerver2023sparse}
J.~Puigcerver, C.~R. Ruiz, B.~Mustafa, and N.~Houlsby, ``{From Sparse to Soft Mixtures of Experts},'' in \emph{The Twelfth International Conference on Learning Representations}, 2023.

\bibitem{muqeeth2023soft}
M.~Muqeeth, H.~Liu, and C.~Raffel, ``{Soft merging of experts with adaptive routing},'' \emph{arXiv preprint arXiv:2306.03745}, 2023.

\bibitem{zhong2024lory}
Z.~Zhong, M.~Xia, D.~Chen, and M.~Lewis, ``{Lory: Fully Differentiable Mixture-of-Experts for Autoregressive Language Model Pre-training},'' \emph{arXiv preprint arXiv:2405.03133}, 2024.

\bibitem{zadouri2023pushing}
T.~Zadouri, A.~{\"U}st{\"u}n, A.~Ahmadian, B.~Ermi{\c{s}}, A.~Locatelli, and S.~Hooker, ``{Pushing mixture of experts to the limit: Extremely parameter efficient moe for instruction tuning},'' \emph{arXiv preprint arXiv:2309.05444}, 2023.

\bibitem{wu2023omni}
J.~Wu, X.~Hu, Y.~Wang, B.~Pang, and R.~Soricut, ``{Omni-SMoLA: Boosting Generalist Multimodal Models with Soft Mixture of Low-rank Experts},'' \emph{arXiv preprint arXiv:2312.00968}, 2023.

\bibitem{wang-etal-2022-adamix}
\BIBentryALTinterwordspacing
Y.~Wang, S.~Agarwal, S.~Mukherjee, X.~Liu, J.~Gao, A.~H. Awadallah, and J.~Gao, ``{{A}da{M}ix: Mixture-of-Adaptations for Parameter-efficient Model Tuning},'' in \emph{Proceedings of the 2022 Conference on Empirical Methods in Natural Language Processing}, Y.~Goldberg, Z.~Kozareva, and Y.~Zhang, Eds.\hskip 1em plus 0.5em minus 0.4em\relax Abu Dhabi, United Arab Emirates: Association for Computational Linguistics, Dec. 2022, pp. 5744--5760. [Online]. Available: \url{https://aclanthology.org/2022.emnlp-main.388}
\BIBentrySTDinterwordspacing

\bibitem{dou2023loramoe}
S.~Dou, E.~Zhou, Y.~Liu, S.~Gao, J.~Zhao, W.~Shen, Y.~Zhou, Z.~Xi, X.~Wang, X.~Fan \emph{et~al.}, ``{Loramoe: Revolutionizing mixture of experts for maintaining world knowledge in language model alignment},'' \emph{arXiv preprint arXiv:2312.09979}, 2023.

\bibitem{gou2023mixture}
Y.~Gou, Z.~Liu, K.~Chen, L.~Hong, H.~Xu, A.~Li, D.-Y. Yeung, J.~T. Kwok, and Y.~Zhang, ``{Mixture of cluster-conditional lora experts for vision-language instruction tuning},'' \emph{arXiv preprint arXiv:2312.12379}, 2023.

\bibitem{luo2024moelora}
T.~Luo, J.~Lei, F.~Lei, W.~Liu, S.~He, J.~Zhao, and K.~Liu, ``{Moelora: Contrastive learning guided mixture of experts on parameter-efficient fine-tuning for large language models},'' \emph{arXiv preprint arXiv:2402.12851}, 2024.

\bibitem{wu2024mixture}
\BIBentryALTinterwordspacing
X.~Wu, S.~Huang, and F.~Wei, ``{Mixture of Lo{RA} Experts},'' in \emph{The Twelfth International Conference on Learning Representations}, 2024. [Online]. Available: \url{https://openreview.net/forum?id=uWvKBCYh4S}
\BIBentrySTDinterwordspacing

\bibitem{komatsuzaki2022sparse}
A.~Komatsuzaki, J.~Puigcerver, J.~Lee-Thorp, C.~R. Ruiz, B.~Mustafa, J.~Ainslie, Y.~Tay, M.~Dehghani, and N.~Houlsby, ``{Sparse Upcycling: Training Mixture-of-Experts from Dense Checkpoints},'' in \emph{The Eleventh International Conference on Learning Representations}, 2022.

\bibitem{zhang2022moefication}
Z.~Zhang, Y.~Lin, Z.~Liu, P.~Li, M.~Sun, and J.~Zhou, ``{MoEfication: Transformer Feed-forward Layers are Mixtures of Experts},'' in \emph{Findings of the Association for Computational Linguistics: ACL 2022}, 2022, pp. 877--890.

\bibitem{llama-moe-2023}
\BIBentryALTinterwordspacing
L.-M. Team, ``{LLaMA-MoE: Building Mixture-of-Experts from LLaMA with Continual Pre-training},'' Dec 2023. [Online]. Available: \url{https://github.com/pjlab-sys4nlp/llama-moe}
\BIBentrySTDinterwordspacing

\bibitem{xue2022one}
F.~Xue, X.~He, X.~Ren, Y.~Lou, and Y.~You, ``{One student knows all experts know: From sparse to dense},'' \emph{arXiv preprint arXiv:2201.10890}, 2022.

\bibitem{chen2022task}
T.~Chen, S.~Huang, Y.~Xie, B.~Jiao, D.~Jiang, H.~Zhou, J.~Li, and F.~Wei, ``{Task-specific expert pruning for sparse mixture-of-experts},'' \emph{arXiv preprint arXiv:2206.00277}, 2022.

\bibitem{sukhbaatar2024branch}
S.~Sukhbaatar, O.~Golovneva, V.~Sharma, H.~Xu, X.~V. Lin, B.~Rozi{\`e}re, J.~Kahn, D.~Li, W.-t. Yih, J.~Weston \emph{et~al.}, ``{Branch-Train-MiX: Mixing Expert LLMs into a Mixture-of-Experts LLM},'' \emph{arXiv preprint arXiv:2403.07816}, 2024.

\bibitem{chen2023lifelong}
W.~Chen, Y.~Zhou, N.~Du, Y.~Huang, J.~Laudon, Z.~Chen, and C.~Cui, ``{Lifelong language pretraining with distribution-specialized experts},'' in \emph{International Conference on Machine Learning}.\hskip 1em plus 0.5em minus 0.4em\relax PMLR, 2023, pp. 5383--5395.

\bibitem{antoniak2023mixture}
S.~Antoniak, S.~Jaszczur, M.~Krutul, M.~Pi{\'o}ro, J.~Krajewski, J.~Ludziejewski, T.~Odrzyg{\'o}{\'z}d{\'z}, and M.~Cygan, ``{Mixture of Tokens: Efficient LLMs through Cross-Example Aggregation},'' \emph{arXiv preprint arXiv:2310.15961}, 2023.

\bibitem{raposo2024mixture}
D.~Raposo, S.~Ritter, B.~Richards, T.~Lillicrap, P.~C. Humphreys, and A.~Santoro, ``{Mixture-of-Depths: Dynamically allocating compute in transformer-based language models},'' \emph{arXiv preprint arXiv:2404.02258}, 2024.

\bibitem{xue2022go}
F.~Xue, Z.~Shi, F.~Wei, Y.~Lou, Y.~Liu, and Y.~You, ``{Go wider instead of deeper},'' in \emph{Proceedings of the AAAI Conference on Artificial Intelligence}, vol.~36, no.~8, 2022, pp. 8779--8787.

\bibitem{tan2023sparse}
S.~Tan, Y.~Shen, Z.~Chen, A.~Courville, and C.~Gan, ``{Sparse Universal Transformer},'' in \emph{Proceedings of the 2023 Conference on Empirical Methods in Natural Language Processing}, 2023, pp. 169--179.

\bibitem{choi2023smop}
J.-Y. Choi, J.~Kim, J.-H. Park, W.-L. Mok, and S.~Lee, ``{SMoP: Towards Efficient and Effective Prompt Tuning with Sparse Mixture-of-Prompts},'' in \emph{The 2023 Conference on Empirical Methods in Natural Language Processing}, 2023.

\bibitem{ma2018modeling}
J.~Ma, Z.~Zhao, X.~Yi, J.~Chen, L.~Hong, and E.~H. Chi, ``{Modeling task relationships in multi-task learning with multi-gate mixture-of-experts},'' in \emph{Proceedings of the 24th ACM SIGKDD international conference on knowledge discovery \& data mining}, 2018, pp. 1930--1939.

\bibitem{nie2021evomoe}
X.~Nie, X.~Miao, S.~Cao, L.~Ma, Q.~Liu, J.~Xue, Y.~Miao, Y.~Liu, Z.~Yang, and B.~Cui, ``{Evomoe: An evolutional mixture-of-experts training framework via dense-to-sparse gate},'' \emph{arXiv preprint arXiv:2112.14397}, 2021.

\bibitem{wu2023mole}
X.~Wu, S.~Huang, and F.~Wei, ``{MoLE: Mixture of LoRA Experts},'' in \emph{The Twelfth International Conference on Learning Representations}, 2023.

\bibitem{pan2024dense}
B.~Pan, Y.~Shen, H.~Liu, M.~Mishra, G.~Zhang, A.~Oliva, C.~Raffel, and R.~Panda, ``{Dense Training, Sparse Inference: Rethinking Training of Mixture-of-Experts Language Models},'' \emph{arXiv preprint arXiv:2404.05567}, 2024.

\bibitem{clark2022unified}
A.~Clark, D.~de~Las~Casas, A.~Guy, A.~Mensch, M.~Paganini, J.~Hoffmann, B.~Damoc, B.~Hechtman, T.~Cai, S.~Borgeaud \emph{et~al.}, ``{Unified scaling laws for routed language models},'' in \emph{International conference on machine learning}.\hskip 1em plus 0.5em minus 0.4em\relax PMLR, 2022, pp. 4057--4086.

\bibitem{rajbhandari2022deepspeed}
S.~Rajbhandari, C.~Li, Z.~Yao, M.~Zhang, R.~Y. Aminabadi, A.~A. Awan, J.~Rasley, and Y.~He, ``{Deepspeed-moe: Advancing mixture-of-experts inference and training to power next-generation ai scale},'' in \emph{International conference on machine learning}.\hskip 1em plus 0.5em minus 0.4em\relax PMLR, 2022, pp. 18\,332--18\,346.

\bibitem{wei2024skywork}
T.~Wei, B.~Zhu, L.~Zhao, C.~Cheng, B.~Li, W.~L{\"u}, P.~Cheng, J.~Zhang, X.~Zhang, L.~Zeng \emph{et~al.}, ``{Skywork-MoE: A Deep Dive into Training Techniques for Mixture-of-Experts Language Models},'' \emph{arXiv preprint arXiv:2406.06563}, 2024.

\bibitem{lieber2024jamba}
O.~Lieber, B.~Lenz, H.~Bata, G.~Cohen, J.~Osin, I.~Dalmedigos, E.~Safahi, S.~Meirom, Y.~Belinkov, S.~Shalev-Shwartz \emph{et~al.}, ``{Jamba: A hybrid transformer-mamba language model},'' \emph{arXiv preprint arXiv:2403.19887}, 2024.

\bibitem{dai2024deepseekmoe}
D.~Dai, C.~Deng, C.~Zhao, R.~Xu, H.~Gao, D.~Chen, J.~Li, W.~Zeng, X.~Yu, Y.~Wu \emph{et~al.}, ``{Deepseekmoe: Towards ultimate expert specialization in mixture-of-experts language models},'' \emph{arXiv preprint arXiv:2401.06066}, 2024.

\bibitem{yang2021m6}
A.~Yang, J.~Lin, R.~Men, C.~Zhou, L.~Jiang, X.~Jia, A.~Wang, J.~Zhang, J.~Wang, Y.~Li \emph{et~al.}, ``{M6-t: Exploring sparse expert models and beyond},'' \emph{arXiv preprint arXiv:2105.15082}, 2021.

\bibitem{chen2023mod}
Z.~Chen, Y.~Shen, M.~Ding, Z.~Chen, H.~Zhao, E.~G. Learned-Miller, and C.~Gan, ``{Mod-squad: Designing mixtures of experts as modular multi-task learners},'' in \emph{Proceedings of the IEEE/CVF Conference on Computer Vision and Pattern Recognition}, 2023, pp. 11\,828--11\,837.

\bibitem{dai2022stablemoe}
D.~Dai, L.~Dong, S.~Ma, B.~Zheng, Z.~Sui, B.~Chang, and F.~Wei, ``{StableMoE: Stable Routing Strategy for Mixture of Experts},'' in \emph{Proceedings of the 60th Annual Meeting of the Association for Computational Linguistics (Volume 1: Long Papers)}, 2022, pp. 7085--7095.

\bibitem{shen2023moduleformer}
Y.~Shen, Z.~Zhang, T.~Cao, S.~Tan, Z.~Chen, and C.~Gan, ``{Moduleformer: Learning modular large language models from uncurated data},'' \emph{arXiv preprint arXiv:2306.04640}, 2023.

\bibitem{lewis2021base}
M.~Lewis, S.~Bhosale, T.~Dettmers, N.~Goyal, and L.~Zettlemoyer, ``{Base layers: Simplifying training of large, sparse models},'' in \emph{International Conference on Machine Learning}.\hskip 1em plus 0.5em minus 0.4em\relax PMLR, 2021, pp. 6265--6274.

\bibitem{hazimeh2021dselect}
H.~Hazimeh, Z.~Zhao, A.~Chowdhery, M.~Sathiamoorthy, Y.~Chen, R.~Mazumder, L.~Hong, and E.~Chi, ``{Dselect-k: Differentiable selection in the mixture of experts with applications to multi-task learning},'' \emph{Advances in Neural Information Processing Systems}, vol.~34, pp. 29\,335--29\,347, 2021.

\bibitem{kim2021scalable}
Y.~J. Kim, A.~A. Awan, A.~Muzio, A.~F.~C. Salinas, L.~Lu, A.~Hendy, S.~Rajbhandari, Y.~He, and H.~H. Awadalla, ``{Scalable and efficient moe training for multitask multilingual models},'' \emph{arXiv preprint arXiv:2109.10465}, 2021.

\bibitem{kudugunta2021beyond}
S.~Kudugunta, Y.~Huang, A.~Bapna, M.~Krikun, D.~Lepikhin, M.-T. Luong, and O.~Firat, ``{Beyond Distillation: Task-level Mixture-of-Experts for Efficient Inference},'' in \emph{Findings of the Association for Computational Linguistics: EMNLP 2021}, 2021, pp. 3577--3599.

\bibitem{costa2022no}
M.~R. Costa-juss{\`a}, J.~Cross, O.~{\c{C}}elebi, M.~Elbayad, K.~Heafield, K.~Heffernan, E.~Kalbassi, J.~Lam, D.~Licht, J.~Maillard \emph{et~al.}, ``{No language left behind: Scaling human-centered machine translation},'' \emph{arXiv preprint arXiv:2207.04672}, 2022.

\bibitem{ye2022eliciting}
Q.~Ye, J.~Zha, and X.~Ren, ``{Eliciting and Understanding Cross-task Skills with Task-level Mixture-of-Experts},'' in \emph{Findings of the Association for Computational Linguistics: EMNLP 2022}, 2022, pp. 2567--2592.

\bibitem{chi2022representation}
Z.~Chi, L.~Dong, S.~Huang, D.~Dai, S.~Ma, B.~Patra, S.~Singhal, P.~Bajaj, X.~Song, X.-L. Mao \emph{et~al.}, ``{On the representation collapse of sparse mixture of experts},'' \emph{Advances in Neural Information Processing Systems}, vol.~35, pp. 34\,600--34\,613, 2022.

\bibitem{zhu2022uni}
J.~Zhu, X.~Zhu, W.~Wang, X.~Wang, H.~Li, X.~Wang, and J.~Dai, ``{Uni-perceiver-moe: Learning sparse generalist models with conditional moes},'' \emph{Advances in Neural Information Processing Systems}, vol.~35, pp. 2664--2678, 2022.

\bibitem{zhang2022mixture}
X.~Zhang, Y.~Shen, Z.~Huang, J.~Zhou, W.~Rong, and Z.~Xiong, ``{Mixture of Attention Heads: Selecting Attention Heads Per Token},'' in \emph{Proceedings of the 2022 Conference on Empirical Methods in Natural Language Processing}, 2022, pp. 4150--4162.

\bibitem{shen2024jetmoe}
Y.~Shen, Z.~Guo, T.~Cai, and Z.~Qin, ``{JetMoE: Reaching Llama2 Performance with 0.1 M Dollars},'' \emph{arXiv preprint arXiv:2404.07413}, 2024.

\bibitem{wu2024yuan}
S.~Wu, J.~Luo, X.~Chen, L.~Li, X.~Zhao, T.~Yu, C.~Wang, Y.~Wang, F.~Wang, W.~Qiao \emph{et~al.}, ``{Yuan 2.0-M32: Mixture of Experts with Attention Router},'' \emph{arXiv preprint arXiv:2405.17976}, 2024.

\bibitem{zeng2024adamoe}
Z.~Zeng, Y.~Miao, H.~Gao, H.~Zhang, and Z.~Deng, ``Adamoe: Token-adaptive routing with null experts for mixture-of-experts language models,'' \emph{arXiv preprint arXiv:2406.13233}, 2024.

\bibitem{shi2024unchosen}
C.~Shi, C.~Yang, X.~Zhu, J.~Wang, T.~Wu, S.~Li, D.~Cai, Y.~Yang, and Y.~Meng, ``Unchosen experts can contribute too: Unleashing moe models' power by self-contrast,'' \emph{arXiv preprint arXiv:2405.14507}, 2024.

\bibitem{guo2024dynamic}
Y.~Guo, Z.~Cheng, X.~Tang, and T.~Lin, ``Dynamic mixture of experts: An auto-tuning approach for efficient transformer models,'' \emph{arXiv preprint arXiv:2405.14297}, 2024.

\bibitem{caiflextron}
R.~Cai, S.~Muralidharan, G.~Heinrich, H.~Yin, Z.~Wang, J.~Kautz, and P.~Molchanov, ``Flextron: Many-in-one flexible large language model,'' in \emph{Forty-first International Conference on Machine Learning}.

\bibitem{roller2021hash}
S.~Roller, S.~Sukhbaatar, J.~Weston \emph{et~al.}, ``{Hash layers for large sparse models},'' \emph{Advances in Neural Information Processing Systems}, vol.~34, pp. 17\,555--17\,566, 2021.

\bibitem{zuo2021taming}
S.~Zuo, X.~Liu, J.~Jiao, Y.~J. Kim, H.~Hassan, R.~Zhang, J.~Gao, and T.~Zhao, ``{Taming Sparsely Activated Transformer with Stochastic Experts},'' in \emph{International Conference on Learning Representations}, 2021.

\bibitem{gururangan2022demix}
S.~Gururangan, M.~Lewis, A.~Holtzman, N.~A. Smith, and L.~Zettlemoyer, ``{DEMix Layers: Disentangling Domains for Modular Language Modeling},'' in \emph{Proceedings of the 2022 Conference of the North American Chapter of the Association for Computational Linguistics: Human Language Technologies}, 2022, pp. 5557--5576.

\bibitem{fan2021beyond}
A.~Fan, S.~Bhosale, H.~Schwenk, Z.~Ma, A.~El-Kishky, S.~Goyal, M.~Baines, O.~Celebi, G.~Wenzek, V.~Chaudhary \emph{et~al.}, ``{Beyond english-centric multilingual machine translation},'' \emph{Journal of Machine Learning Research}, vol.~22, no. 107, pp. 1--48, 2021.

\bibitem{ren2023pangu}
X.~Ren, P.~Zhou, X.~Meng, X.~Huang, Y.~Wang, W.~Wang, P.~Li, X.~Zhang, A.~Podolskiy, G.~Arshinov \emph{et~al.}, ``{Pangu-{$\Sigma$}: Towards trillion parameter language model with sparse heterogeneous computing},'' \emph{arXiv preprint arXiv:2303.10845}, 2023.

\bibitem{zhou2022mixture}
Y.~Zhou, T.~Lei, H.~Liu, N.~Du, Y.~Huang, V.~Zhao, A.~M. Dai, Q.~V. Le, J.~Laudon \emph{et~al.}, ``{Mixture-of-experts with expert choice routing},'' \emph{Advances in Neural Information Processing Systems}, vol.~35, pp. 7103--7114, 2022.

\bibitem{zhou2023brainformers}
Y.~Zhou, N.~Du, Y.~Huang, D.~Peng, C.~Lan, D.~Huang, S.~Shakeri, D.~So, A.~M. Dai, Y.~Lu \emph{et~al.}, ``{Brainformers: Trading simplicity for efficiency},'' in \emph{International Conference on Machine Learning}.\hskip 1em plus 0.5em minus 0.4em\relax PMLR, 2023, pp. 42\,531--42\,542.

\bibitem{dat2023homoe}
D.~H. Dat, P.~Y. Mao, T.~H. Nguyen, W.~Buntine, and M.~Bennamoun, ``{HOMOE: A Memory-Based and Composition-Aware Framework for Zero-Shot Learning with Hopfield Network and Soft Mixture of Experts},'' \emph{arXiv preprint arXiv:2311.14747}, 2023.

\bibitem{he2023merging}
S.~He, R.-Z. Fan, L.~Ding, L.~Shen, T.~Zhou, and D.~Tao, ``Merging experts into one: Improving computational efficiency of mixture of experts,'' in \emph{Proceedings of the 2023 Conference on Empirical Methods in Natural Language Processing}, 2023, pp. 14\,685--14\,691.

\bibitem{zhang2023emergent}
Z.~Zhang, Z.~Zeng, Y.~Lin, C.~Xiao, X.~Wang, X.~Han, Z.~Liu, R.~Xie, M.~Sun, and J.~Zhou, ``Emergent modularity in pre-trained transformers,'' in \emph{Findings of the Association for Computational Linguistics: ACL 2023}, 2023, pp. 4066--4083.

\bibitem{chowdhury2023patch}
M.~N.~R. Chowdhury, S.~Zhang, M.~Wang, S.~Liu, and P.-Y. Chen, ``{Patch-level routing in mixture-of-experts is provably sample-efficient for convolutional neural networks},'' in \emph{International Conference on Machine Learning}.\hskip 1em plus 0.5em minus 0.4em\relax PMLR, 2023, pp. 6074--6114.

\bibitem{zhang2023robust}
Y.~Zhang, R.~Cai, T.~Chen, G.~Zhang, H.~Zhang, P.-Y. Chen, S.~Chang, Z.~Wang, and S.~Liu, ``{Robust Mixture-of-Expert Training for Convolutional Neural Networks},'' in \emph{Proceedings of the IEEE/CVF International Conference on Computer Vision}, 2023, pp. 90--101.

\bibitem{chen2022towards}
Z.~Chen, Y.~Deng, Y.~Wu, Q.~Gu, and Y.~Li, ``{Towards understanding mixture of experts in deep learning},'' \emph{arXiv preprint arXiv:2208.02813}, 2022.

\bibitem{wang2020deep}
X.~Wang, F.~Yu, L.~Dunlap, Y.-A. Ma, R.~Wang, A.~Mirhoseini, T.~Darrell, and J.~E. Gonzalez, ``{Deep mixture of experts via shallow embedding},'' in \emph{Uncertainty in artificial intelligence}.\hskip 1em plus 0.5em minus 0.4em\relax PMLR, 2020, pp. 552--562.

\bibitem{artetxe2021efficient}
M.~Artetxe, S.~Bhosale, N.~Goyal, T.~Mihaylov, M.~Ott, S.~Shleifer, X.~V. Lin, J.~Du, S.~Iyer, R.~Pasunuru \emph{et~al.}, ``{Efficient large scale language modeling with mixtures of experts},'' \emph{arXiv preprint arXiv:2112.10684}, 2021.

\bibitem{qwen_moe}
\BIBentryALTinterwordspacing
Q.~Team, ``{Qwen1.5-MoE: Matching 7B Model Performance with 1/3 Activated Parameters"},'' February 2024. [Online]. Available: \url{https://qwenlm.github.io/blog/qwen-moe/}
\BIBentrySTDinterwordspacing

\bibitem{he2024mixture}
X.~O. He, ``Mixture of a million experts,'' \emph{arXiv preprint arXiv:2407.04153}, 2024.

\bibitem{lin2024moe}
B.~Lin, Z.~Tang, Y.~Ye, J.~Cui, B.~Zhu, P.~Jin, J.~Zhang, M.~Ning, and L.~Yuan, ``{Moe-llava: Mixture of experts for large vision-language models},'' \emph{arXiv preprint arXiv:2401.15947}, 2024.

\bibitem{glorot2011deep}
X.~Glorot, A.~Bordes, and Y.~Bengio, ``{Deep sparse rectifier neural networks},'' in \emph{Proceedings of the fourteenth international conference on artificial intelligence and statistics}.\hskip 1em plus 0.5em minus 0.4em\relax JMLR Workshop and Conference Proceedings, 2011, pp. 315--323.

\bibitem{hendrycks2016gaussian}
D.~Hendrycks and K.~Gimpel, ``{Gaussian error linear units (gelus)},'' \emph{arXiv preprint arXiv:1606.08415}, 2016.

\bibitem{shazeer2020glu}
N.~Shazeer, ``{Glu variants improve transformer},'' \emph{arXiv preprint arXiv:2002.05202}, 2020.

\bibitem{cai2024shortcut}
W.~Cai, J.~Jiang, L.~Qin, J.~Cui, S.~Kim, and J.~Huang, ``{Shortcut-connected Expert Parallelism for Accelerating Mixture-of-Experts},'' \emph{arXiv preprint arXiv:2404.05019}, 2024.

\bibitem{he2023pad}
S.~He, L.~Ding, D.~Dong, B.~Liu, F.~Yu, and D.~Tao, ``Pad-net: An efficient framework for dynamic networks,'' in \emph{Proceedings of the 61st Annual Meeting of the Association for Computational Linguistics (Volume 1: Long Papers)}, 2023, pp. 14\,354--14\,366.

\bibitem{zhao2024hypermoe}
H.~Zhao, Z.~Qiu, H.~Wu, Z.~Wang, Z.~He, and J.~Fu, ``Hypermoe: Towards better mixture of experts via transferring among experts,'' \emph{arXiv preprint arXiv:2402.12656}, 2024.

\bibitem{diao2023mixture}
S.~Diao, T.~Xu, R.~Xu, J.~Wang, and T.~Zhang, ``{Mixture-of-Domain-Adapters: Decoupling and Injecting Domain Knowledge to Pre-trained Language Models' Memories},'' in \emph{The 61st Annual Meeting Of The Association For Computational Linguistics}, 2023.

\bibitem{li2024mixlora}
D.~Li, Y.~Ma, N.~Wang, Z.~Cheng, L.~Duan, J.~Zuo, C.~Yang, and M.~Tang, ``{MixLoRA: Enhancing Large Language Models Fine-Tuning with LoRA based Mixture of Experts},'' \emph{arXiv preprint arXiv:2404.15159}, 2024.

\bibitem{chen2024llava}
S.~Chen, Z.~Jie, and L.~Ma, ``{Llava-mole: Sparse mixture of lora experts for mitigating data conflicts in instruction finetuning mllms},'' \emph{arXiv preprint arXiv:2401.16160}, 2024.

\bibitem{zhu2023sira}
Y.~Zhu, N.~Wichers, C.-C. Lin, X.~Wang, T.~Chen, L.~Shu, H.~Lu, C.~Liu, L.~Luo, J.~Chen \emph{et~al.}, ``{Sira: Sparse mixture of low rank adaptation},'' \emph{arXiv preprint arXiv:2311.09179}, 2023.

\bibitem{mao2022unipelt}
Y.~Mao, L.~Mathias, R.~Hou, A.~Almahairi, H.~Ma, J.~Han, S.~Yih, and M.~Khabsa, ``{UniPELT: A Unified Framework for Parameter-Efficient Language Model Tuning},'' in \emph{Proceedings of the 60th Annual Meeting of the Association for Computational Linguistics (Volume 1: Long Papers)}, 2022, pp. 6253--6264.

\bibitem{gao2024higher}
C.~Gao, K.~Chen, J.~Rao, B.~Sun, R.~Liu, D.~Peng, Y.~Zhang, X.~Guo, J.~Yang, and V.~Subrahmanian, ``{Higher Layers Need More LoRA Experts},'' \emph{arXiv preprint arXiv:2402.08562}, 2024.

\bibitem{liu2023moelora}
Q.~Liu, X.~Wu, X.~Zhao, Y.~Zhu, D.~Xu, F.~Tian, and Y.~Zheng, ``{Moelora: An moe-based parameter efficient fine-tuning method for multi-task medical applications},'' \emph{arXiv preprint arXiv:2310.18339}, 2023.

\bibitem{liu2024intuition}
Y.~Liu, R.~Zhang, H.~Yang, K.~Keutzer, Y.~Du, L.~Du, and S.~Zhang, ``{Intuition-aware Mixture-of-Rank-1-Experts for Parameter Efficient Finetuning},'' \emph{arXiv preprint arXiv:2404.08985}, 2024.

\bibitem{xu2024meteora}
J.~Xu, J.~Lai, and Y.~Huang, ``Meteora: Multiple-tasks embedded lora for large language models,'' \emph{arXiv preprint arXiv:2405.13053}, 2024.

\bibitem{wu2022residual}
L.~Wu, M.~Liu, Y.~Chen, D.~Chen, X.~Dai, and L.~Yuan, ``{Residual mixture of experts},'' \emph{arXiv preprint arXiv:2204.09636}, 2022.

\bibitem{dua2022tricks}
D.~Dua, S.~Bhosale, V.~Goswami, J.~Cross, M.~Lewis, and A.~Fan, ``{Tricks for Training Sparse Translation Models},'' in \emph{Proceedings of the 2022 Conference of the North American Chapter of the Association for Computational Linguistics: Human Language Technologies}, 2022, pp. 3340--3345.

\bibitem{chen2022sparse}
T.~Chen, Z.~Zhang, A.~K. JAISWAL, S.~Liu, and Z.~Wang, ``{Sparse MoE as the New Dropout: Scaling Dense and Self-Slimmable Transformers},'' in \emph{The Eleventh International Conference on Learning Representations}, 2022.

\bibitem{zuo2022moebert}
S.~Zuo, Q.~Zhang, C.~Liang, P.~He, T.~Zhao, and W.~Chen, ``{MoEBERT: from BERT to Mixture-of-Experts via Importance-Guided Adaptation},'' in \emph{Proceedings of the 2022 Conference of the North American Chapter of the Association for Computational Linguistics: Human Language Technologies}, 2022, pp. 1610--1623.

\bibitem{qiu2024unlocking}
Z.~Qiu, Z.~Huang, and J.~Fu, ``Unlocking emergent modularity in large language models,'' in \emph{Proceedings of the 2024 Conference of the North American Chapter of the Association for Computational Linguistics: Human Language Technologies (Volume 1: Long Papers)}, 2024, pp. 2638--2660.

\bibitem{huang2023experts}
Y.~Huang, P.~Ye, X.~Huang, S.~Li, T.~Chen, and W.~Ouyang, ``{Experts weights averaging: A new general training scheme for vision transformers},'' \emph{arXiv preprint arXiv:2308.06093}, 2023.

\bibitem{he2024demystifying}
S.~He, D.~Dong, L.~Ding, and A.~Li, ``Demystifying the compression of mixture-of-experts through a unified framework,'' \emph{arXiv preprint arXiv:2406.02500}, 2024.

\bibitem{li2022branch}
M.~Li, S.~Gururangan, T.~Dettmers, M.~Lewis, T.~Althoff, N.~A. Smith, and L.~Zettlemoyer, ``{Branch-Train-Merge: Embarrassingly Parallel Training of Expert Language Models},'' in \emph{First Workshop on Interpolation Regularizers and Beyond at NeurIPS 2022}, 2022.

\bibitem{wang2023fusing}
H.~Wang, F.~M. Polo, Y.~Sun, S.~Kundu, E.~Xing, and M.~Yurochkin, ``{Fusing Models with Complementary Expertise},'' in \emph{The Twelfth International Conference on Learning Representations}, 2023.

\bibitem{he2021fastmoe}
J.~He, J.~Qiu, A.~Zeng, Z.~Yang, J.~Zhai, and J.~Tang, ``{Fastmoe: A fast mixture-of-expert training system},'' \emph{arXiv preprint arXiv:2103.13262}, 2021.

\bibitem{hwang2023tutel}
C.~Hwang, W.~Cui, Y.~Xiong, Z.~Yang, Z.~Liu, H.~Hu, Z.~Wang, R.~Salas, J.~Jose, P.~Ram \emph{et~al.}, ``{Tutel: Adaptive mixture-of-experts at scale},'' \emph{Proceedings of Machine Learning and Systems}, vol.~5, 2023.

\bibitem{shen2022se}
L.~Shen, Z.~Wu, W.~Gong, H.~Hao, Y.~Bai, H.~Wu, X.~Wu, J.~Bian, H.~Xiong, D.~Yu \emph{et~al.}, ``{Se-moe: A scalable and efficient mixture-of-experts distributed training and inference system},'' \emph{arXiv preprint arXiv:2205.10034}, 2022.

\bibitem{he2022fastermoe}
J.~He, J.~Zhai, T.~Antunes, H.~Wang, F.~Luo, S.~Shi, and Q.~Li, ``{Fastermoe: modeling and optimizing training of large-scale dynamic pre-trained models},'' in \emph{Proceedings of the 27th ACM SIGPLAN Symposium on Principles and Practice of Parallel Programming}, 2022, pp. 120--134.

\bibitem{singh2023hybrid}
S.~Singh, O.~Ruwase, A.~A. Awan, S.~Rajbhandari, Y.~He, and A.~Bhatele, ``{A Hybrid Tensor-Expert-Data Parallelism Approach to Optimize Mixture-of-Experts Training},'' in \emph{Proceedings of the 37th International Conference on Supercomputing}, 2023, pp. 203--214.

\bibitem{nie2022hetumoe}
X.~Nie, P.~Zhao, X.~Miao, T.~Zhao, and B.~Cui, ``{HetuMoE: An efficient trillion-scale mixture-of-expert distributed training system},'' \emph{arXiv preprint arXiv:2203.14685}, 2022.

\bibitem{nie2023flexmoe}
X.~Nie, X.~Miao, Z.~Wang, Z.~Yang, J.~Xue, L.~Ma, G.~Cao, and B.~Cui, ``{Flexmoe: Scaling large-scale sparse pre-trained model training via dynamic device placement},'' \emph{Proceedings of the ACM on Management of Data}, vol.~1, no.~1, pp. 1--19, 2023.

\bibitem{zhai2023smartmoe}
M.~Zhai, J.~He, Z.~Ma, Z.~Zong, R.~Zhang, and J.~Zhai, ``{$\{$SmartMoE$\}$: Efficiently Training $\{$Sparsely-Activated$\}$ Models through Combining Offline and Online Parallelization},'' in \emph{2023 USENIX Annual Technical Conference (USENIX ATC 23)}, 2023, pp. 961--975.

\bibitem{gale2023megablocks}
T.~Gale, D.~Narayanan, C.~Young, and M.~Zaharia, ``{Megablocks: Efficient sparse training with mixture-of-experts},'' \emph{Proceedings of Machine Learning and Systems}, vol.~5, 2023.

\bibitem{tan2024scattered}
S.~Tan, Y.~Shen, R.~Panda, and A.~Courville, ``{Scattered Mixture-of-Experts Implementation},'' \emph{arXiv preprint arXiv:2403.08245}, 2024.

\bibitem{zheng2023pit}
N.~Zheng, H.~Jiang, Q.~Zhang, Z.~Han, L.~Ma, Y.~Yang, F.~Yang, C.~Zhang, L.~Qiu, M.~Yang \emph{et~al.}, ``{Pit: Optimization of dynamic sparse deep learning models via permutation invariant transformation},'' in \emph{Proceedings of the 29th Symposium on Operating Systems Principles}, 2023, pp. 331--347.

\bibitem{yao2024exploiting}
J.~Yao, Q.~Anthony, A.~Shafi, H.~Subramoni \emph{et~al.}, ``{Exploiting Inter-Layer Expert Affinity for Accelerating Mixture-of-Experts Model Inference},'' \emph{arXiv preprint arXiv:2401.08383}, 2024.

\bibitem{chen2022ta}
C.~Chen, M.~Li, Z.~Wu, D.~Yu, and C.~Yang, ``{Ta-moe: Topology-aware large scale mixture-of-expert training},'' \emph{Advances in Neural Information Processing Systems}, vol.~35, pp. 22\,173--22\,186, 2022.

\bibitem{zhang2024mpmoe}
Z.~Zhang, Y.~Xia, H.~Wang, D.~Yang, C.~Hu, X.~Zhou, and D.~Cheng, ``{MPMoE: Memory Efficient MoE for Pre-trained Models with Adaptive Pipeline Parallelism},'' \emph{IEEE Transactions on Parallel and Distributed Systems}, 2024.

\bibitem{jiang2024lancet}
C.~Jiang, Y.~Tian, Z.~Jia, S.~Zheng, C.~Wu, and Y.~Wang, ``{Lancet: Accelerating Mixture-of-Experts Training via Whole Graph Computation-Communication Overlapping},'' \emph{arXiv preprint arXiv:2404.19429}, 2024.

\bibitem{shi2023pipemoe}
S.~Shi, X.~Pan, X.~Chu, and B.~Li, ``Pipemoe: Accelerating mixture-of-experts through adaptive pipelining,'' in \emph{IEEE INFOCOM 2023-IEEE Conference on Computer Communications}.\hskip 1em plus 0.5em minus 0.4em\relax IEEE, 2023, pp. 1--10.

\bibitem{shi2024schemoe}
S.~Shi, X.~Pan, Q.~Wang, C.~Liu, X.~Ren, Z.~Hu, Y.~Yang, B.~Li, and X.~Chu, ``Schemoe: An extensible mixture-of-experts distributed training system with tasks scheduling,'' in \emph{Proceedings of the Nineteenth European Conference on Computer Systems}, 2024, pp. 236--249.

\bibitem{punniyamurthy2023optimizing}
K.~Punniyamurthy, K.~Hamidouche, and B.~M. Beckmann, ``Optimizing distributed ml communication with fused computation-collective operations,'' \emph{arXiv preprint arXiv:2305.06942}, 2023.

\bibitem{hwang2023pre}
R.~Hwang, J.~Wei, S.~Cao, C.~Hwang, X.~Tang, T.~Cao, M.~Yang, and M.~Rhu, ``{Pre-gated MoE: An Algorithm-System Co-Design for Fast and Scalable Mixture-of-Expert Inference},'' \emph{arXiv preprint arXiv:2308.12066}, 2023.

\bibitem{yi2023edgemoe}
R.~Yi, L.~Guo, S.~Wei, A.~Zhou, S.~Wang, and M.~Xu, ``{Edgemoe: Fast on-device inference of moe-based large language models},'' \emph{arXiv preprint arXiv:2308.14352}, 2023.

\bibitem{du2024mogu}
Y.~Du, S.~Zhao, D.~Zhao, M.~Ma, Y.~Chen, L.~Huo, Q.~Yang, D.~Xu, and B.~Qin, ``Mogu: A framework for enhancing safety of open-sourced llms while preserving their usability,'' \emph{arXiv preprint arXiv:2405.14488}, 2024.

\bibitem{tang2020progressive}
H.~Tang, J.~Liu, M.~Zhao, and X.~Gong, ``{Progressive layered extraction (ple): A novel multi-task learning (mtl) model for personalized recommendations},'' in \emph{Proceedings of the 14th ACM Conference on Recommender Systems}, 2020, pp. 269--278.

\bibitem{jiang2023adamct}
J.~Jiang, P.~Zhang, Y.~Luo, C.~Li, J.~B. Kim, K.~Zhang, S.~Wang, X.~Xie, and S.~Kim, ``{AdaMCT: adaptive mixture of CNN-transformer for sequential recommendation},'' in \emph{Proceedings of the 32nd ACM International Conference on Information and Knowledge Management}, 2023, pp. 976--986.

\bibitem{zhang2024m3oe}
Z.~Zhang, S.~Liu, J.~Yu, Q.~Cai, X.~Zhao, C.~Zhang, Z.~Liu, Q.~Liu, H.~Zhao, L.~Hu \emph{et~al.}, ``M3oe: Multi-domain multi-task mixture-of experts recommendation framework,'' in \emph{Proceedings of the 47th International ACM SIGIR Conference on Research and Development in Information Retrieval}, 2024, pp. 893--902.

\bibitem{mustafa2022multimodal}
B.~Mustafa, C.~Riquelme, J.~Puigcerver, R.~Jenatton, and N.~Houlsby, ``{Multimodal contrastive learning with limoe: the language-image mixture of experts},'' \emph{Advances in Neural Information Processing Systems}, vol.~35, pp. 9564--9576, 2022.

\bibitem{shen2023scaling}
S.~Shen, Z.~Yao, C.~Li, T.~Darrell, K.~Keutzer, and Y.~He, ``{Scaling vision-language models with sparse mixture of experts},'' \emph{arXiv preprint arXiv:2303.07226}, 2023.

\bibitem{li2023pace}
Y.~Li, B.~Hui, Z.~Yin, M.~Yang, F.~Huang, and Y.~Li, ``Pace: Unified multi-modal dialogue pre-training with progressive and compositional experts,'' in \emph{Proceedings of the 61st Annual Meeting of the Association for Computational Linguistics (Volume 1: Long Papers)}, 2023, pp. 13\,402--13\,416.

\bibitem{li2024uni}
Y.~Li, S.~Jiang, B.~Hu, L.~Wang, W.~Zhong, W.~Luo, L.~Ma, and M.~Zhang, ``{Uni-MoE: Scaling Unified Multimodal LLMs with Mixture of Experts},'' \emph{arXiv preprint arXiv:2405.11273}, 2024.

\bibitem{mckinzie2024mm1}
B.~McKinzie, Z.~Gan, J.-P. Fauconnier, S.~Dodge, B.~Zhang, P.~Dufter, D.~Shah, X.~Du, F.~Peng, F.~Weers \emph{et~al.}, ``{Mm1: Methods, analysis \& insights from multimodal llm pre-training},'' \emph{arXiv preprint arXiv:2403.09611}, 2024.

\bibitem{jiang2023mistral}
A.~Q. Jiang, A.~Sablayrolles, A.~Mensch, C.~Bamford, D.~S. Chaplot, D.~d.~l. Casas, F.~Bressand, G.~Lengyel, G.~Lample, L.~Saulnier \emph{et~al.}, ``{Mistral 7B},'' \emph{arXiv preprint arXiv:2310.06825}, 2023.

\bibitem{touvron2023llama}
H.~Touvron, L.~Martin, K.~Stone, P.~Albert, A.~Almahairi, Y.~Babaei, N.~Bashlykov, S.~Batra, P.~Bhargava, S.~Bhosale \emph{et~al.}, ``{Llama 2: Open foundation and fine-tuned chat models},'' \emph{arXiv preprint arXiv:2307.09288}, 2023.

\bibitem{gpt-3.5-turbo}
OpenAI, ``{Chatgpt: Optimizing language models for dialogue},'' \url{https://openai.com/blog/chatgpt}, 2022.

\bibitem{bi2024deepseek}
X.~Bi, D.~Chen, G.~Chen, S.~Chen, D.~Dai, C.~Deng, H.~Ding, K.~Dong, Q.~Du, Z.~Fu \emph{et~al.}, ``{Deepseek llm: Scaling open-source language models with longtermism},'' \emph{arXiv preprint arXiv:2401.02954}, 2024.

\bibitem{qwen1.5}
\BIBentryALTinterwordspacing
Q.~Team, ``{Introducing Qwen1.5},'' February 2024. [Online]. Available: \url{https://qwenlm.github.io/blog/qwen1.5/}
\BIBentrySTDinterwordspacing

\bibitem{bengio2013estimating}
Y.~Bengio, N.~L{\'e}onard, and A.~Courville, ``{Estimating or propagating gradients through stochastic neurons for conditional computation},'' \emph{arXiv preprint arXiv:1308.3432}, 2013.

\bibitem{davis2013low}
A.~Davis and I.~Arel, ``{Low-rank approximations for conditional feedforward computation in deep neural networks},'' \emph{arXiv preprint arXiv:1312.4461}, 2013.

\bibitem{almahairi2016dynamic}
A.~Almahairi, N.~Ballas, T.~Cooijmans, Y.~Zheng, H.~Larochelle, and A.~Courville, ``{Dynamic capacity networks},'' in \emph{International Conference on Machine Learning}.\hskip 1em plus 0.5em minus 0.4em\relax PMLR, 2016, pp. 2549--2558.

\bibitem{bengio2015conditional}
E.~Bengio, P.-L. Bacon, J.~Pineau, and D.~Precup, ``{Conditional computation in neural networks for faster models},'' \emph{arXiv preprint arXiv:1511.06297}, 2015.

\bibitem{rosenbaum2017routing}
C.~Rosenbaum, T.~Klinger, and M.~Riemer, ``{Routing networks: Adaptive selection of non-linear functions for multi-task learning},'' \emph{arXiv preprint arXiv:1711.01239}, 2017.

\bibitem{rosenbaum2019routing}
C.~Rosenbaum, I.~Cases, M.~Riemer, and T.~Klinger, ``{Routing networks and the challenges of modular and compositional computation},'' \emph{arXiv preprint arXiv:1904.12774}, 2019.

\bibitem{raffel2020exploring}
C.~Raffel, N.~Shazeer, A.~Roberts, K.~Lee, S.~Narang, M.~Matena, Y.~Zhou, W.~Li, and P.~J. Liu, ``{Exploring the limits of transfer learning with a unified text-to-text transformer},'' \emph{Journal of machine learning research}, vol.~21, no. 140, pp. 1--67, 2020.

\bibitem{li2022lazy}
Z.~Li, C.~You, S.~Bhojanapalli, D.~Li, A.~S. Rawat, S.~J. Reddi, K.~Ye, F.~Chern, F.~Yu, R.~Guo \emph{et~al.}, ``The lazy neuron phenomenon: On emergence of activation sparsity in transformers,'' \emph{arXiv preprint arXiv:2210.06313}, 2022.

\bibitem{gross2017hard}
S.~Gross, M.~Ranzato, and A.~Szlam, ``{Hard mixtures of experts for large scale weakly supervised vision},'' in \emph{Proceedings of the IEEE Conference on Computer Vision and Pattern Recognition}, 2017, pp. 6865--6873.

\bibitem{hu2021lora}
E.~J. Hu, P.~Wallis, Z.~Allen-Zhu, Y.~Li, S.~Wang, L.~Wang, W.~Chen \emph{et~al.}, ``{LoRA: Low-Rank Adaptation of Large Language Models},'' in \emph{International Conference on Learning Representations}, 2021.

\bibitem{muennighoff2024olmoe}
N.~Muennighoff, L.~Soldaini, D.~Groeneveld, K.~Lo, J.~Morrison, S.~Min, W.~Shi, P.~Walsh, O.~Tafjord, N.~Lambert \emph{et~al.}, ``Olmoe: Open mixture-of-experts language models,'' \emph{arXiv preprint arXiv:2409.02060}, 2024.

\bibitem{liu2024deepseek}
A.~Liu, B.~Feng, B.~Xue, B.~Wang, B.~Wu, C.~Lu, C.~Zhao, C.~Deng, C.~Zhang, C.~Ruan \emph{et~al.}, ``Deepseek-v3 technical report,'' \emph{arXiv preprint arXiv:2412.19437}, 2024.

\bibitem{lample2019large}
G.~Lample, A.~Sablayrolles, M.~Ranzato, L.~Denoyer, and H.~J{\'e}gou, ``Large memory layers with product keys,'' \emph{Advances in Neural Information Processing Systems}, vol.~32, 2019.

\bibitem{gu2023mamba}
A.~Gu and T.~Dao, ``{Mamba: Linear-time sequence modeling with selective state spaces},'' \emph{arXiv preprint arXiv:2312.00752}, 2023.

\bibitem{hendrycks2020measuring}
D.~Hendrycks, C.~Burns, S.~Basart, A.~Zou, M.~Mazeika, D.~Song, and J.~Steinhardt, ``{Measuring massive multitask language understanding},'' \emph{arXiv preprint arXiv:2009.03300}, 2020.

\bibitem{cobbe2021training}
K.~Cobbe, V.~Kosaraju, M.~Bavarian, M.~Chen, H.~Jun, L.~Kaiser, M.~Plappert, J.~Tworek, J.~Hilton, R.~Nakano \emph{et~al.}, ``{Training verifiers to solve math word problems},'' \emph{arXiv preprint arXiv:2110.14168}, 2021.

\bibitem{hendrycks2021measuring}
D.~Hendrycks, C.~Burns, S.~Kadavath, A.~Arora, S.~Basart, E.~Tang, D.~Song, and J.~Steinhardt, ``{Measuring mathematical problem solving with the math dataset},'' \emph{arXiv preprint arXiv:2103.03874}, 2021.

\bibitem{chen2021evaluating}
M.~Chen, J.~Tworek, H.~Jun, Q.~Yuan, H.~P. d.~O. Pinto, J.~Kaplan, H.~Edwards, Y.~Burda, N.~Joseph, G.~Brockman \emph{et~al.}, ``{Evaluating large language models trained on code},'' \emph{arXiv preprint arXiv:2107.03374}, 2021.

\bibitem{devlin2018bert}
J.~Devlin, M.-W. Chang, K.~Lee, and K.~Toutanova, ``{Bert: Pre-training of deep bidirectional transformers for language understanding},'' \emph{arXiv preprint arXiv:1810.04805}, 2018.

\bibitem{zhang2019root}
B.~Zhang and R.~Sennrich, ``{Root mean square layer normalization},'' \emph{Advances in Neural Information Processing Systems}, vol.~32, 2019.

\bibitem{ainslie2023gqa}
J.~Ainslie, J.~Lee-Thorp, M.~de~Jong, Y.~Zemlyanskiy, F.~Lebron, and S.~Sanghai, ``{GQA: Training Generalized Multi-Query Transformer Models from Multi-Head Checkpoints},'' in \emph{Proceedings of the 2023 Conference on Empirical Methods in Natural Language Processing}, 2023, pp. 4895--4901.

\bibitem{su2024roformer}
J.~Su, M.~Ahmed, Y.~Lu, S.~Pan, W.~Bo, and Y.~Liu, ``{Roformer: Enhanced transformer with rotary position embedding},'' \emph{Neurocomputing}, vol. 568, p. 127063, 2024.

\bibitem{ding2022delta}
N.~Ding, Y.~Qin, G.~Yang, F.~Wei, Z.~Yang, Y.~Su, S.~Hu, Y.~Chen, C.-M. Chan, W.~Chen \emph{et~al.}, ``{Delta tuning: A comprehensive study of parameter efficient methods for pre-trained language models},'' \emph{arXiv preprint arXiv:2203.06904}, 2022.

\bibitem{han2024parameter}
Z.~Han, C.~Gao, J.~Liu, S.~Q. Zhang \emph{et~al.}, ``{Parameter-efficient fine-tuning for large models: A comprehensive survey},'' \emph{arXiv preprint arXiv:2403.14608}, 2024.

\bibitem{lialin2023scaling}
V.~Lialin, V.~Deshpande, and A.~Rumshisky, ``{Scaling down to scale up: A guide to parameter-efficient fine-tuning},'' \emph{arXiv preprint arXiv:2303.15647}, 2023.

\bibitem{liu2022few}
H.~Liu, D.~Tam, M.~Muqeeth, J.~Mohta, T.~Huang, M.~Bansal, and C.~A. Raffel, ``{Few-shot parameter-efficient fine-tuning is better and cheaper than in-context learning},'' \emph{Advances in Neural Information Processing Systems}, vol.~35, pp. 1950--1965, 2022.

\bibitem{ostapenko2023case}
O.~Ostapenko, L.~Caccia, Z.~Su, N.~Le~Roux, L.~Charlin, and A.~Sordoni, ``{A Case Study of Instruction Tuning with Mixture of Parameter-Efficient Experts},'' in \emph{NeurIPS 2023 Workshop on Instruction Tuning and Instruction Following}, 2023.

\bibitem{houlsby2019parameter}
N.~Houlsby, A.~Giurgiu, S.~Jastrzebski, B.~Morrone, Q.~De~Laroussilhe, A.~Gesmundo, M.~Attariyan, and S.~Gelly, ``{Parameter-efficient transfer learning for NLP},'' in \emph{International Conference on Machine Learning}.\hskip 1em plus 0.5em minus 0.4em\relax PMLR, 2019, pp. 2790--2799.

\bibitem{li2021prefix}
X.~L. Li and P.~Liang, ``{Prefix-Tuning: Optimizing Continuous Prompts for Generation},'' in \emph{Proceedings of the 59th Annual Meeting of the Association for Computational Linguistics and the 11th International Joint Conference on Natural Language Processing (Volume 1: Long Papers)}, 2021, pp. 4582--4597.

\bibitem{chen2023sparse}
\BIBentryALTinterwordspacing
T.~Chen, Z.~Zhang, A.~K. JAISWAL, S.~Liu, and Z.~Wang, ``{Sparse MoE as the New Dropout: Scaling Dense and Self-Slimmable Transformers},'' in \emph{The Eleventh International Conference on Learning Representations}, 2023. [Online]. Available: \url{https://openreview.net/forum?id=w1hwFUb_81}
\BIBentrySTDinterwordspacing

\bibitem{wei2023skywork}
T.~Wei, L.~Zhao, L.~Zhang, B.~Zhu, L.~Wang, H.~Yang, B.~Li, C.~Cheng, W.~L{\"u}, R.~Hu \emph{et~al.}, ``{Skywork: A more open bilingual foundation model},'' \emph{arXiv preprint arXiv:2310.19341}, 2023.

\bibitem{kirkpatrick2017overcoming}
J.~Kirkpatrick, R.~Pascanu, N.~Rabinowitz, J.~Veness, G.~Desjardins, A.~A. Rusu, K.~Milan, J.~Quan, T.~Ramalho, A.~Grabska-Barwinska \emph{et~al.}, ``{Overcoming catastrophic forgetting in neural networks},'' \emph{Proceedings of the national academy of sciences}, vol. 114, no.~13, pp. 3521--3526, 2017.

\bibitem{ott2019fairseq}
M.~Ott, S.~Edunov, A.~Baevski, A.~Fan, S.~Gross, N.~Ng, D.~Grangier, and M.~Auli, ``{fairseq: A fast, extensible toolkit for sequence modeling},'' \emph{arXiv preprint arXiv:1904.01038}, 2019.

\bibitem{shazeer2018mesh}
N.~Shazeer, Y.~Cheng, N.~Parmar, D.~Tran, A.~Vaswani, P.~Koanantakool, P.~Hawkins, H.~Lee, M.~Hong, C.~Young \emph{et~al.}, ``{Mesh-tensorflow: Deep learning for supercomputers},'' \emph{Advances in neural information processing systems}, vol.~31, 2018.

\bibitem{rajbhandari2020zero}
S.~Rajbhandari, J.~Rasley, O.~Ruwase, and Y.~He, ``{Zero: Memory optimizations toward training trillion parameter models},'' in \emph{SC20: International Conference for High Performance Computing, Networking, Storage and Analysis}.\hskip 1em plus 0.5em minus 0.4em\relax IEEE, 2020, pp. 1--16.

\bibitem{ren2021zero}
J.~Ren, S.~Rajbhandari, R.~Y. Aminabadi, O.~Ruwase, S.~Yang, M.~Zhang, D.~Li, and Y.~He, ``{$\{$Zero-offload$\}$: Democratizing $\{$billion-scale$\}$ model training},'' in \emph{2021 USENIX Annual Technical Conference (USENIX ATC 21)}, 2021, pp. 551--564.

\bibitem{rajbhandari2021zero}
S.~Rajbhandari, O.~Ruwase, J.~Rasley, S.~Smith, and Y.~He, ``{Zero-infinity: Breaking the gpu memory wall for extreme scale deep learning},'' in \emph{Proceedings of the international conference for high performance computing, networking, storage and analysis}, 2021, pp. 1--14.

\bibitem{ma2022bagualu}
Z.~Ma, J.~He, J.~Qiu, H.~Cao, Y.~Wang, Z.~Sun, L.~Zheng, H.~Wang, S.~Tang, T.~Zheng \emph{et~al.}, ``{BaGuaLu: targeting brain scale pretrained models with over 37 million cores},'' in \emph{Proceedings of the 27th ACM SIGPLAN Symposium on Principles and Practice of Parallel Programming}, 2022, pp. 192--204.

\bibitem{zheng2022alpa}
L.~Zheng, Z.~Li, H.~Zhang, Y.~Zhuang, Z.~Chen, Y.~Huang, Y.~Wang, Y.~Xu, D.~Zhuo, E.~P. Xing \emph{et~al.}, ``{Alpa: Automating inter-and $\{$Intra-Operator$\}$ parallelism for distributed deep learning},'' in \emph{16th USENIX Symposium on Operating Systems Design and Implementation (OSDI 22)}, 2022, pp. 559--578.

\bibitem{shoeybi2019megatron}
M.~Shoeybi, M.~Patwary, R.~Puri, P.~LeGresley, J.~Casper, and B.~Catanzaro, ``{Megatron-lm: Training multi-billion parameter language models using model parallelism},'' \emph{arXiv preprint arXiv:1909.08053}, 2019.

\bibitem{smith2022using}
S.~Smith, M.~Patwary, B.~Norick, P.~LeGresley, S.~Rajbhandari, J.~Casper, Z.~Liu, S.~Prabhumoye, G.~Zerveas, V.~Korthikanti \emph{et~al.}, ``{Using deepspeed and megatron to train megatron-turing nlg 530b, a large-scale generative language model},'' \emph{arXiv preprint arXiv:2201.11990}, 2022.

\bibitem{narayanan2021efficient}
D.~Narayanan, M.~Shoeybi, J.~Casper, P.~LeGresley, M.~Patwary, V.~Korthikanti, D.~Vainbrand, P.~Kashinkunti, J.~Bernauer, B.~Catanzaro \emph{et~al.}, ``{Efficient large-scale language model training on gpu clusters using megatron-lm},'' in \emph{Proceedings of the International Conference for High Performance Computing, Networking, Storage and Analysis}, 2021, pp. 1--15.

\bibitem{huang2019gpipe}
Y.~Huang, Y.~Cheng, A.~Bapna, O.~Firat, D.~Chen, M.~Chen, H.~Lee, J.~Ngiam, Q.~V. Le, Y.~Wu \emph{et~al.}, ``{Gpipe: Efficient training of giant neural networks using pipeline parallelism},'' \emph{Advances in neural information processing systems}, vol.~32, 2019.

\bibitem{narayanan2019pipedream}
D.~Narayanan, A.~Harlap, A.~Phanishayee, V.~Seshadri, N.~R. Devanur, G.~R. Ganger, P.~B. Gibbons, and M.~Zaharia, ``{PipeDream: generalized pipeline parallelism for DNN training},'' in \emph{Proceedings of the 27th ACM symposium on operating systems principles}, 2019, pp. 1--15.

\bibitem{qi2023zero}
P.~Qi, X.~Wan, G.~Huang, and M.~Lin, ``{Zero Bubble Pipeline Parallelism},'' in \emph{The Twelfth International Conference on Learning Representations}, 2023.

\bibitem{li2021sequence}
S.~Li, F.~Xue, C.~Baranwal, Y.~Li, and Y.~You, ``{Sequence parallelism: Long sequence training from system perspective},'' \emph{arXiv preprint arXiv:2105.13120}, 2021.

\bibitem{korthikanti2023reducing}
V.~A. Korthikanti, J.~Casper, S.~Lym, L.~McAfee, M.~Andersch, M.~Shoeybi, and B.~Catanzaro, ``{Reducing activation recomputation in large transformer models},'' \emph{Proceedings of Machine Learning and Systems}, vol.~5, 2023.

\bibitem{jacobs2023deepspeed}
S.~A. Jacobs, M.~Tanaka, C.~Zhang, M.~Zhang, L.~Song, S.~Rajbhandari, and Y.~He, ``{Deepspeed ulysses: System optimizations for enabling training of extreme long sequence transformer models},'' \emph{arXiv preprint arXiv:2309.14509}, 2023.

\bibitem{dosovitskiy2020image}
A.~Dosovitskiy, L.~Beyer, A.~Kolesnikov, D.~Weissenborn, X.~Zhai, T.~Unterthiner, M.~Dehghani, M.~Minderer, G.~Heigold, S.~Gelly \emph{et~al.}, ``{An image is worth 16x16 words: Transformers for image recognition at scale},'' \emph{arXiv preprint arXiv:2010.11929}, 2020.

\bibitem{zheng2022survey}
Y.~Zheng and D.~X. Wang, ``{A survey of recommender systems with multi-objective optimization},'' \emph{Neurocomputing}, vol. 474, pp. 141--153, 2022.

\bibitem{ngiam2011multimodal}
J.~Ngiam, A.~Khosla, M.~Kim, J.~Nam, H.~Lee, and A.~Y. Ng, ``{Multimodal deep learning},'' in \emph{Proceedings of the 28th international conference on machine learning (ICML-11)}, 2011, pp. 689--696.

\bibitem{baltruvsaitis2018multimodal}
T.~Baltru{\v{s}}aitis, C.~Ahuja, and L.-P. Morency, ``{Multimodal machine learning: A survey and taxonomy},'' \emph{IEEE transactions on pattern analysis and machine intelligence}, vol.~41, no.~2, pp. 423--443, 2018.

\bibitem{uppal2022multimodal}
S.~Uppal, S.~Bhagat, D.~Hazarika, N.~Majumder, S.~Poria, R.~Zimmermann, and A.~Zadeh, ``{Multimodal research in vision and language: A review of current and emerging trends},'' \emph{Information Fusion}, vol.~77, pp. 149--171, 2022.

\bibitem{zhou2020unified}
L.~Zhou, H.~Palangi, L.~Zhang, H.~Hu, J.~Corso, and J.~Gao, ``{Unified vision-language pre-training for image captioning and vqa},'' in \emph{Proceedings of the AAAI conference on artificial intelligence}, vol.~34, no.~07, 2020, pp. 13\,041--13\,049.

\bibitem{dao2022flashattention}
T.~Dao, D.~Fu, S.~Ermon, A.~Rudra, and C.~R{\'e}, ``{Flashattention: Fast and memory-efficient exact attention with io-awareness},'' \emph{Advances in Neural Information Processing Systems}, vol.~35, pp. 16\,344--16\,359, 2022.

\bibitem{kwon2023efficient}
W.~Kwon, Z.~Li, S.~Zhuang, Y.~Sheng, L.~Zheng, C.~H. Yu, J.~Gonzalez, H.~Zhang, and I.~Stoica, ``{Efficient memory management for large language model serving with pagedattention},'' in \emph{Proceedings of the 29th Symposium on Operating Systems Principles}, 2023, pp. 611--626.

\bibitem{shen2023mixture}
S.~Shen, L.~Hou, Y.~Zhou, N.~Du, S.~Longpre, J.~Wei, H.~W. Chung, B.~Zoph, W.~Fedus, X.~Chen \emph{et~al.}, ``{Mixture-of-experts meets instruction tuning: A winning combination for large language models},'' \emph{arXiv preprint arXiv:2305.14705}, 2023.

\bibitem{dou2023art}
S.~Dou, E.~Zhou, Y.~Liu, S.~Gao, J.~Zhao, W.~Shen, Y.~Zhou, Z.~Xi, X.~Wang, X.~Fan \emph{et~al.}, ``{The Art of Balancing: Revolutionizing Mixture of Experts for Maintaining World Knowledge in Language Model Alignment},'' \emph{arXiv preprint arXiv:2312.09979}, 2023.

\end{thebibliography}
